\documentclass[acmlarge]{acmart}
\AtBeginDocument{%
  }

\setcopyright{acmlicensed}
\acmJournal{IMWUT}
\acmYear{2023} \acmVolume{7} \acmNumber{2} \acmArticle{55} \acmMonth{6} \acmPrice{15.00}\acmDOI{10.1145/3596256}

\usepackage{graphicx}
\usepackage{xcolor}
\usepackage{color}
\usepackage[utf8]{inputenc}

\usepackage{cleveref}
\crefformat{section}{\S#2#1#3} 
\crefformat{subsection}{\S#2#1#3}
\crefformat{subsubsection}{\S#2#1#3}

\usepackage{algorithm}
\usepackage[noend]{algpseudocode}
\usepackage{float}
\usepackage{amsmath}
\usepackage{commath}
\algrenewcommand\algorithmicrequire{\textbf{Input:}}

\algrenewcommand\algorithmicensure{\textbf{Output:}}
\usepackage{flexisym}
\usepackage{relsize}

\usepackage{multirow, booktabs, makecell}  
\usepackage{arydshln}

\usepackage{subcaption} 
\usepackage{bbold} 

\usepackage{enumitem}

\definecolor{revc}{RGB}{0, 0, 0} 
\definecolor{REVC}{RGB}{0, 0, 0} 

\newcommand{\rev}[1]{{\color{revc}#1}}
\newcommand{\revi}[1]{\textcolor{black}{#1}}

\newcommand{\system}{DAPPER}

\begin{document}
\title{\system{}: Label-Free Performance Estimation after Personalization for Heterogeneous Mobile Sensing}



\author{Taesik Gong}
\orcid{https://orcid.org/0000-0002-8967-3652}
\affiliation{%
  \institution{School of Electrical Engineering, KAIST}
  \country{Republic of Korea}
  }
\email{taesik.gong@kaist.ac.kr}
\affiliation{%
  \institution{Nokia Bell Labs}
  \country{UK}
  }
\email{taesik.gong@nokia-bell-labs.com}

\author{Yewon Kim}
\orcid{https://orcid.org/0009-0006-2013-4278}
\affiliation{%
  \institution{School of Electrical Engineering, KAIST}
  \country{Republic of Korea}
  }
\email{yewon.e.kim@kaist.ac.kr}

\author{Adiba Orzikulova}
\orcid{https://orcid.org/0009-0006-4085-2701}
\affiliation{%
  \institution{School of Electrical Engineering, KAIST}
  \country{Republic of Korea}
  }
\email{adiorz@kaist.ac.kr}

\author{Yunxin Liu}
\orcid{https://orcid.org/0000-0001-7352-8955}
\affiliation{%
  \institution{Institute for AI Industry Research (AIR), Tsinghua University and Shanghai Artificial Intelligence Laboratory}
  \country{China}
  }
\email{liuyunxin@air.tsinghua.edu.cn}

\author{Sung Ju Hwang}
\orcid{https://orcid.org/0000-0002-9675-2324}
\affiliation{%
  \institution{Graduate School of AI, KAIST}
  \country{Republic of Korea}
  }
\email{sjhwang82@kaist.ac.kr}

\author{Jinwoo Shin}
\orcid{https://orcid.org/0000-0003-4313-4669}
\affiliation{%
  \institution{Graduate School of AI, KAIST}
  \country{Republic of Korea}
  }
\email{jinwoos@kaist.ac.kr}

\author{Sung-Ju Lee}
\orcid{https://orcid.org/0000-0002-5518-2126}
\affiliation{%
  \institution{School of Electrical Engineering, KAIST}
  \country{Republic of Korea}
  }
\email{profsj@kaist.ac.kr}

\renewcommand{\shortauthors}{Gong et al.}

\begin{abstract}
\label{rev:pp}

Many applications utilize sensors in mobile devices and machine learning to provide novel services. 
However, various factors such as different users, devices, and environments impact the performance of such applications, thus making the domain shift (i.e., distributional shift between the training domain and the target domain) a critical issue in mobile sensing. Despite attempts in domain adaptation to solve this challenging problem, their performance is unreliable due to the complex interplay among diverse factors. In principle, the performance uncertainty can be identified and redeemed by performance validation with ground-truth labels. However, it is infeasible for every user to collect high-quality, sufficient labeled data. To address the issue, we present \system{}~(Domain AdaPtation Performance EstimatoR) that estimates the adaptation performance in a target domain with only \emph{unlabeled} target data. Our key idea is to approximate the model performance based on the mutual information 
between the model inputs and corresponding outputs. Our evaluation with four real-world sensing datasets compared against six baselines shows that on average, \system{} outperforms the state-of-the-art baseline by \rev{39.8}\% in estimation accuracy. Moreover, our on-device experiment shows that \system{} achieves up to \revi{396$\times$} less computation overhead compared with the baselines.

\end{abstract}



\maketitle

\section{Introduction}\label{sec:introduction}

Mobile sensing utilizes the sensors from mobile devices (e.g., smartphones and wearables) to infer user contexts and provide appropriate services accordingly. Integrated with deep learning, which enables understanding of multi-dimensional sensory data, mobile sensing has broad applications ranging from human activity recognition~\cite{RONAO2016235, 10.1145/3161174, s19030714} to context recognition~\cite{9242267, 10.1145/3307334.3326074}, authentication~\cite{10.1145/3267242.3267252, 10.1145/3241539.3241575, 10.1145/3287036}, emotion recognition~\cite{10.1145/3191753, 10.1145/3264937}, and healthcare ~\cite{10.1145/3372224.3380889, 10.1145/3300061.3300125, 10.1145/3351281}. While a body of research in this area has demonstrated its potential, existing methods have been limited in real-world deployments due to the \textit{domain shift} issue, i.e., distribution shift of target data with respect to training (or source) data~\cite{hhar, 10.1145/2494091.2496039, metasense}. Specifically, users have different physical conditions and behaviors, and their mobile devices have various specifications, making each user's sensing distribution different from others. Therefore, a pre-trained model typically suffers from poor generalization to an unseen user or device. A popular approach in addressing this problem is \textit{domain adaptation} that utilizes labeled/unlabeled samples~\cite{metasense, systematic, scaling, xhar2020, generalization_fitness2021, multi_source2020, 10.1145/3432230} to adapt to the target condition. 

Domain adaptation for mobile sensing is particularly challenging as countless combinations between users and devices, the quality and quantity of the target samples, numerous hyperparameters, and even software/hardware are known to affect the model's performance~\cite{OLORISADE20171, 10.5555/3045118.3045343}. As such, the performance gain after adaptation is usually uncertain under such heterogeneous mobile sensing environments.

Two common practices used for model evaluation in domain adaptation have critical limitations to solve the aforementioned problem. First, many studies trained until a fixed number of training epochs (e.g., 100) and reported the last epoch's performance~\cite{shot, 10.5555/3045118.3045244, 10.1007/978-3-319-49409-8_35, shu2018a, NEURIPS2018_99607461, 10.1007/978-3-030-01225-0_28, 8953760, pmlr-v80-xie18c, pmlr-v70-long17a, Haoran_2020_ECCV, digging2020, xhar2020, generalization_fitness2021}. However, simply using the last epoch's model often leads to sub-optimal (or even worse) performance with unnecessary computational overhead. 
To avoid such undesirable situations, performance validation is critical regardless of adaptation algorithms. Second, 
existing studies used labeled test data from each target domain and reported the best accuracy~\cite{dou2019domain, contrastive2019, conditional2018, transferable2019, Tang_2020_CVPR, cui2020gvb, dada, xu2020adversarial, metasense, multi_source2020}. However, it is extremely costly and unrealistic to require every user to collect data and manually label them for validation. Our research question arises here: \emph{can we validate the model performance on the target domain without user effort?}

\begin{figure}[t]
    \centering
    \includegraphics[width=0.5\linewidth]{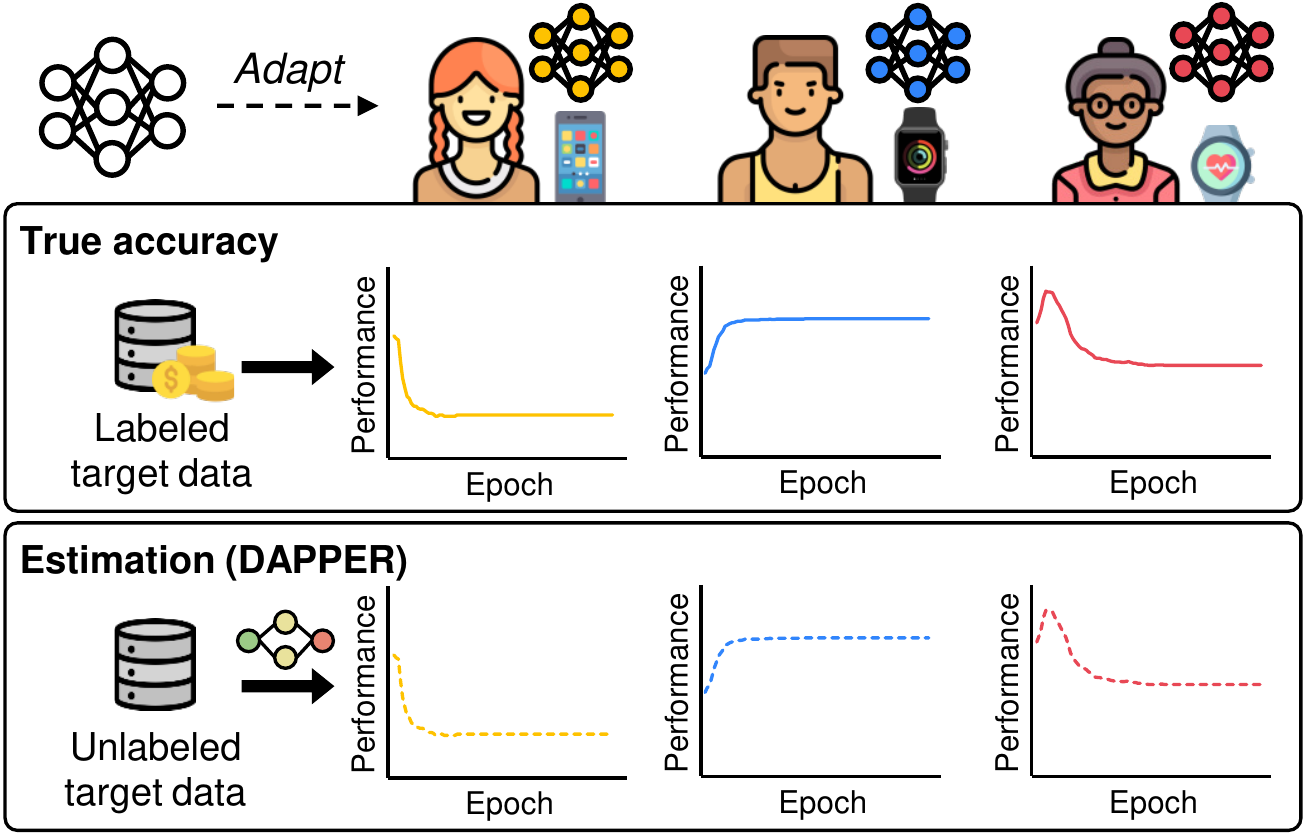}
    \caption{\system{} estimates the adaptation performance with only unlabeled data from the target user.}
    \vspace{-0.3cm}
    \label{fig:concept}
\end{figure}

We propose \textit{\system{}}~(Domain AdaPtation Performance EstimatoR), a label-free performance estimation for domain adaptation in mobile sensing (Figure~\ref{fig:concept}). Once deployed to users, \system{} leverages only unlabeled target data for performance estimation. As unlabeled time-series sensory data could be easily collected in mobile sensing applications, \system{} does not require any labeling effort. As \system{} needs training only once before deployment, it does not incur extra training overhead on target devices, which is beneficial in resource-constrained mobile devices. To the best of our knowledge, \system{} is the first proposal of performance estimation with unlabeled data in mobile sensing. Knowing the adapted performance has many benefits for the deployment of sensing models. One can avoid the aforementioned undesirable accuracy degradation or computational overhead. It could also help model maintenance; model developers could monitor whether their sensing applications provide desired performance for their customers and conduct additional actions when needed. Moreover, recent studies found that demonstrating the model's performance improves user trust in AI systems~\cite{10.1145/3290605.3300509, 10.1145/3287560.3287590, 10.1145/3301275.3302277}.

\system{}'s performance estimation with unlabeled data is based on information theory~\cite{shannon}. Specifically, we focus on the information gain in the model predictions given unlabeled target data, and then we demonstrate the correlation between the information gain and the actual performance on the target domain. Based on the analysis, we design a lightweight label-free performance estimation network that leverages domain knowledge, such as dataset-specific characteristics and the nature of the adaptation process. Note that training neural networks requires a large number of training instances for generalization. Without additional data collection, we generate virtual training data by simulating various adaptation scenarios only with source data.

We evaluate our proposed method with four real-world datasets: Heterogeneity Human Activity Recognition (HHAR)~\cite{hhar}, Wearable Stress and Affect Detection (WESAD)~\cite{schmidt2018introducing}, Individual-Condition Human Activity Recognition (ICHAR)~\cite{metasense}, and Individual-Condition Speech Recognition (ICSR)~\cite{metasense}. We demonstrate the effectiveness of our method compared with six baselines, including the state-of-the-art performance estimation algorithm for domain adaptation~\cite{ensrm}. Our results show that \system{} is the most accurate performance estimator, outperforming the state-of-the-art baseline by \rev{39.8}\% on average in estimation accuracy. In fact, we discovered that the state-of-the-art baselines do not perform well due to their assumptions that do not hold in mobile sensing scenarios.
We further evaluate the computational overhead by implementing \system{} and the baselines on mobile devices with the Mobile Neural Network~(MNN) framework~\cite{alibaba2020mnn}. The results show that \system{} requires 180$\sim$396$\times$ fewer computations than the state-of-the-art training-based algorithms and only 0.2\% extra computational overhead compared with validation from the labeled test data.

We summarize our contributions as follows:
\begin{itemize}
    \item We highlight the importance of performance validation in domain adaptation for mobile sensing by uncovering the performance dynamics under domain shifts and suggesting useful applications of performance validation.
    \item We present \system{}, the first domain adaptation performance estimation in mobile sensing with \emph{unlabeled} target data. \system{} combats the uncertainty of the adapted performance by modeling the relationship between model outputs and the performance. 
    \item We conduct extensive experiments with four real-world datasets against six baselines including the state-of-the-art performance estimation algorithm. Our evaluation indicates that \system{} not only outperforms the baselines in estimation accuracy but also requires significantly less computational overhead.
\end{itemize}

\section{Background and Motivation}~\label{sec:background}
\subsection{Performance Dynamics in Mobile Sensing}~\label{sec:background:dynamics}
In mobile sensing, users have different behaviors and physical characteristics. For instance, an elderly's jogging might be slower than a young person's jogging. Moreover, their mobile devices also have diverse hardware specifications, such as sensing sampling rates. In addition, the placement of mobile devices varies according to the type of device (smartphone vs. smartwatch) and user's preference (hand vs. pocket). These differences collectively create a unique sensing condition for each individual~\cite{hhar, 10.1145/2494091.2496039, metasense}. Unlike other research fields where domains are relatively well-defined and categorized (e.g., real-world objects vs. object icons in computer vision), every target user becomes a unique domain in mobile sensing. Therefore, \textit{domain shift}, i.e., distribution shift of the target data with respect to the source data, is inevitable in mobile sensing and is regarded as a major hurdle for realizing mobile sensing in the wild. 
To tackle this problem, various \textit{domain adaptation} (DA) algorithms have been proposed in mobile sensing; given a model trained on one or more source domains, a domain adaptation algorithm utilizes few labeled and/or unlabeled target data to adapt to the target domain~\cite{metasense, systematic, scaling, xhar2020, generalization_fitness2021, multi_source2020, 10.1145/3432230}. 

While previous research has focused on the algorithmic aspect of domain adaptation, we argue that the performance dynamics in the adaptation stage must be addressed. In addition to the user behaviors, numerous other factors, such as physical characteristics, the device's sensor specifications, and device placements, also affect the adapted performance. The quantity and quality of the data from the target user are usually not guaranteed (e.g., mislabeling, incorrect time stamps, skewed data distribution, etc.)~\cite{8366913}. When the model is deployed, both software (OSes, library versions, etc.) and hardware (GPUs, etc.) specifications on which adaptation is applied also impact the performance~\cite{OLORISADE20171}. These factors are dependent on every possible target user and difficult to control in the model development stage. Furthermore, traditional machine learning design choices, such as model architecture, set of hyperparameters, or training algorithms, lie in a huge space and are known to be extremely difficult to optimize for every single domain~\cite{10.5555/3045118.3045343}.

We conducted a preliminary empirical study to see how various factors affect adaptation performance. We used the Heterogeneity Human Activity Recognition (HHAR) dataset~\cite{hhar} and considered a common multi-source domain adaptation scenario: for each target domain, we pre-trained a base model with the remaining domains, and the base model was adapted to the target domain. We used a recent unsupervised domain adaptation algorithm (SHOT~\cite{shot}) with 300 unlabeled samples except for the mislabeling rate experiment, where we used fine-tuning with 25 labeled data. We used 1D convolutional neural networks (CNN) and the Adam optimizer with a fixed learning rate of 0.001 as the default. Additional details about the dataset and the training procedure are in~\cref{sec:experiment:settings}. We varied diverse user-related factors in mobile sensing: the user, the device, the number of given target samples, the mislabeling rate, and the distribution of samples with respect to classes. We also varied machine-learning hyperparameters: the learning rate, the optimizer, and the algorithm-specific hyperparameter (balancing factor). For each experiment, all other factors remained the same.

\begin{figure*}[t]
    \begin{subfigure}[t]{0.25\linewidth}
        \centering
  \includegraphics[width=0.95\linewidth]{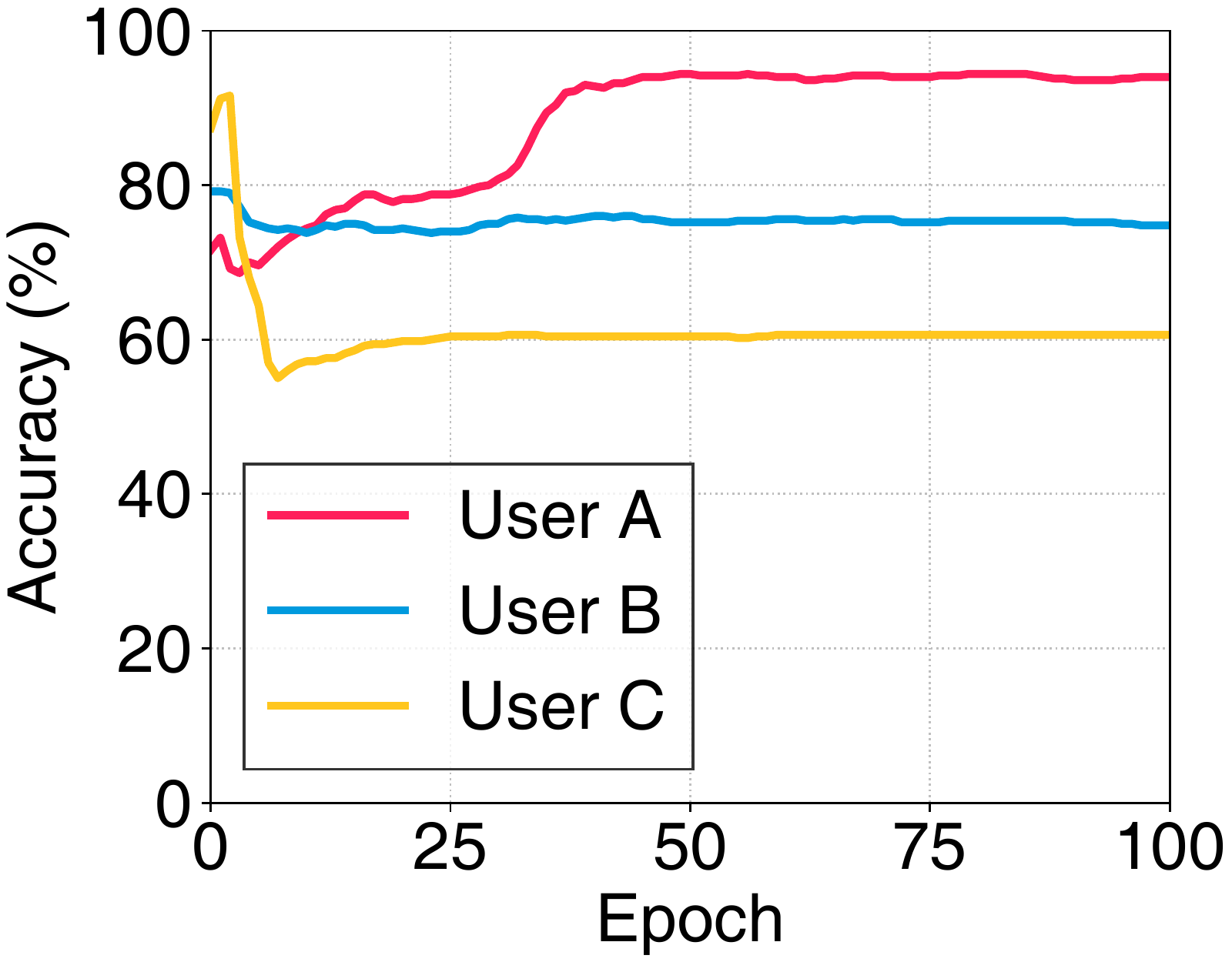}
        \vspace{-0.2cm}
        \caption{Target user.}
        \vspace{0.2cm}
        \label{fig:motivation1:target_user}
    \end{subfigure}
    ~
    \begin{subfigure}[t]{0.25\linewidth}
        \centering
  \includegraphics[width=0.95\linewidth]{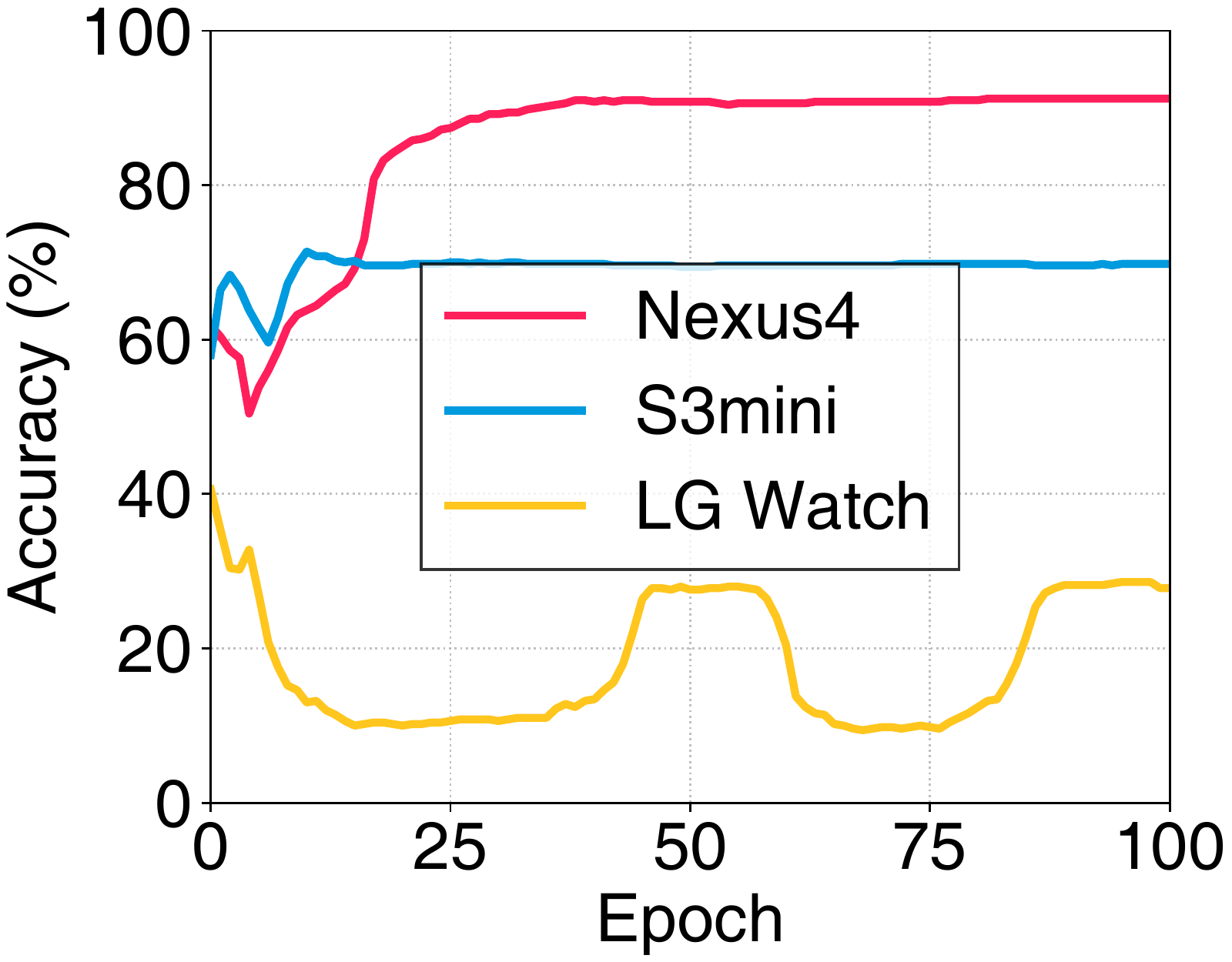}
        \vspace{-0.2cm}
        \caption{Target device.}
        \vspace{0.2cm}
        \label{fig:motivation1:target_device}
    \end{subfigure}
    ~
    \begin{subfigure}[t]{0.25\linewidth}
        \centering
  \includegraphics[width=0.95\linewidth]{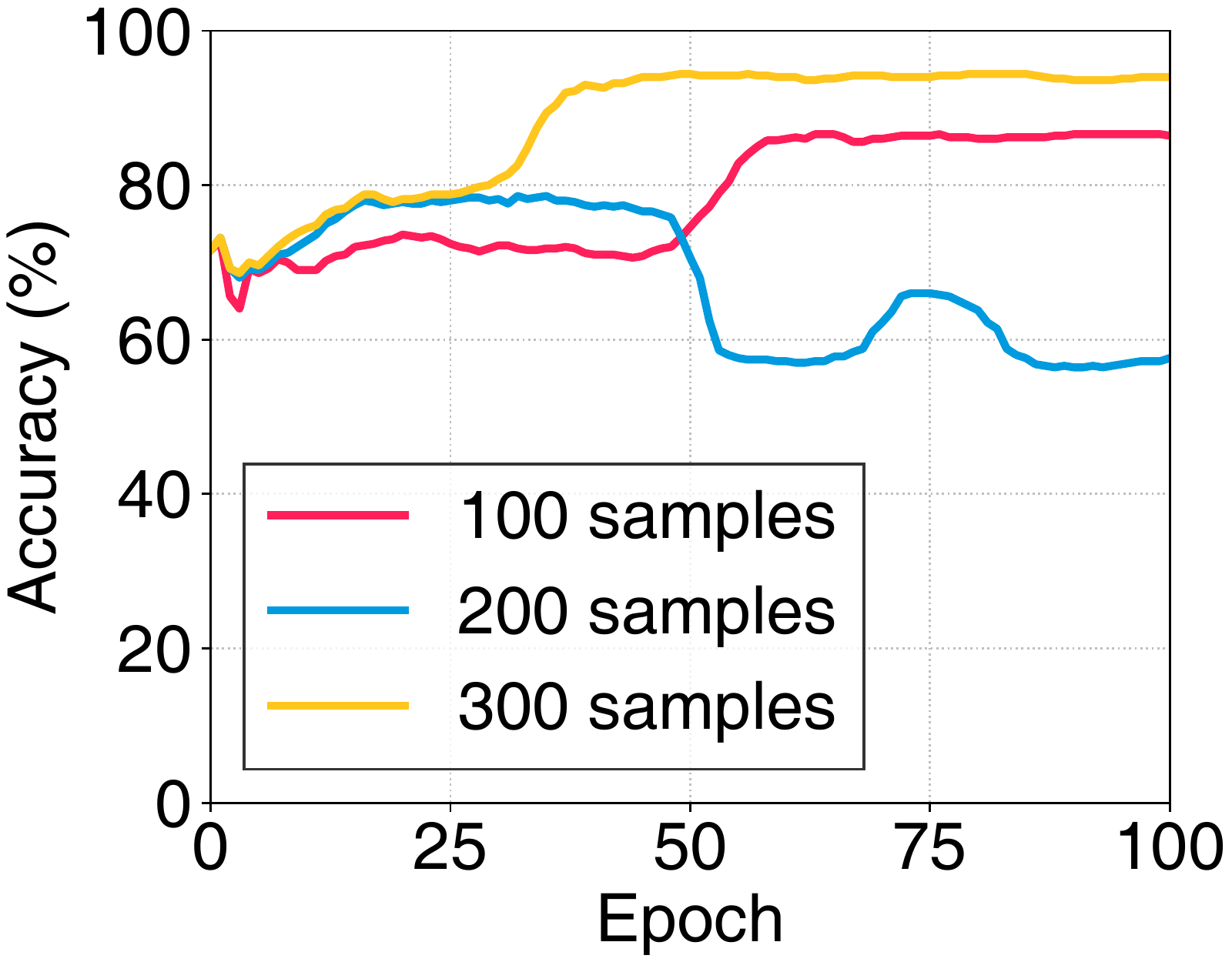}
        \vspace{-0.2cm}
        \caption{Number of given samples.}
        \vspace{0.2cm}
        \label{fig:motivation1:num_samples}
    \end{subfigure}
    ~
    \begin{subfigure}[t]{0.25\linewidth}
        \centering
  \includegraphics[width=0.95\linewidth]{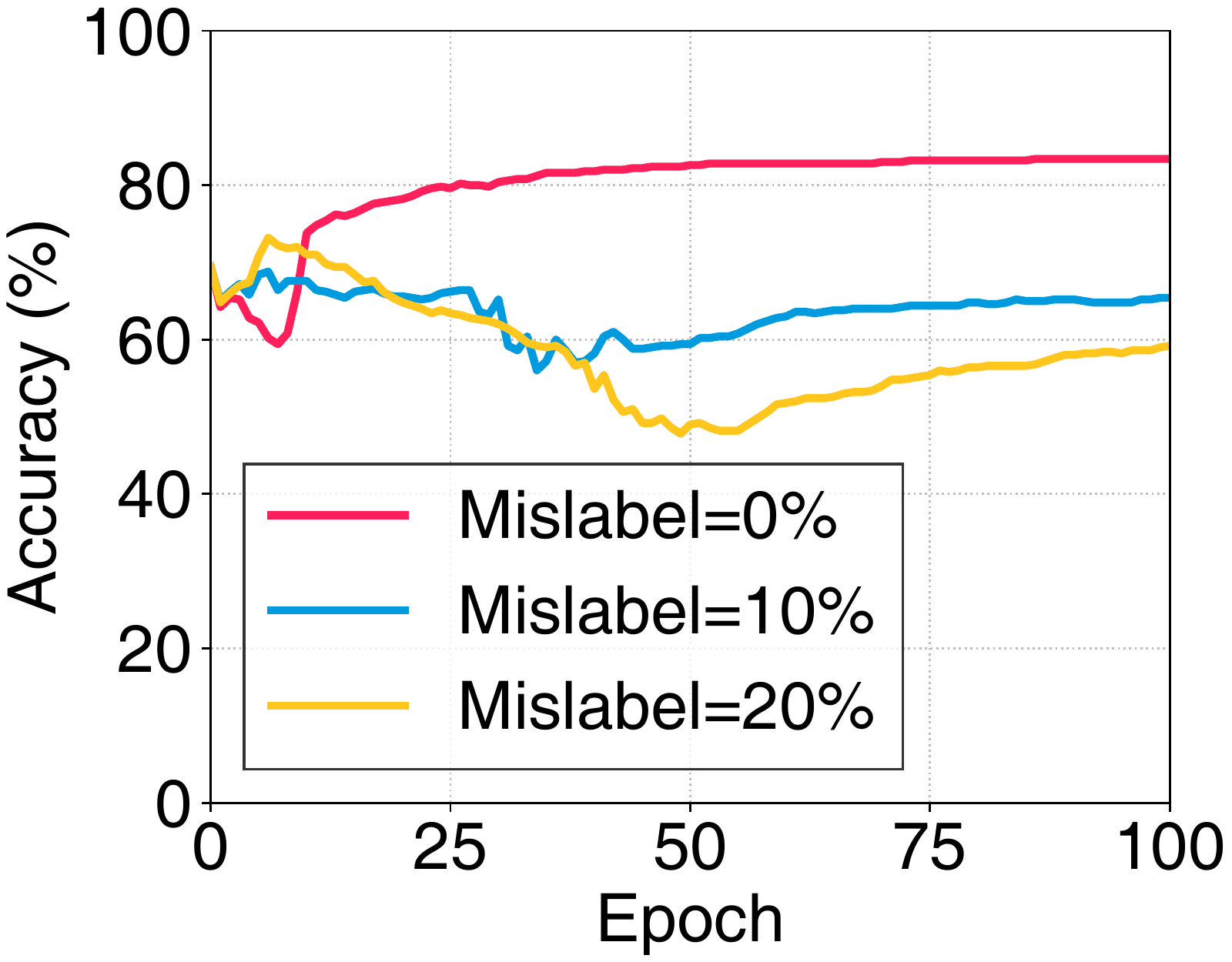}
        \vspace{-0.2cm}
        \caption{Mislabeling.}
        \vspace{0.2cm}
        \label{fig:motivation1:mislabel}
    \end{subfigure}
    
    \begin{subfigure}[t]{0.25\linewidth}
        \centering
  \includegraphics[width=0.95\linewidth]{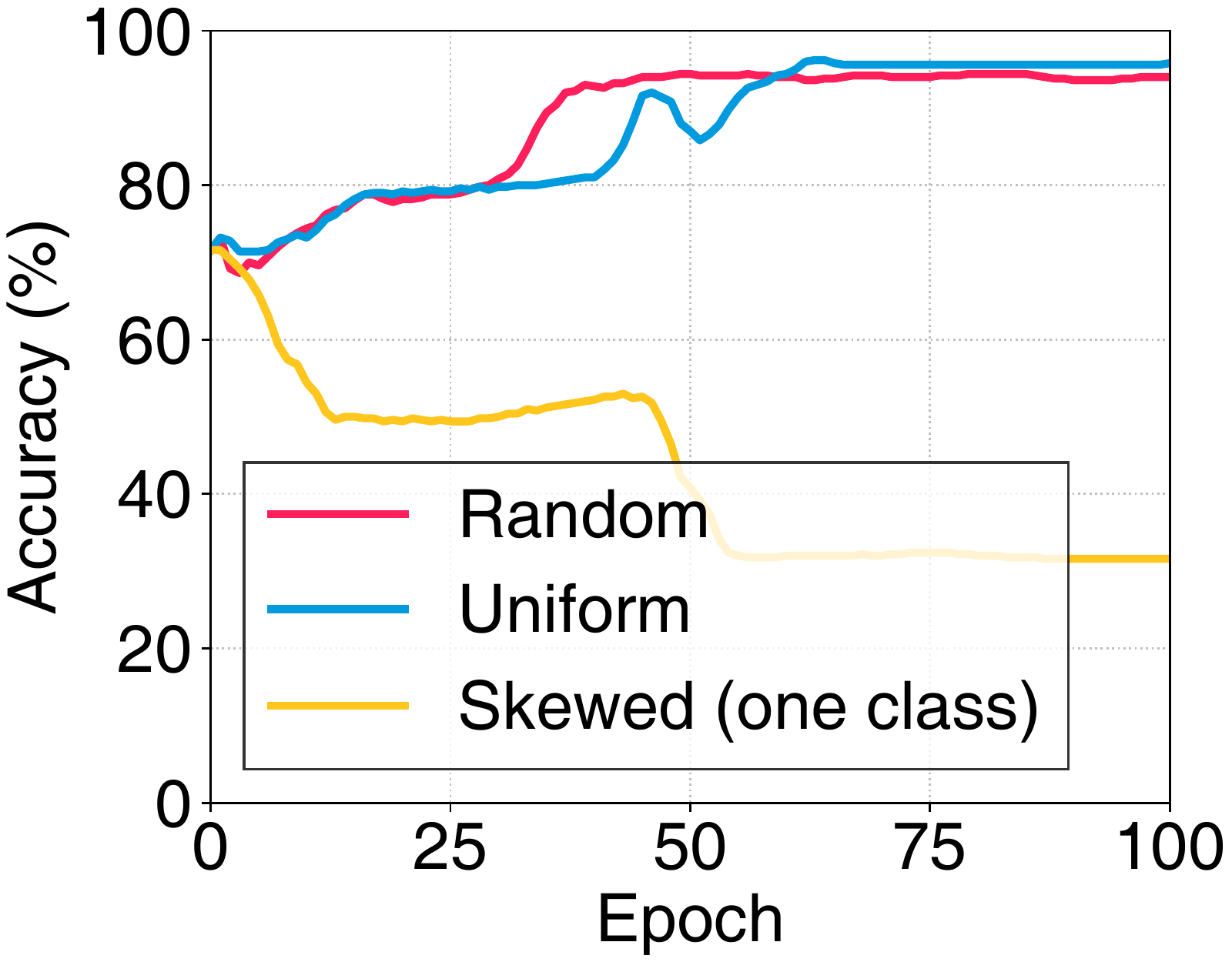}
        \vspace{-0.2cm}
        \caption{User data distribution.}
        \vspace{0.2cm}
        \label{fig:motivation1:sample_distribution}
    \end{subfigure}
    ~
    \begin{subfigure}[t]{0.25\linewidth}
        \centering
  \includegraphics[width=0.95\linewidth]{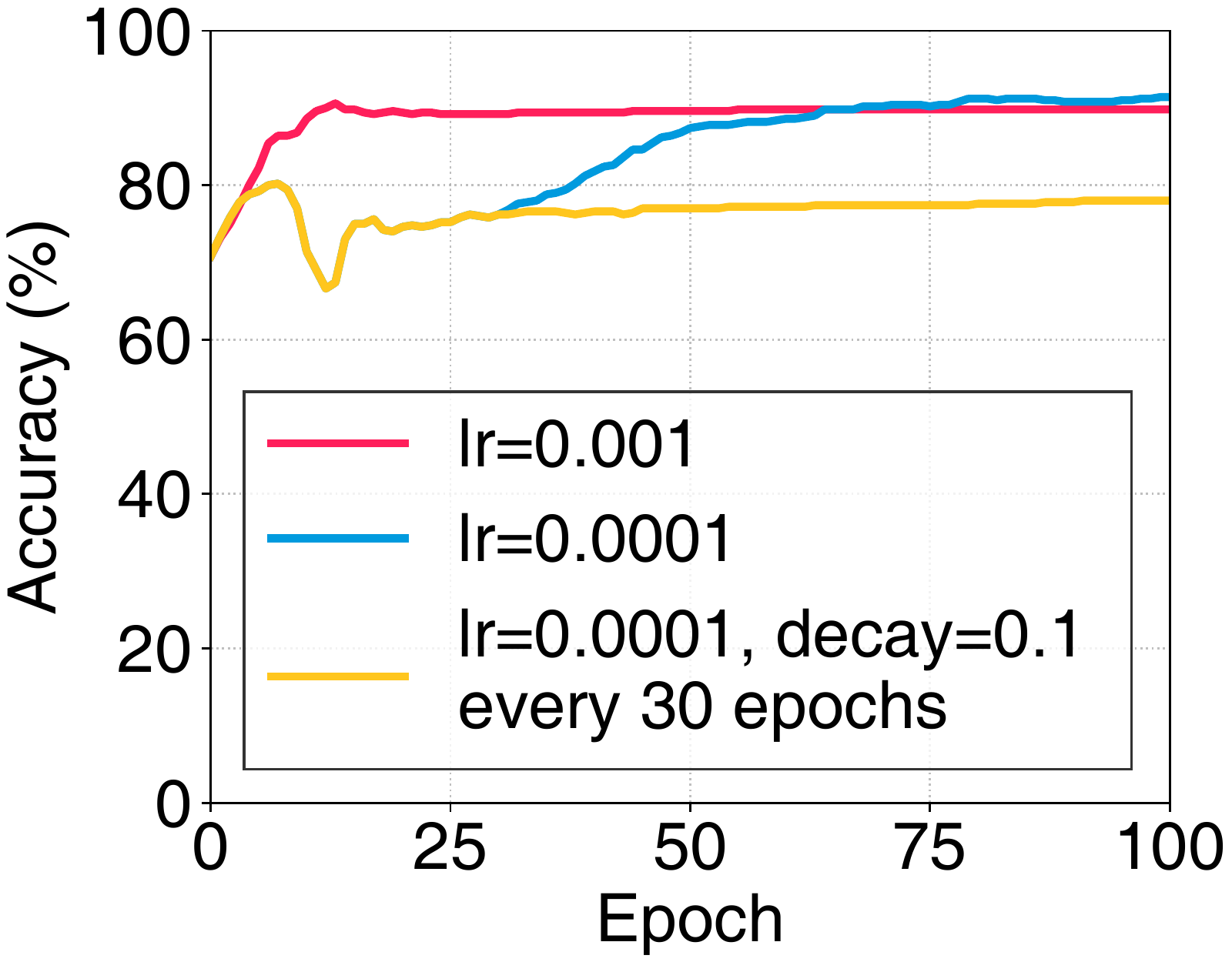}
        \vspace{-0.2cm}
        \caption{Learning rate.}
        \label{fig:motivation1:learning_rate}
    \end{subfigure}
    ~
    \begin{subfigure}[t]{0.25\linewidth}
        \centering
  \includegraphics[width=0.95\linewidth]{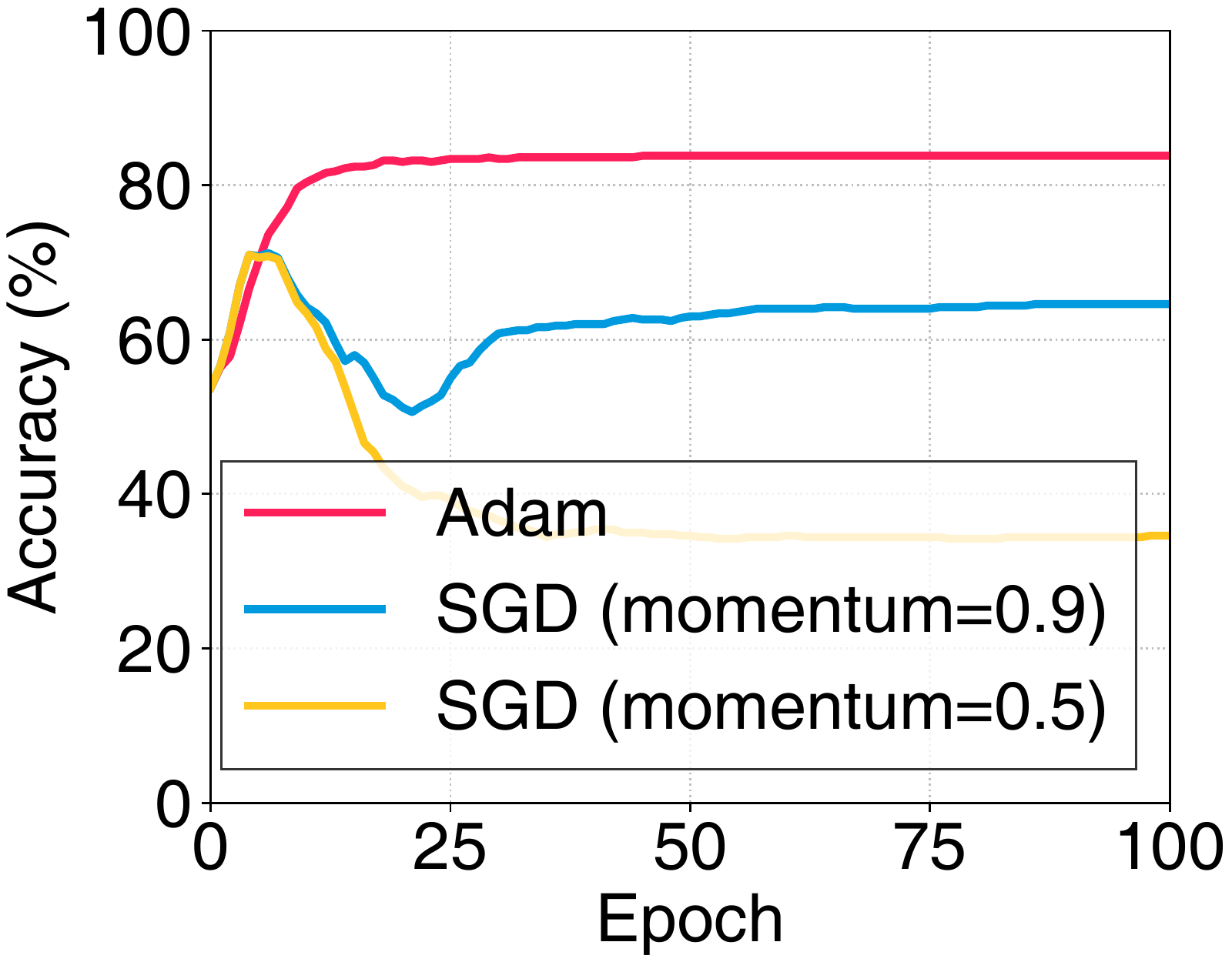}
        \vspace{-0.2cm}
        \caption{Optimizer.}
        \label{fig:motivation1:optimizer}
    \end{subfigure}
    ~
    \begin{subfigure}[t]{0.25\linewidth}
        \centering
  \includegraphics[width=0.95\linewidth]{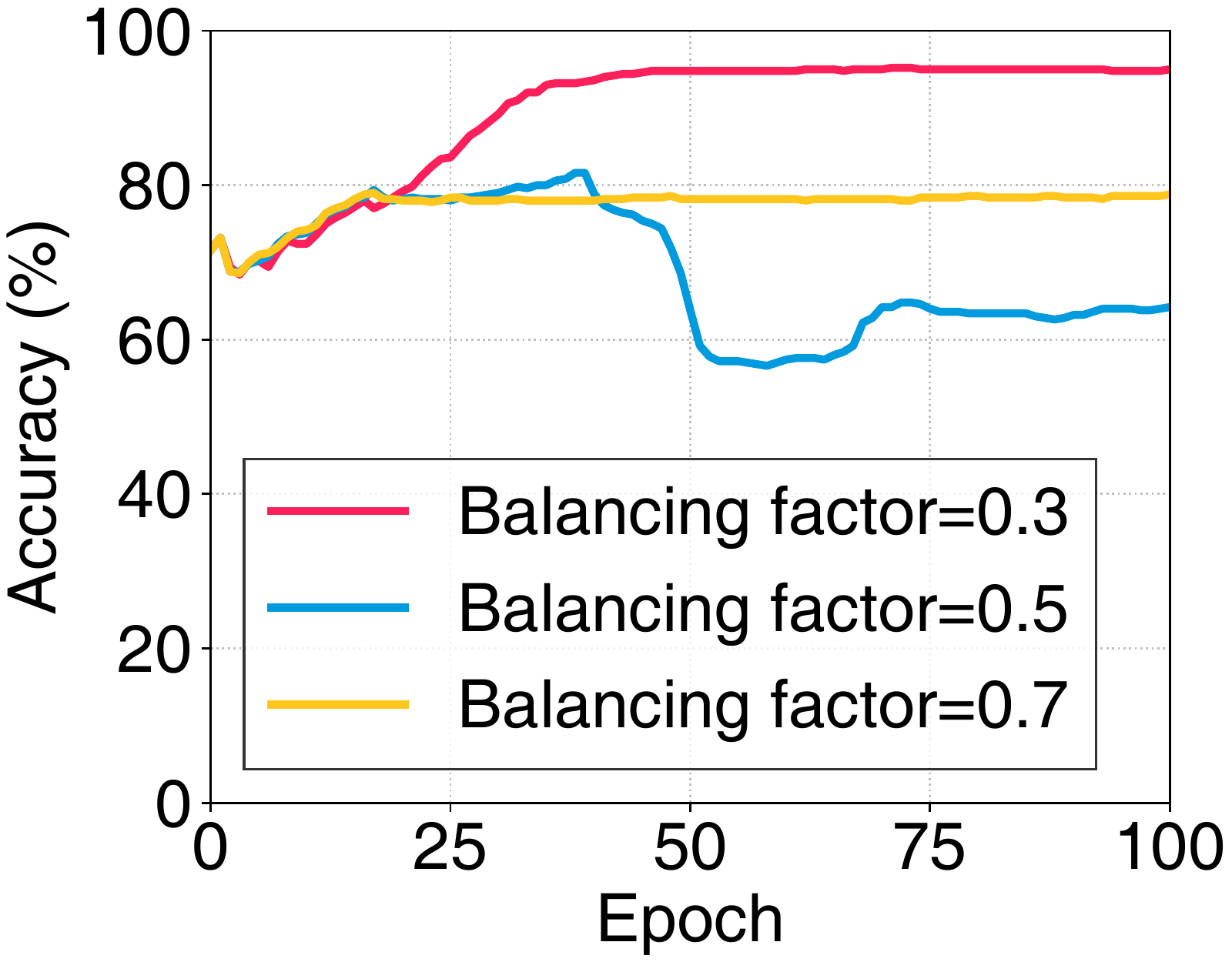}
        \vspace{-0.2cm}
        \caption{Balancing factor.}
        \label{fig:motivation1:balancing_factor}
    \end{subfigure}
    \vspace{-0.5cm}
  \caption{Performance dynamics in mobile sensing domain adaptation. We varied the (i) user-related factors in mobile sensing (the target user, the target device, the number of given samples, the mislabeling rate, and the sample distribution) and (ii) varied machine-learning hyperparameters (the learning rate, the optimizer, and the algorithm-specific hyperparameter).}
    \vspace{-0.4cm}
    \label{fig:motivation1}
\end{figure*}

Figure~\ref{fig:motivation1} visualizes the accuracy on the target test dataset as the adaptation proceeded for 100 training epochs. The adapted accuracy has diverse patterns according to a variety of factors. As multiple factors interplay and some are beyond the developer's control, it is difficult to predict the adapted performance. For instance, Figure~\ref{fig:motivation1:num_samples} shows that ``100 samples'' converged to a higher accuracy than ``200 samples,'' which contradicts the common expectation that training with more samples would result in higher accuracy. \rev{We found that this often happens as a higher number of samples does not guarantee a higher quality of the samples with respect to the algorithm's objective.}

\subsection{Limitations of Previous Approaches}~\label{sec:background:common}
\begin{figure}[t]
    \centering
    \begin{subfigure}[t]{0.25\linewidth}
        \centering
  \includegraphics[width=0.95\linewidth]{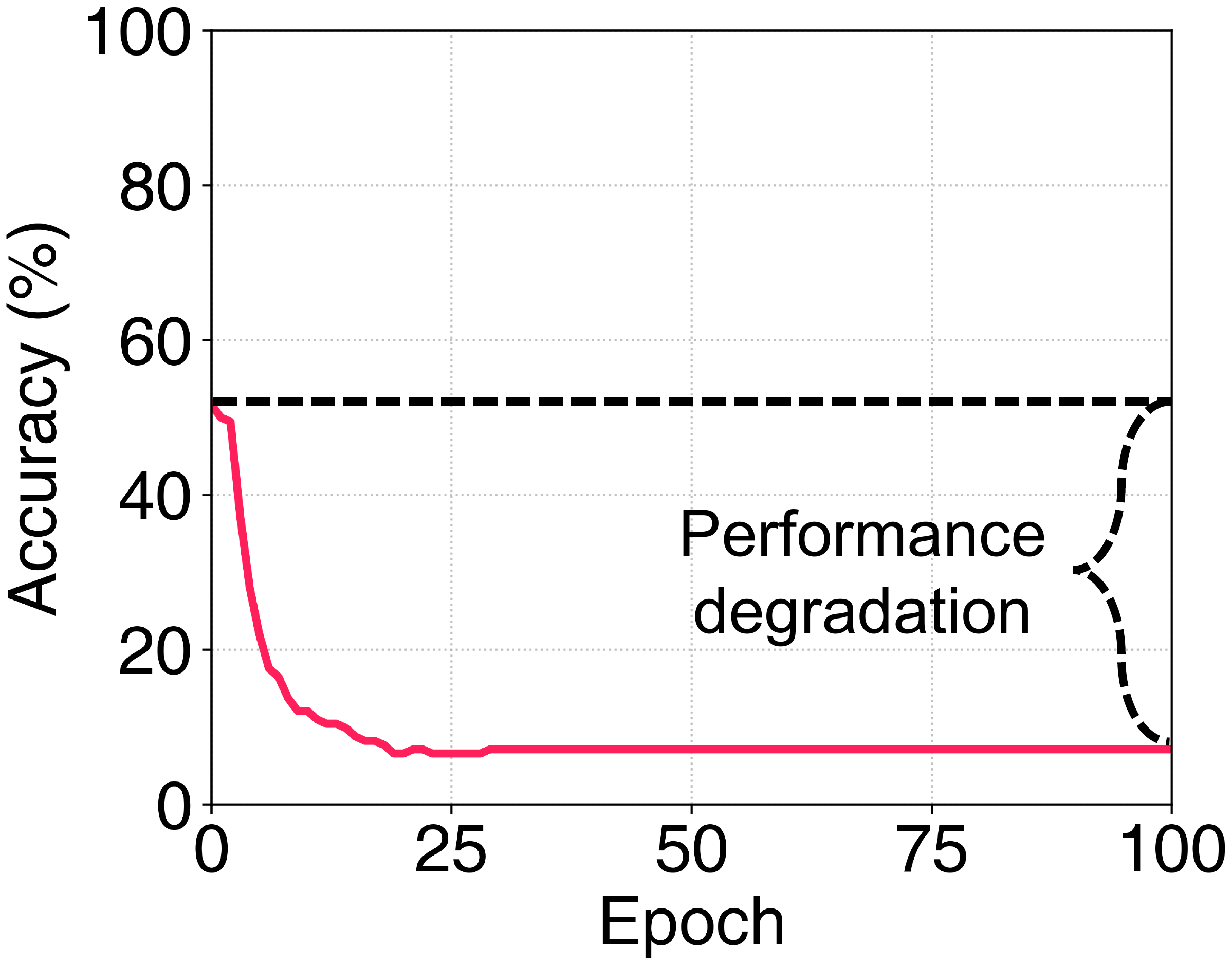}
        \label{fig:motivation2:degradation}
    \end{subfigure}
    ~
    \begin{subfigure}[t]{0.25\linewidth}
        \centering
  \includegraphics[width=0.95\linewidth]{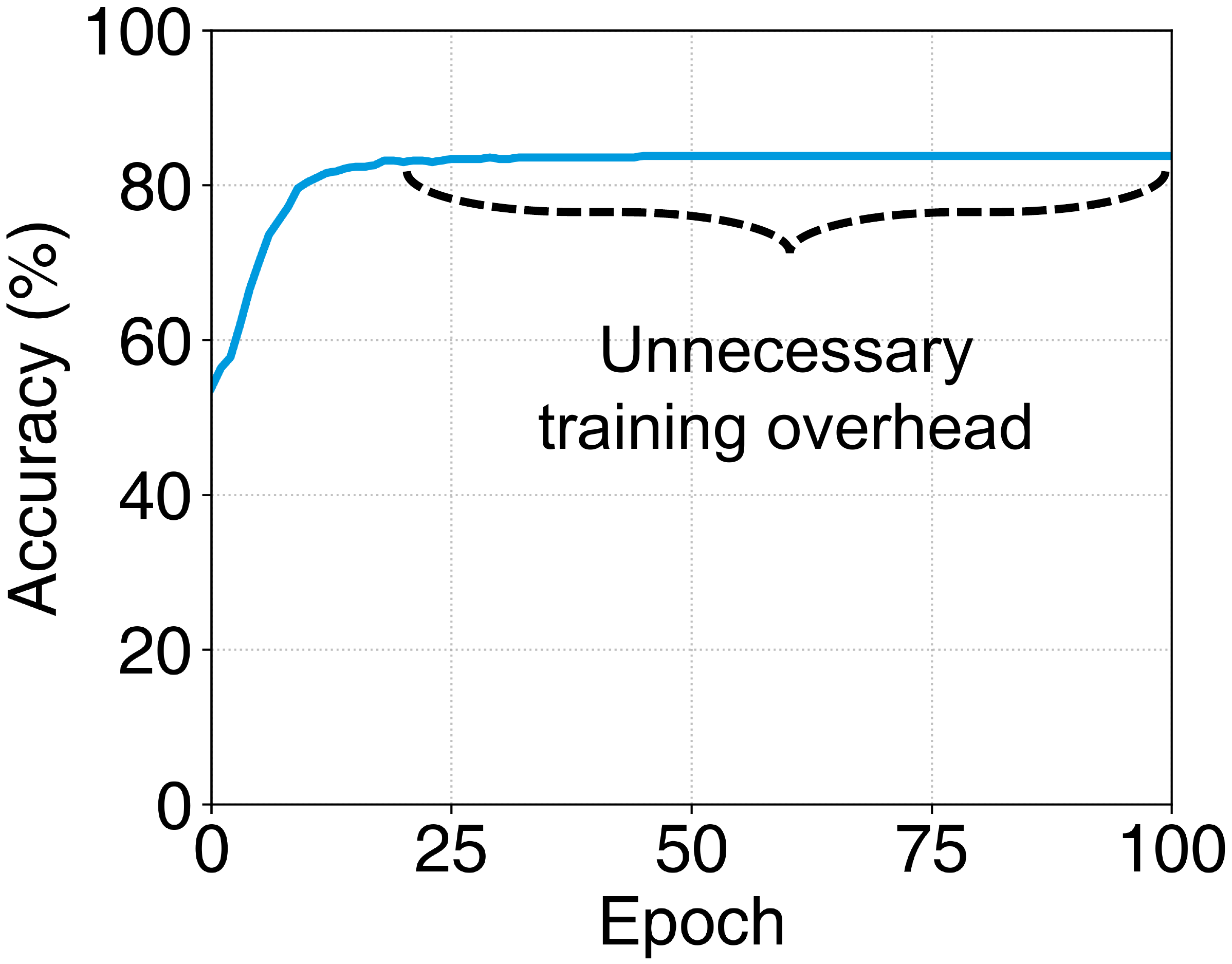}
        \label{fig:motivation2:overhead}
    \end{subfigure}
    ~
    \begin{subfigure}[t]{0.25\linewidth}
        \centering
  \includegraphics[width=0.95\linewidth]{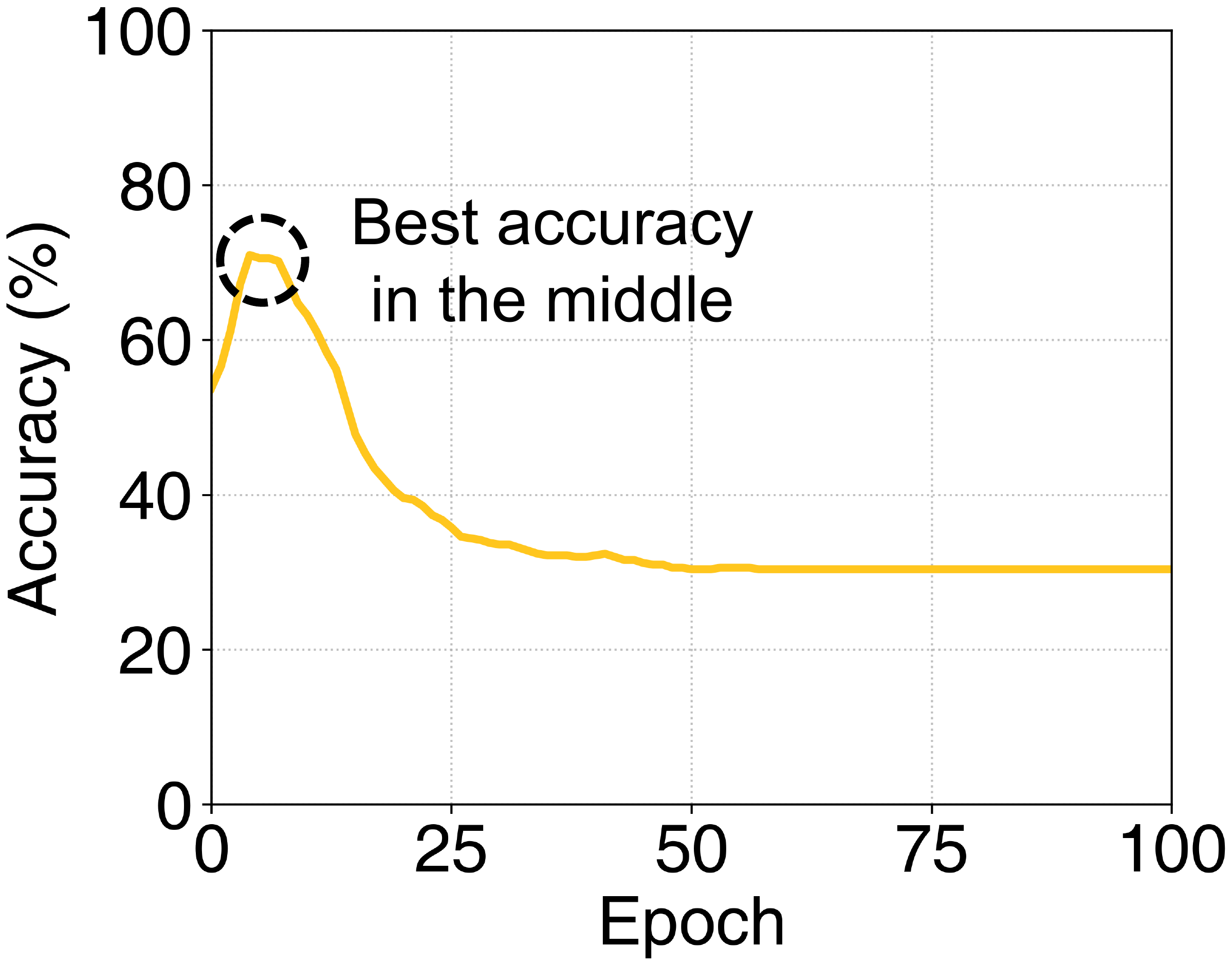}
        \label{fig:motivation2:flucutation}
    \end{subfigure}
    \vspace{-0.3cm}
  \caption{
  Three observed problems in domain adaptation for a fixed epoch without validation: (i) performance degradation, (ii) unnecessary training overhead, and (iii) best accuracy in the middle.}
  \vspace{-0.3cm}
    \label{fig:motivation2}
\end{figure}

We found that common approaches in performance validation for domain adaptation have critical limitations. Many studies 
used a fixed number of training epochs and reported the accuracy at the last epoch~\cite{shot, 10.5555/3045118.3045244, 10.1007/978-3-319-49409-8_35, shu2018a, NEURIPS2018_99607461, 10.1007/978-3-030-01225-0_28, 8953760, pmlr-v80-xie18c, pmlr-v70-long17a, Haoran_2020_ECCV, digging2020, xhar2020, generalization_fitness2021}. However, simply selecting the model at the last epoch suffers from several critical issues, as we illustrate in Figure~\ref{fig:motivation2}. First, in some cases, adaptation results in undesirable performance degradation due to various factors, as discussed in~\cref{sec:background:dynamics}.
Second, if the model's accuracy has rapidly saturated in the first few epochs, training for more epochs will bring in unnecessary computational overhead. Third, the model's accuracy can fluctuate and reach its maximum in the middle of training. Thus, selecting the last epoch's model without proper performance validation might lead to selecting a sub-optimal model. In summary, merely choosing the model at a certain point without validation would often be subject to performance issues, highlighting the importance of performance validation.

Note that the accuracies reported in Figures~\ref{fig:motivation1} and \ref{fig:motivation2} are calculated from the labeled test data. Similarly, existing studies often used hold-out labeled test data to compute accuracy and reported the best model's accuracy among epochs to validate the model's performance precisely ~\cite{dou2019domain, contrastive2019, conditional2018, transferable2019, Tang_2020_CVPR, cui2020gvb, dada, xu2020adversarial, metasense, multi_source2020}. However, it is extremely unrealistic for every user to collect an adequate amount of high-quality data and manually label them. Moreover, non-expert users could often mislabel data, which further harms the reliability of the labels~\cite{10.1145/3314419}.

\subsection{Applications of Performance Validation in Mobile Sensing}~\label{sec:background:importance}
Knowing the adapted performance has many benefits in the deployment of sensing models. We summarize the importance of performance validation in practical settings.
\subsubsection{Avoiding Performance Degradation} With the user and device factor added to the complex hyperparameter space of deep learning, domain adaptation in mobile sensing does not necessarily guarantee performance improvement for an arbitrary user. Moreover, the performance often degrades after adaptation, as shown in Figure~\ref{fig:motivation2}. If the accuracy after adaptation could be estimated, one can avoid such accuracy degradation by discarding the updated model.
\subsubsection{Reducing Computational Overhead} Performance validation could reduce unnecessary computation overhead, which is crucial in resource-constrained devices. For instance, a model sometimes converges before it reaches the last training epoch, as illustrated in Figure~\ref{fig:motivation2}. With the performance knowledge of the adapted model, the system could halt adaptation and avoid redundant computation when the model's performance does not improve for consecutive training iterations or reaches a satisfactory performance.
\subsubsection{Model Development and Maintenance} Performance validation could help model maintenance. Model developers (or companies) could monitor whether their sensing applications are providing desired performance for their customers and conduct additional actions when needed. For instance, developers could group poorly-performing models and improve them by identifying the common characteristics of the users. Furthermore, developers could provide personalization such as selecting the best hyperparameters/models for a specific user to improve performance.
\subsubsection{User-Interactive Adaptation} Recent studies have demonstrated that providing the model performance enhances users' trust and acceptance of ML systems, which leads to improved user experiences~\cite{10.1145/3290605.3300509, 10.1145/3287560.3287590, 10.1145/3301275.3302277}. 
Knowing the performance of a model allows designing a user-interactive adaptation algorithm. For instance, the model could proactively ask the user to collect more (labeled) data when the current performance is poor. When the presented performance is not satisfactory, the user could accept the request to gather more data. After adaptation, the model provides the user with the performance gain from the help of the user's data collection effort. If the model performs well, there is no need to collect further data. This AI-assisted decision-making not only lets users understand the model's current behavior better but also effectively utilizes users' feedback for model improvement.

\section{\system{}: A Label-free Performance Estimator for Mobile Sensing}

\begin{figure*}[t]
    \centering
    \includegraphics[width=1\linewidth]{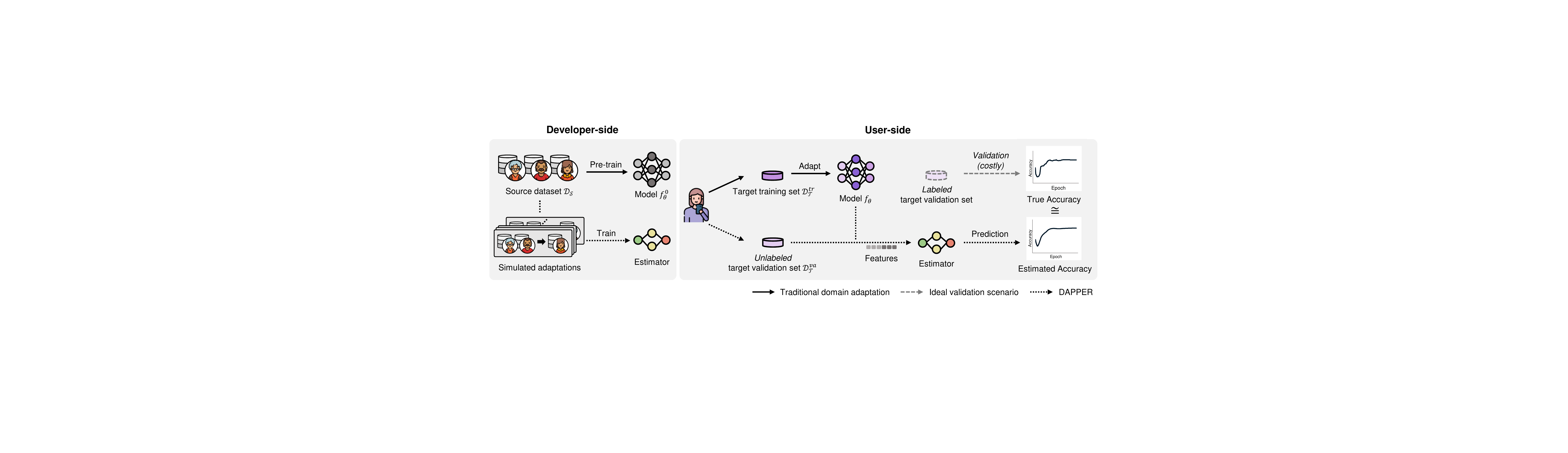}
    \vspace{-0.3cm}
    \caption{An overview of \system{} that estimates the performance of an adapted model with unlabeled target data. }
    \vspace{-0.3cm}
    \label{fig:overview}
\end{figure*}

\subsection{Overview}\label{sec:method:overview}

Our goal is to validate the performance of a model under heterogeneous mobile sensing scenarios without user effort. 
Figure~\ref{fig:overview} shows the overall workflow of our performance estimation framework, \textit{\system{}}.
We follow a common domain adaptation setting, where a pre-trained model with source data is deployed as an initial model and is adapted to the target user with some training data from the target domain~\cite{metasense, shot}. \system{} estimates the change in the performance as close as possible to the change in the performance calculated from the labeled data, utilizing only unlabeled data from the target domain. Note that it is extremely challenging, if not impossible, to compute the exact performance for a target domain without labeled data. 
We approach this problem with information theory~\cite{shannon}. By analyzing the information gain between the inputs and the outputs of the model, we demonstrate the correlation between the information gain and the actual performance on the target domain. Furthermore, we design a performance estimator to overcome the challenges in applying the theoretical analysis in practice.

\noindent\textbf{Design principles of \system{} for mobile sensing:} \system{} is designed by considering the characteristics and challenges of mobile sensing. First, collecting unlabeled time-series sensory data from a target user could be done in mobile sensing without the user's manual effort~\cite{BHATTACHARYA2014242, Tang_2021}. Taking human activity recognition as an example, while a user is walking, unlabeled sensory data of walking comes to the recognition app. \system{} utilizes such labor-free unlabeled data to estimate the performance of the adapted model. Second, obtaining high-quality labeled data in mobile sensing is costly as it involves users in the data collection and labeling process. Moreover, lots of labeled data are necessary to train an accurate estimator. \system{} mitigates this problem by training with \textit{simulated} data in which diverse adaptation scenarios are generated from the already collected source data. Third, minimizing the computation overhead of an estimation model is critical in resource-constrained mobile/wearable devices. By design, \system{} requires training only once from the developer side and no further training from the user side. Moreover, \system{}'s lightweight features and architecture collectively make the inference overhead on the user side negligible.

\subsection{Problem Formulation}\label{sec:method:problem}
Formally, let $\mathcal{D}_{\mathcal{S}} = \{\boldsymbol{x}_{\mathcal{S}}^{(i)}, y_{\mathcal{S}}^{(i)}\}_{i=1}^{N_{\mathcal{S}}}$ be the source data collected from one or more domains and $\mathcal{D}_{\mathcal{T}}=\{\boldsymbol{x}_{\mathcal{T}}^{(i)}, y_{\mathcal{T}}^{(i)}\}_{i=1}^{N_{\mathcal{T}}}$ be the target data to adapt to, where $x^{(i)}$ and $y^{(i)}$ denote data instances and their corresponding labels regardless of their subscripts. We follow the standard covariate shift assumption in domain adaptation, which is defined as $P_{\mathcal{S}}(\mathbf{x}) \neq P_{\mathcal{T}}(\mathbf{x})$ and $P_{\mathcal{S}}(y|\mathbf{x}) = P_{\mathcal{T}}(y|\mathbf{x})$~\cite{quinonero2008dataset} where $P_{\mathcal{S}}(\mathbf{x})$ and $P_{\mathcal{T}}(\mathbf{x})$ refer to the probability distribution of the source and target domains, respectively. We consider a realistic scenario where $\mathcal{D}_{\mathcal{T}} \nsubset \mathcal{D}_{\mathcal{S}}$. Let $\mathcal{D}_{\mathcal{T}}^{tr}$ be training data for adaptation and $\mathcal{D}_{\mathcal{T}}^{va}$ be unlabeled validation data, where $\mathcal{D}_{\mathcal{T}}^{tr}, \mathcal{D}_{\mathcal{T}}^{va} \subset \mathcal{D}_{\mathcal{T}}$, and $\mathcal{D}_{\mathcal{T}}^{tr} \cap \mathcal{D}_{\mathcal{T}}^{va} = \varnothing$. We want to estimate the adapted model's performance with unlabeled target validation data $\mathcal{D}_{\mathcal{T}}^{va}$. We are interested in the target \textit{accuracy,}\footnote[1]{We have also tried other metrics, such as the F1 score, and observed similar findings throughout the experiments.} which is the most representative metric of performance. We focus on estimating the \textit{change in the target accuracy} as adaptation proceeds. We use the terms ``performance'' and ``accuracy'' interchangeably throughout the paper. Specifically, the target accuracy at epoch $e \in E=\{0,1,2,...,T\}$ is defined as 
\begin{equation}
    a^{(e)} = \frac{\sum_{(\boldsymbol{x}_{\mathcal{T}}, y_{\mathcal{T}}) \in \mathcal{D}_{\mathcal{T}}^{va}}\mathbb{1}_{[\hat{y}_{\mathcal{T}}^{(e)} = y_{\mathcal{T}}]}}{N^{\mathcal{T}}},
\end{equation}
where $\hat{y}_{\mathcal{T}}^{(e)}$ is a predicted label with an adapted model at epoch~$e$, i.e., ${\hat{f}_{\theta}^{(e)}(\mathbf{x}_{\mathcal{T}})}$, 
and $\mathbb{1}_{[\hat{y}_{\mathcal{T}}^{(e)}=y_{\mathcal{T}}]}$ is one if $\hat{y}_{\mathcal{T}}^{(e)} = y_{\mathcal{T}}$ and zero otherwise.
Accordingly, the \textit{change in the target accuracy} at epoch $e$ is defined as follows:
\begin{equation}
    \Delta a^{(e)} = a^{(e)} - a^{(0)},
\end{equation}
which represents how accuracy changed after adaptation compared against before adaptation. Our objective is to minimize the difference between the change in the ground-truth accuracy and the estimated change in accuracy computed from the unlabeled target validation data $\mathcal{D}_{\mathcal{T}}^{va}$ during the training epochs, which is defined as follows:
\begin{equation}\label{eq:objective}
    \textit{minimize} \sum_{e = 1}^{T} \quad \abs{\Delta  a^{(e)} - \Delta \hat{a}^{(e)}},
\end{equation}
where $\Delta\hat{a}^{(e)}$ is the estimated accuracy change of $f_{\theta}^{(e)}$ through $\mathcal{D}_{\mathcal{T}}^{va}$. \rev{We chose this objective as it is intuitive and one of the most widely used objectives (i.e., L1 distance).} 

\subsection{Theoretical Insight}\label{sec:method:theory}

We tackle the problem of performance estimation with unlabeled data based on \emph{mutual information}~\cite{shannon}. In information theory, mutual information is a measure that quantifies the amount of information gained about a random variable after observing the other random variable. The \emph{infomax principle}~\cite{36} explains that neural networks can be viewed as an information channel and proposes a principle to self-organize such networks by conveying as much information as possible of given data. This principle has been widely applied in deep learning, such as feature learning~\cite{NEURIPS2019_ddf35421} and clustering~\cite{ji2019invariant}. Grounded on this, we explain the performance of the model in the aspect of the information gain between the inputs and the corresponding outputs of a model. We then demonstrate that it can be derived via Monte Carlo estimation. Finally, we investigate the relationship between mutual information and model performance.

Specifically, the mutual information (also known as information gain) between two random variables $X$ (i.e., inputs) and $Y$ (i.e., predictions) is defined as:

\begin{equation}
    I(X, Y) = H(Y) - H(Y|X)
\end{equation}
where $H(\cdot)$ is the entropy~\cite{shannon} of a probability distribution. 
Note that entropy is a quantitative measure of \emph{uncertainty} about a random variable. Mutual information represents the \emph{reduction of the uncertainty} about the variable $Y$ after observing $X$. Mutual information becomes zero when $X$ and $Y$ are independent, i.e., when $H(Y|X)=H(Y)$. In the context of the domain shift problem, high mutual information is interpreted as a high reduction of the uncertainty of the model predictions $Y$, given inputs $X$ from a target domain. On the other hand, low mutual information means a low reduction of the uncertainty of $Y$ for inputs $X$ from the target domain, which indicates that the model does not successfully correlate $X$ with $Y$.

Let $p(y|x)$ denote the probability distribution of the output $y$ from the model $f_{\theta}$ given an input $x$. Mutual information can be expressed with $p(y|x)$:
\begin{multline}\label{eq:mutual_information}
    H(Y) - H(Y|X)
    = \sum_{y} p(y) \mathit{log}p(y) - \sum_{x} p(x) \left(\sum_{y} p(y|x) \mathit{log}p(y|x)\right)\\
    = \sum_{y} \left(\sum_{x}p(x) p(y|x)\right) \mathit{log}\left(\sum_{x}p(x) p(y|x)\right) - \mathop{\mathbb{E}}_{x} \left(\sum_{y} p(y|x) \mathit{log}p(y|x)\right)\\
    = \sum_{y} \left(\mathop{\mathbb{E}}_{x} p(y|x)\right) \mathit{log}\left(\mathop{\mathbb{E}}_{x} p(y|x)\right) - \mathop{\mathbb{E}}_{x} \left(\sum_{y} p(y|x) \mathit{log}p(y|x)\right).
\end{multline}
In practice, $p(y|x)$ can be calculated by applying the softmax function to the output logits of neural networks.

We now show that mutual information can be driven from unlabeled target validation data $\mathcal{D}_{\mathcal{T}}^{va}$ via Monte Carlo approximation:
\begin{multline}\label{eq:approx}
    H(Y) - H(Y|X) = \sum_{y} \left(\mathop{\mathbb{E}}_{x} p(y|x)\right) \mathit{log}\left(\mathop{\mathbb{E}}_{x} p(y|x)\right) - \mathop{\mathbb{E}}_{x} \left(\sum_{y} p(y|x) \mathit{log}p(y|x)\right)\\
    \approx \underbrace{\sum_{y} \left(\frac{1}{N_{\mathcal{T}}}\sum_{x\in \mathcal{D}_{\mathcal{T}}^{va}} p(y|x)\right) \mathit{log}\left(\frac{1}{N_{\mathcal{T}}}\sum_{x\in \mathcal{D}_{\mathcal{T}}^{va}} p(y|x)\right)}_{\text{Global Diversity}} - \underbrace{\frac{1}{N_{\mathcal{T}}}\sum_{x\in \mathcal{D}_{\mathcal{T}}^{va}} \sum_{y} p(y|x) \mathit{log}p(y|x)}_{\text{Individual Uncertainty}}.
\end{multline}
Note that the first term is the entropy of the averaged predictions of the model and the second term is the average of the entropy for each prediction. As a high-level interpretation, the first term becomes high if the predictions are globally diverse, and the second term becomes low when the uncertainties of individual predictions are low. Therefore, mutual information is high when the predictions on the target validation data are both globally diversified and individually certain. Intuitively, an accurate classifier should provide confident class predictions, while the output class distribution should be diverse for class-balanced inputs, and thus have high mutual information~\cite{NIPS1991_a8abb4bb}. This also aligns with the \emph{infomax principle} for neural networks, which prescribes that a classifier should maximize the mutual information between inputs and outputs~\cite{36}.

\begin{figure}[t]
    \centering
    \begin{subfigure}[t]{0.25\linewidth}
        \centering
        \includegraphics[width=0.95\linewidth]{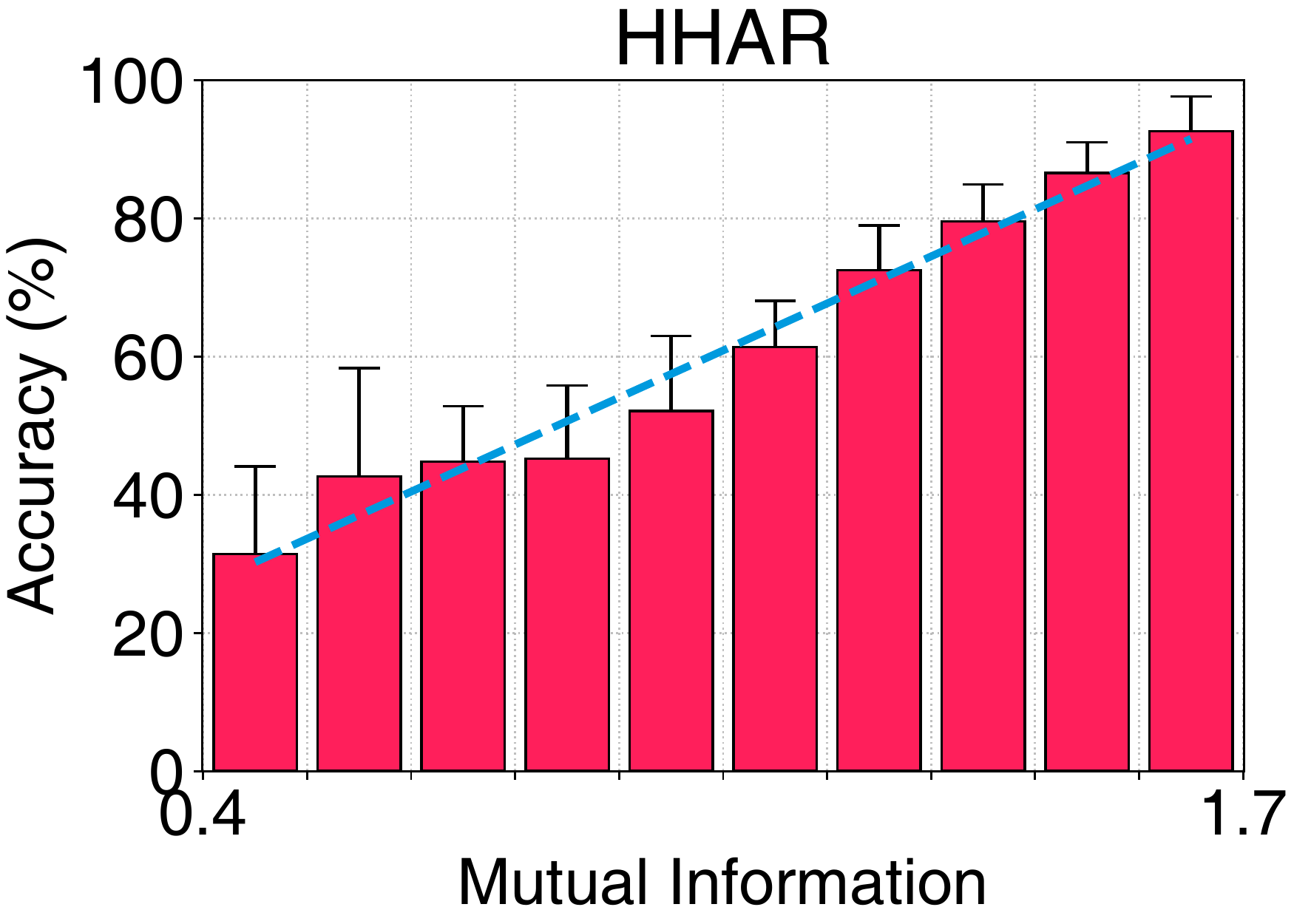}
        \label{fig:correl_mi:hhar}
    \end{subfigure}
    ~
    \begin{subfigure}[t]{0.25\linewidth}
        \centering
        \includegraphics[width=0.95\linewidth]{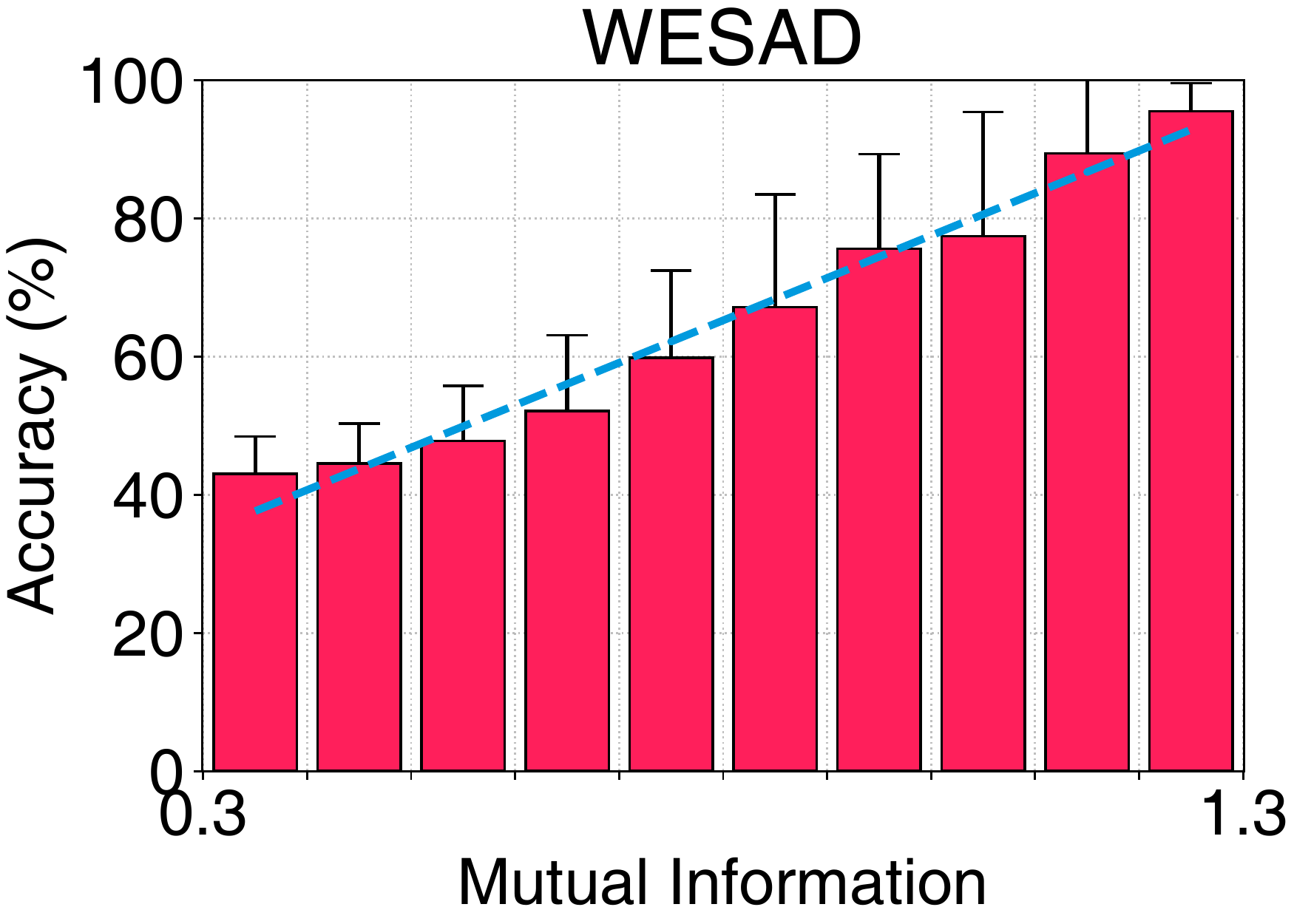}
        \label{fig:correl_mi:wesad}
    \end{subfigure}
    ~
    \begin{subfigure}[t]{0.25\linewidth}
        \centering
        \includegraphics[width=0.95\linewidth]{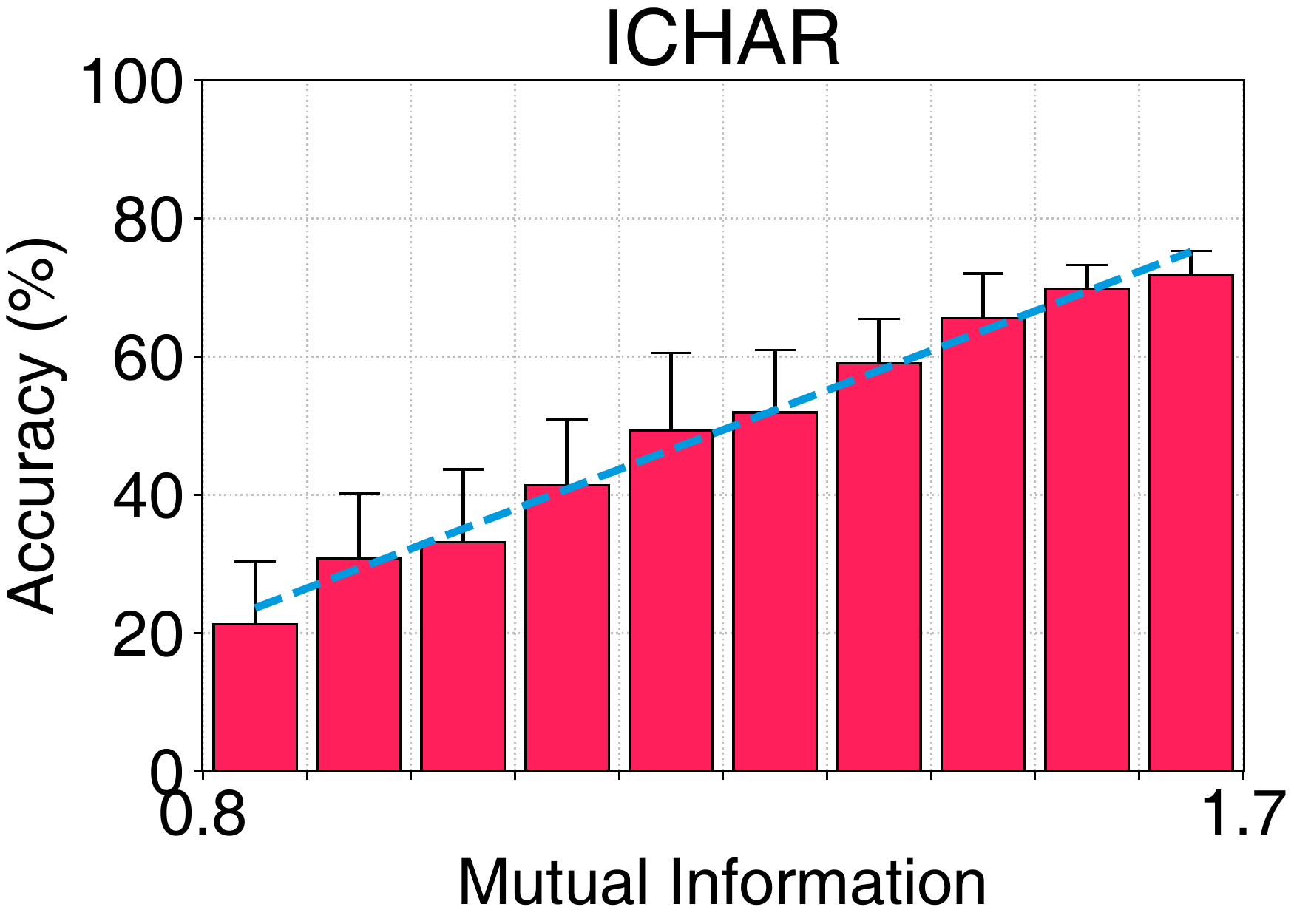}
        \label{fig:correl_mi:act}
    \end{subfigure}
    ~
    \begin{subfigure}[t]{0.25\linewidth}
        \centering
  \includegraphics[width=0.95\linewidth]{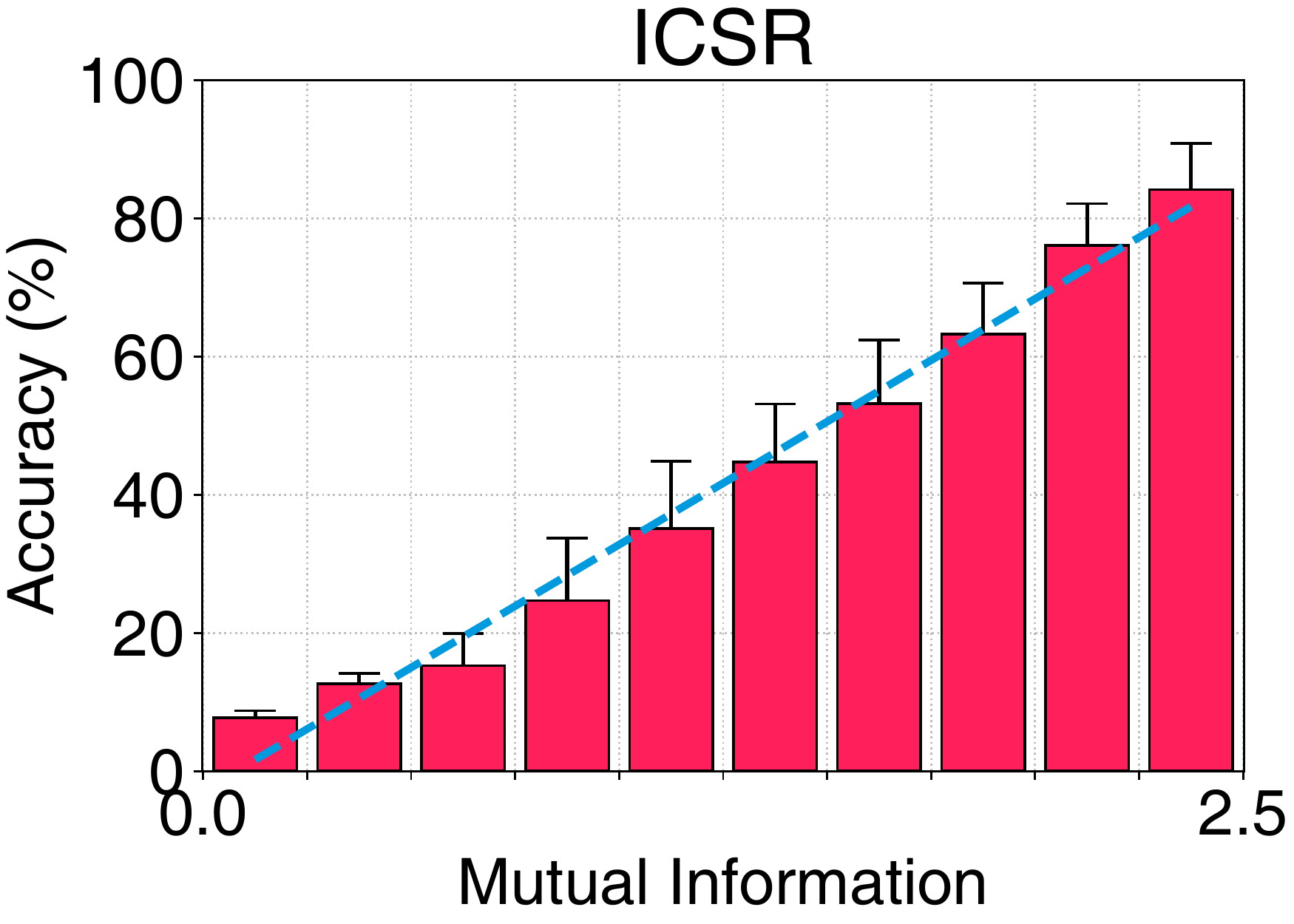}
        \label{fig:correl_mi:spe}
    \end{subfigure}
    \vspace{-0.4cm}
    \caption{Correlation between mutual information and ground-truth accuracy across four datasets.}
    \vspace{-0.3cm}
    \label{fig:correl_mi}
\end{figure}

With the aforementioned theoretical interpretation, we conducted a preliminary study to empirically understand the relationship between mutual information and accuracy. For this study, we used a total of 20k domain adaptation results where we logged both mutual information and ground-truth accuracy using four sensory datasets: Heterogeneity Human Activity Recognition~(HHAR)~\cite{hhar}, Wearable Stress and Affect Detection~(WESAD)~\cite{schmidt2018introducing}, Individual-Condition Human Activity Recognition (ICHAR)~\cite{metasense}, and Individual-Condition Speech Recognition (ICSR)~\cite{metasense} (details in~\cref{sec:experiment:settings}). Figure~\ref{fig:correl_mi} shows the correlation between mutual information and accuracy across four sensory datasets. As shown, mutual information has a positive correlation with performance demonstrating the feasibility of performance estimation via unlabeled target data. Still, the relationship between the accuracy and mutual information is not straightforward, and one might argue that model calibration~\cite{10.5555/3305381.3305518} might be useful in this context. \rev{We discuss the relevance of our method to model calibration in~\cref{sec:discussion:calibration}.} In the following section, we design a performance estimator by addressing the limitations of mutual information.

\subsection{Performance Estimator}\label{sec:method:dapper}

While mutual information provides theoretical insights for performance estimation with unlabeled data, using it directly as a proxy for performance has several practical limitations: (i)~The unit of mutual information denotes the amount of information (e.g., bits with the log base 2). It does not represent the target performance metric of our interest, e.g., accuracy. Therefore, mutual information itself is not an interpretable metric to users. (ii)~As illustrated in Figure~\ref{fig:correl_mi}, the relationship between mutual information and accuracy depends on the given unlabeled data from each dataset; the relationship is not a one-to-one function. Mutual information cannot decode the underlying characteristics of datasets, which makes it different among datasets. (iii)~We are interested in accuracy changes for each iteration of domain adaptation. In such scenarios, previous information (e.g., previously estimated accuracy) could give a hint to performance estimation. Mutual information, however, does not utilize previous information.

To overcome the limitations of our theoretical analysis, we propose to train a domain adaptation performance estimator that gets the outputs of an adapted model on $\mathcal{D}_{\mathcal{T}}^{va}$ and predicts the change in accuracy. The benefits of the proposed performance estimator are three-fold: (i)~machine learning can capture the underlying optimal mapping from mutual information to the desired metric (e.g., accuracy) with sufficient training data and supervised objective, (ii)~dataset-dependent characteristics can be learned by training dataset-specific estimators, and (iii)~machine learning can leverage previous information for better prediction via sequential modeling.

Practical concerns for training such an estimator would be the cost of data collection and computation overhead. \system{} resolves the concerns by generating lots of virtual training data using already collected source data and decoupling the training process from the user side. \system{} utilizes the virtual training data to simulate heterogeneous domain adaptation scenarios and learn to estimate accuracy. Unlike state-of-the-art performance estimation algorithms~\cite{chuang2020estimating, ensrm}, the entire training process of \system{} is conducted on the developer side. Therefore, once deployed to users, only inference is required for estimation.

\begin{figure*}[t]
    \centering
    \includegraphics[width=1\linewidth]{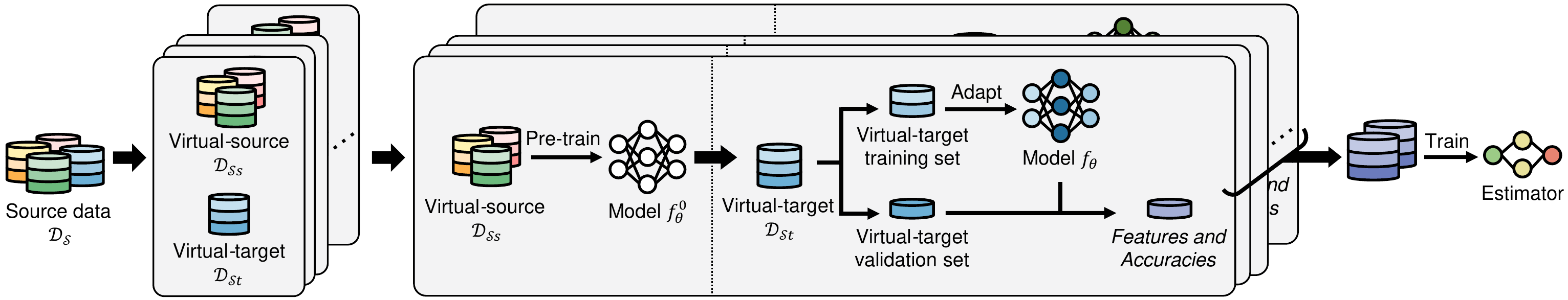}
    \vspace{-0.3cm}
1    \caption{An illustration of the training process of the performance estimator. We generate training data via simulation. The whole process runs on the developer side.}
    \vspace{-0.3cm}
    \label{fig:dapper}
\end{figure*}

\subsubsection{Virtual Training Data Generation}\label{sec:method:dapper:virtual}
We propose a simulation-based virtual training data generation to train the estimator. Our data generation methodology is inspired by the principle of domain randomization, postulating that the disparities between source and target domains could be represented as variations within the source domain~\cite{10.1109/ICRA.2018.8460528, 10.1109/IROS.2017.8202133}. Specifically, we generate training data with the following objectives. First, we create training data only from source data. Collecting target data is not required to train the estimator, which is beneficial in terms of cost and effort. Second, we train the estimator with as diverse data as possible to make the estimator learn various adaptation scenarios.

Figure~\ref{fig:dapper} is an illustration of the training process. The training starts from virtual training data generation with the source data. Note that we do not have access to the target data that is ideal for training the estimator. We instead \textit{simulate} adaptation with the source data to generate the training data for the estimator. We assume the model developer collected or obtained the source datasets $\mathcal{D}_{\mathcal{S}}$ from multiple domains, and we randomly split the source data into non-overlapping virtual-source domains $\mathcal{D}_{\mathcal{S}_{s}}$ and virtual-target domains $\mathcal{D}_{\mathcal{S}_{t}}$. With the virtual-source and virtual-target, we generate the training data by \textit{simulating} adaptation; the model is pre-trained with the virtual-source data, and the adaptation algorithm is applied to the pre-trained model to adapt to the virtual-target. We randomly sample the virtual-target training and validation data to simulate diverse adaptation scenarios. While adapting to the virtual-target data, features are calculated for every epoch from the unlabeled virtual-target validation data. We also log accuracy as the ground truth. \rev{This process is repeated 10,000 times to generate enough training data, which takes around one day with a server equipped with eight NVIDIA GeForce RTX 3090 GPUs. We then train the estimator with dataset-specific virtual training data.} This way, we expect the estimator can be trained from myriad combinations of virtual-source and virtual-target domains, and, more importantly, learn the dataset-specific domain-invariant relationship between the features and ground-truth accuracy, which we explain in detail in the following section.

\subsubsection{Training}\label{sec:method:dapper:training}
We train the performance estimator $g_{\phi}$ of \system{} with the data generated from the source data (Figure~\ref{fig:dapper}). To train an accurate and lightweight estimator, we (i)~use computationally light features as the inputs for the estimator and (ii)~utilize an LSTM architecture for sequential modeling. 

\vspace{0.1cm}
For features, we use global diversity and individual uncertainty in Equation~\eqref{eq:approx} to learn an optimal relationship between them. Given unlabeled target validation data $\mathcal{D}_{\mathcal{T}}^{va}$, global diversity (GD) is defined as:
\begin{equation}
    \mathit{GD}(\mathcal{D}_{\mathcal{T}}^{va}) = \sum_{y} \left(\frac{1}{N_{\mathcal{T}}}\sum_{x\in \mathcal{D}_{\mathcal{T}}^{va}} p(y|x)\right) \mathit{log}\left(\frac{1}{N_{\mathcal{T}}}\sum_{x\in \mathcal{D}_{\mathcal{T}}^{va}} p(y|x)\right).
\end{equation}
Similarly, the definition of individual uncertainty (IU) is:
\begin{equation}
    \mathit{IU}(\mathcal{D}_{\mathcal{T}}^{va}) = \frac{1}{N_{\mathcal{T}}}\sum_{x\in \mathcal{D}_{\mathcal{T}}^{va}} \sum_{y} p(y|x) \mathit{log}p(y|x).
\end{equation}
In addition, we consider prediction distribution (PD) as a feature, which is defined as:
\begin{equation}
    \mathit{PD}(\mathcal{D}_{\mathcal{T}}^{va}) = \frac{1}{N_{\mathcal{T}}}\sum_{x\in \mathcal{D}_{\mathcal{T}}^{va}} p(y|x).
\end{equation}

Note that GD and IU are single values while PD contains $K$-dimensional information where $K$ is the number of classes. As GD and IU are single values (i.e., encoded values) summed over all $y$s, relying on only those features might lose important data-specific characteristics that can be observed in class information $y$. PD reveals class-wise information (i.e., decoded values) that is beneficial in capturing data characteristics and thus improves performance estimation. Our features are extremely lightweight as opposed to using raw sensory inputs as features, e.g., one-second accelerometer input corresponding to 400$\times$3 values under 400~Hz sampling rate.

In addition, we use the difference from the previous values as additional input features, i.e.,
\begin{equation}
    \Delta \mathit{Feat}_{e} = \mathit{Feat}_{e} - \mathit{Feat}_{e-1}, \quad
    \text{where} \quad \mathit{Feat}_{e} \in \mathcal{F}_{e}, \quad \text{and} \quad \mathcal{F}_{e} = \{\mathit{GD}_{e}, \mathit{IU}_{e}, \mathit{PD}_{e}\},
\end{equation}

where $e$ is the training epoch. By providing the value differences as additional features, not only can the estimator consider the current features, but it also leverages the changes in the features from the previous model. This explicit use of the absolute values (current) and the relative values (difference) of features helps the estimator better correlate the given information with the change in performance. We evaluate the effectiveness of our features via an ablative study in~\cref{sec:experiment:ablation}.

We notice that domain adaptation is a sequential process, i.e., the model is likely to be updated periodically with multiple training epochs. In such scenarios, previous information about the target model (e.g., previous features) could provide hints about the current accuracy estimation. To leverage such previous ``states'' of the model, we use Long Short-Term Memory (LSTM)~~\cite{6795963} as the backbone network. LSTM is widely used for sequential modeling as it is simple and robust to the vanishing gradient problem in taking long sequences as inputs. Specifically, we use two stacks of LSTM layers with 200 hidden dimensions, followed by two fully connected layers with one ReLU activation. We use an L1 loss to minimize the difference between the predicted change in accuracy and the ground-truth change in accuracy (Equation~\eqref{eq:objective}). For optimization, we adopt the Adam optimizer~\cite{kingma2014adam} with a learning rate of $10^{-5}$. The resulting model size is affordable for mobile devices (1.9~MB) and incurs negligible computation overhead. We conduct a comprehensive analysis for the computation overhead in~\cref{sec:experiment:overhead}.

\subsubsection{Inference}\label{sec:method:dapper:inference}
Given unlabeled target validation data $\mathcal{D}_{\mathcal{T}}^{va}$ and a target model $f_{\theta}^{(e)}$, \system{} calculates GD, IU, PD, and the differences of each of them. Those features are fed into the performance estimator $g_{\phi}$, and the estimator predicts the accuracy change from the initial model $f_{\theta}^{(0)}$, i.e.,
\begin{equation}
    \Delta \hat{a}^{(e)} = g_{\phi}(\mathit{GD}_{e}, \mathit{IU}_{e}, \mathit{PD}_{e}, \Delta\mathit{GD}_{e}, \Delta\mathit{IU}_{e}, \Delta\mathit{PD}_{e}).
\end{equation}
Note that the performance estimator $g_{\phi}$ is pre-trained by the developer, and only inferences are made from the estimator after deployment. We conduct a detailed computation overhead analysis of \system{} and comparison with baselines in~\cref{sec:experiment:overhead}.
\section{Experiment Settings}~\label{sec:experiment:settings}
\subsection{Datasets and Preprocessing}\label{sec:experiment:settings:dataset}
We evaluate our method with the following four real-world mobile sensing datasets.
\subsubsection{HHAR}
HHAR~\cite{hhar} contains six activity classes of IMU data, obtained from nine users and 12 devices (eight smartphones and four smartwatches). Smartphones were kept in a tight pouch around their waist and smartwatches were on their arms. Sampling rates and sensor specifications vary by device. After removing two users whose Samsung Galaxy Gear data were missing, we created domains with seven users and five devices. We randomly selected four users and three devices and grouped them as sources (12 domains), and used the remaining devices and users as targets (six domains).

\subsubsection{WESAD}
WESAD~\cite{schmidt2018introducing} is a multi-modal dataset for wearable stress and affect detection with four classes. It was collected from 15 users equipped with the same medical devices around their chest and wrist. WESAD includes diverse sensing modalities: three-axis acceleration, electrocardiogram, electrodermal activity, electromyogram, respiration, body temperature, and blood volume pulse. Following the previous study~\cite{9361223}, we down-sampled each modality to 4~Hz, which is the lowest sampling rate. Among 15 domains, we randomly chose 10 domains as sources and the others as targets. 

\subsubsection{ICHAR}
ICHAR~\cite{metasense} contains nine classes of IMU data collected from 10 pairs of different users and devices. Each participant was carrying a unique smartphone or smartwatch and there was no restriction on how to hold the device. This makes the domain adaptation task in ICHAR more challenging than HHAR which had clear instructions on how to carry devices. Among 10 individual conditions, i.e., domains, we randomly chose seven domains as sources and the remaining three as targets.

\subsubsection{ICSR}
ICSR~\cite{metasense} is a speech recognition dataset collected under the same condition as ICHAR. During data collection, each subject held one unique device in a preferred way and spoke out 14 different words with their own styles (loudness, pitch, etc.). Among 10 resulting domains, we randomly chose seven domains as sources and the remaining three as targets.

\subsection{Adaptation Methods}\label{sec:experiment:settings:adaptation}
We introduce domain adaptation methods upon which \system{} estimates performance. We consider a common multi-source DA scenario: for each target domain, we pre-trained a base model with the source domains, and the base model was adapted to the target domain with a DA algorithm. We experimented with four common adaptation methods as follows.

\subsubsection{Fine-tuning}\label{sec:experiment:settings:adaptation:sda}
Fine-tuning utilizes labeled data from a target domain for adaptation. We used a well-known fine-tuning method where a pre-trained model updated its parameters with a few labeled data from the target domain. Specifically, we randomly chose a target domain first and selected 1$\sim$50 random labeled data from the target with which we performed adaptation. Here, we used a separate unlabeled dataset that was not used for adaptation to estimate performance. This process was repeated 50 times for each dataset, and we reported the average. 

\subsubsection{DANN}
Domain-Adversarial Training of Neural Network~(DANN)~\cite{10.5555/2946645.2946704} is a widely used unsupervised domain adaptation (UDA) algorithm that adapts to a target domain with labeled source and unlabeled target data. Similar to fine-tuning, we chose a target domain randomly, selected 50$\sim$500 random unlabeled data from the target, performed adaptation, and estimated the performance with separate unlabeled data, which was repeated 50 times for each dataset. We provided more samples in DANN than fine-tuning, considering that unlabeled data are easier to collect.

\subsubsection{CDAN}
Conditional Domain Adversarial Network (CDAN)~\cite{long2018conditional} improves the adversarial training of DANN by jointly utilizing feature representations and classifier predictions with labeled source and unlabeled target data. We adopted the same experiment setting as DANN, where we randomly chose a target domain, selected 50$\sim$500 random unlabeled data from the target, performed adaptation, and estimated performance with separate unlabeled data, which was repeated 50 times for each dataset.

\subsubsection{SHOT}
Source HypOthesis Transfer (SHOT)~\cite{shot} is a recent UDA algorithm that does not require source data in adaptation. We adopt SHOT as it has a practical setting since the source data might not be available to the target users due to privacy and intellectual property concerns. Same as DANN, we chose a target domain randomly, selected 50$\sim$500 random unlabeled data from the target, performed adaptation, and estimated performance with a separate unlabeled data, which was repeated 50 times for each dataset. 

\subsubsection{Implementation}
For the backbone networks, we adopted 1D convolutional neural networks (CNN)~\cite{NIPS2012_c399862d} followed by fully-connected layers, which are commonly used in mobile sensing~\cite{metasense, systematic, scaling, 10.1145/3328932}. We used Rectified Linear Unit (ReLU)~\cite{10.5555/3104322.3104425} for activation function and Batch Normalization layer~\cite{10.5555/3045118.3045167} after each CNN layer for regularization and fast convergence. We used L2-regularization~\cite{10.1145/1015330.1015435} as a regularization technique. We trained the model with Adam optimizer~\cite{kingma2014adam}. We used a fixed learning rate of 0.001. We used a conventional standardization for preprocessing the datasets. We used the PyTorch framework~\cite{paszke2017automatic} for implementation and trained the model with NVIDIA TITAN Xp GPUs. We ran each training for 100 epochs in both adaptation scenarios.

\subsection{Baselines}\label{sec:experiment:settings:baselines}
We compare \system{} with five baselines, including the state-of-the-art performance estimation algorithms~\cite{chuang2020estimating, ensrm}. We note that in the context of mobile sensing, no previous work has proposed performance estimation utilizing unlabeled data. Most studies trained until a fixed number of epochs~\cite{digging2020, xhar2020, generalization_fitness2021}, while several studies selected the best models on the labeled target validation data~\cite{metasense, multi_source2020}.

\subsubsection{TgtLabel}
TgtLabel computes accuracy with labeled target validation data. 
This performance validation is often used in domain adaptation research~\cite{dou2019domain, contrastive2019, conditional2018, transferable2019, Tang_2020_CVPR, cui2020gvb, dada, xu2020adversarial, metasense, multi_source2020}. We use TgtLabel as the ground-truth performance.

\subsubsection{FixedEpoch}
Training until a fixed number of epochs (FixedEpoch) is a widely used model selection method~\cite{shot, 10.5555/3045118.3045244, 10.1007/978-3-319-49409-8_35, shu2018a, NEURIPS2018_99607461, 10.1007/978-3-030-01225-0_28, 8953760, pmlr-v80-xie18c, pmlr-v70-long17a, Haoran_2020_ECCV, digging2020, xhar2020, generalization_fitness2021}. Note that as FixedEpoch always chooses the last epoch's model without any validation data, it does not have performance validation.

\subsubsection{SrcLabel}
Validating performance with labeled source data (SrcLabel) is also a popular method~\cite{8578985, Li2018MLDG, Li_2019_ICCV, 10.1007/978-3-319-46493-0_36} as source data are more likely labeled with a larger amount, compared with target data.
SrcLabel computes accuracy from a hold-out labeled source data to estimate the target performance. In order not to affect the performance of the pre-trained model, the hold-out source data for validation were never used in the training process for other baselines as well.

\subsubsection{SoftmaxScore}\label{sec:experiment:settings:softmax}
Using the confidence score from the last softmax layer~\cite{elsahar-galle-2019-annotate} might be the simplest approach utilizing target unlabeled data for performance estimation. This method is hence commonly adopted as a baseline in related studies~\cite{chuang2020estimating, ensrm}. Specifically, it estimates the accuracy of the target domain by averaging the softmax confidence scores computed from unlabeled target validation data.

\subsubsection{DIR} A recent study utilizes Domain-Invariant Representations (DIR) for estimating generalization under domain shifts~\cite{chuang2020estimating}. This study learns an accurate ``check'' model with domain-invariant representations as a proxy for estimating performance. Specifically, it pre-trains a check model with both labeled source data and unlabeled target data via DANN~\cite{10.5555/2946645.2946704}. After each adaptation, to estimate performance, it updates the check model that maximizes disagreement with the adapted model. An estimated proxy risk is the maximum disagreement between the check model and the adapted model. Therefore, this algorithm requires updating the check model for every update of the adapted model. We used this proxy risk to calculate an estimated accuracy. For the check model, we used the same model as the adaptation model and used the unlabeled target validation data to train. We referred to the official implementation
and the check model was updated for 20 epochs.


\subsubsection{EnsRM} Ensemble via Representation Matching (EnsRM)~\cite{ensrm} is the state-of-the-art performance estimation algorithm that utilizes ensemble models and self-training to estimate target accuracy. Similar to DIR~\cite{chuang2020estimating}, EnsRM assumes that an accurate check model is pre-trained via DANN~\cite{10.5555/2946645.2946704}. For performance estimation, EnsRM further fine-tunes the check model to the unlabeled target validation data for multiple epochs and stores the models after each epoch, which jointly become an ensemble of check models. These ensemble models are utilized to identify misclassified target points where the ensemble's class predictions disagree with those of the adapted model, and the misclassified target points are added to training samples. This procedure is repeated for multiple rounds with self-training, and the disagreement ratio from the last round becomes an estimated error of the model. Following the authors' implementation,
we updated the check model for five epochs and five rounds.

\subsubsection{Comparison with Baselines}

\begin{table}[t]
    \centering
    \caption{Comparison of the baselines and \system{}.}
    \vspace{-0.3cm}
    
    \begin{tabular}{
    lcccc
    }
    \Xhline{2\arrayrulewidth}
                        & \textbf{Src data} & \textbf{Tgt data} & \textbf{Extra Training} & \textbf{Validation}  \\\hline
    \textbf{TgtLabel (oracle)}   & No                    & Labeled              & No       & Yes                     \\ 
    
    \textbf{FixedEpoch} & No                    & No                    & No       & No                     \\
    \textbf{SrcLabel}   & Labeled              & No                    & No       & Yes                     \\
    \textbf{SoftmaxScore}        & No              & Unlabeled            & No     & Yes                       \\ 
    \textbf{DIR}        & Labeled              & Unlabeled            & Yes     & Yes                       \\ 
    \textbf{EnsRM}        & Labeled              & Unlabeled            & Yes     & Yes                       \\ 
    \textbf{\system{} (ours)}    & No                    & Unlabeled            & No          & Yes                  \\ 
    
    \Xhline{2\arrayrulewidth}
    \end{tabular}
    \vspace{-0.3cm}
    \label{tab:baselines}
\end{table}

Table~\ref{tab:baselines} compares the baselines and \system{}. \rev{``Src data'' means whether it requires hold-out source data during performance estimation while ``Tgt data'' means whether it requires target data during estimation.} ``Labeled'' means the requirement of labeled data, ``Unlabeled'' means the requirement of unlabeled data, and ``No'' means no requirement of data for that category. ``Extra Training'' is whether it requires \textit{additional} training for every updated model. 
Unlike DIR and EnsRM, \system{} requires training only once before deployment, without additional training for every adapted model. ``Validation'' is whether the method includes performance validation. FixedEpoch does not have performance validation.

\subsection{Performance Estimation Setup}

For performance validation (or estimation), we randomly selected 500 samples throughout experiments unless specified. The validation samples were consistent for each dataset to ensure a fair evaluation. For instance, all baselines that require target data used the same 500 samples from a target for validation, but those samples were labeled for TgtLabel and unlabeled for DIR, EnsRM, and \system{}. SrcLabel used 500 samples from the hold-out source data. For each dataset and adaptation method pair, the result was averaged over 50 training episodes as explained in \cref{sec:experiment:settings:adaptation}.

We use similarity calculated from an L1 distance as a metric to compare with baselines. L1 distance 
is a metric that calculates the distance between two $N$ dimensional vectors. We regard the true accuracy (TgtLabel) as the oracle and calculate the L1 distance between the ground truth and the estimations. We subtract the averaged L1 distance from 1 to compute similarity. Specifically, we define similarity as follows: 

\begin{equation}
    \textit{Similarity} = 1 - \frac{1}{\abs{E}} \sum_{e \in E} \quad \abs{\Delta  a^{(e)} - \Delta \hat{a}^{(e)}},
\end{equation}
which ranges from 0 to 1 (0$\sim$100\%). Therefore, the higher the similarity, the better the estimation.

\section{Evaluation}\label{sec:experiment}

\begin{table}[t]
\centering
\caption{Average similarities (\%) of the baselines and \system{} under four domain adaptation methods (fine-tuning, DANN, CDAN, and SHOT), across four datasets (HHAR, WESAD, ICHAR, and ICSR). D1$\sim$D$n$ represent target domains. Bold type numbers indicate those of the highest average similarity.}
\label{tab:overall_result}
\resizebox{\columnwidth}{!}{%
\begin{tabular}{
lrrrrrrrrrrrrrrrrrr
}
    \Xhline{2\arrayrulewidth}
    \addlinespace[0.12cm] 
 \multicolumn{1}{c}{\multirow{2}{*}{\textbf{Method}}} & \multicolumn{6}{c}{\textbf{HHAR}} & \multicolumn{5}{c}{\textbf{WESAD}} & \multicolumn{3}{c}{\textbf{ICHAR}} & \multicolumn{3}{c}{\textbf{ICSR}} & \multicolumn{1}{c}{\multirow{2}{*}{\textbf{Avg.}}} \\
 \cmidrule(lr){2-7} \cmidrule(lr){8-12} \cmidrule(lr){13-15} \cmidrule(lr){16-18}
 & \multicolumn{1}{c}{D1} & \multicolumn{1}{c}{D2} & \multicolumn{1}{c}{D3} & \multicolumn{1}{c}{D4} & \multicolumn{1}{c}{D5} & \multicolumn{1}{c}{D6} & \multicolumn{1}{c}{D1} & \multicolumn{1}{c}{D2} & \multicolumn{1}{c}{D3} & \multicolumn{1}{c}{D4} & \multicolumn{1}{c}{D5} & \multicolumn{1}{c}{D1} & \multicolumn{1}{c}{D2} & \multicolumn{1}{c}{D3} & \multicolumn{1}{c}{D1} & \multicolumn{1}{c}{D2} & \multicolumn{1}{c}{D3} & \multicolumn{1}{c}{} \\

    \addlinespace[0.02cm] 
    \hline
    \addlinespace[0.2cm] 
    \multicolumn{19}{l}{\textit{Adaptation algorithm: Fine-tuning}}\\
    \hline
    \addlinespace[0.04cm] 

TgtLabel & 100.0 & 100.0 & 100.0 & 100.0 & 100.0 & 100.0 & 100.0 & 100.0 & 100.0 & 100.0 & 100.0 & 100.0 & 100.0 & 100.0 & 100.0 & 100.0 & 100.0 & 100.0 \\
FixedEpoch & N/A & N/A & N/A & N/A & N/A & N/A & N/A & N/A & N/A & N/A & N/A & N/A & N/A & N/A & N/A & N/A & N/A & N/A \\
SrcLabel & 78.5 & 72.3 & 77.6 & 71.4 & 83.5 & 71.7 & 73.0 & 60.2 & 78.9 & 82.8 & 65.7 & 73.9 & 71.5 & 67.8 & 84.2 & 87.0 & 81.8 & 75.4 \\
SoftmaxScore & 85.2 & 89.5 & 88.8 & 88.7 & 91.9 & 87.4 & 82.7 & 86.7 & 82.1 & 74.9 & 90.9 & 82.4 & 75.1 & 76.3 & 70.2 & 78.9 & 85.4 & 83.4 \\
DIR & 70.1 & 69.6 & 62.4 & 64.7 & 70.4 & 66.8 & 68.7 & 28.7 & 52.0 & 45.4 & 57.3 & 64.0 & 61.5 & 64.0 & 71.7 & 69.5 & 71.6 & 62.3 \\
EnsRM & 44.2 & 41.3 & 39.8 & 38.3 & 39.8 & 43.1 & 59.4 & 48.5 & 47.0 & 57.8 & 47.8 & 53.6 & 59.4 & 55.6 & 62.8 & 44.8 & 40.8 & 48.5 \\
DAPPER (ours) & 94.3 & 95.9 & 93.8 & 97.0 & 96.3 & 95.8 & 94.5 & 94.9 & 90.6 & 77.0 & 89.1 & 94.5 & 95.9 & 96.2 & 89.4 & 93.1 & 89.8 & \textbf{92.8} \\

    \addlinespace[0.02cm] 
    \hline
    \addlinespace[0.2cm] 
    \multicolumn{19}{l}{\textit{Adaptation algorithm: DANN}}\\
    \hline
    \addlinespace[0.04cm]

TgtLabel & 100.0 & 100.0 & 100.0 & 100.0 & 100.0 & 100.0 & 100.0 & 100.0 & 100.0 & 100.0 & 100.0 & 100.0 & 100.0 & 100.0 & 100.0 & 100.0 & 100.0 & 100.0 \\
FixedEpoch & N/A & N/A & N/A & N/A & N/A & N/A & N/A & N/A & N/A & N/A & N/A & N/A & N/A & N/A & N/A & N/A & N/A & N/A \\
SrcLabel & 85.7 & 84.3 & 92.1 & 91.1 & 90.7 & 77.9 & 70.4 & 64.0 & 73.2 & 41.3 & 93.6 & 66.0 & 64.1 & 60.6 & 64.7 & 93.3 & 84.0 & 76.3 \\
SoftmaxScore & 89.2 & 84.1 & 94.8 & 93.4 & 96.1 & 85.1 & 65.4 & 49.8 & 83.7 & 42.6 & 86.3 & 76.1 & 76.1 & 73.6 & 75.8 & 95.4 & 88.3 & 79.8 \\
DIR & 82.9 & 87.4 & 54.1 & 66.4 & 63.6 & 76.9 & 93.9 & 82.9 & 70.8 & 81.2 & 70.2 & 92.8 & 83.1 & 87.7 & 74.5 & 63.9 & 70.6 & 76.6 \\
EnsRM & 42.6 & 44.3 & 35.6 & 37.3 & 36.3 & 49.4 & 87.8 & 84.8 & 61.8 & 85.4 & 54.9 & 66.7 & 67.3 & 74.2 & 56.4 & 27.8 & 36.9 & 55.9 \\
DAPPER (ours) & 93.4 & 94.8 & 95.1 & 89.5 & 93.5 & 96.7 & 92.8 & 92.6 & 84.9 & 78.1 & 90.0 & 95.7 & 89.6 & 93.4 & 95.2 & 91.7 & 97.5 & \textbf{92.0} \\
    \addlinespace[0.02cm] 
    \hline
    \addlinespace[0.2cm] 
    \multicolumn{19}{l}{\textit{Adaptation algorithm: CDAN}}\\
    \hline
    \addlinespace[0.04cm] 
    
TgtLabel & 100.0 & 100.0 & 100.0 & 100.0 & 100.0 & 100.0 & 100.0 & 100.0 & 100.0 & 100.0 & 100.0 & 100.0 & 100.0 & 100.0 & 100.0 & 100.0 & 100.0 & 100.0 \\
FixedEpoch & N/A & N/A & N/A & N/A & N/A & N/A & N/A & N/A & N/A & N/A & N/A & N/A & N/A & N/A & N/A & N/A & N/A & N/A \\
SrcLabel & 86.8 & 86.8 & 92.8 & 90.2 & 87.3 & 81.1 & 86.6 & 68.1 & 95.0 & 57.9 & 94.9 & 77.3 & 71.3 & 75.8 & 61.3 & 93.8 & 86.8 & 82.0 \\
SoftmaxScore & 79.6 & 78.1 & 85.2 & 81.9 & 82.3 & 74.2 & 73.6 & 56.1 & 88.1 & 46.5 & 84.9 & 55.2 & 50.8 & 55.4 & 53.7 & 84.3 & 76.6 & 71.0 \\
DIR & 76.4 & 81.4 & 62.3 & 79.3 & 71.0 & 77.3 & 72.1 & 67.9 & 46.9 & 69.8 & 58.0 & 92.0 & 74.6 & 79.1 & 74.9 & 63.5 & 67.2 & 71.4 \\
EnsRM & 41.1 & 42.8 & 35.8 & 39.7 & 41.3 & 47.4 & 77.2 & 91.5 & 46.4 & 69.9 & 52.1 & 64.0 & 68.5 & 64.4 & 60.4 & 25.3 & 35.6 & 53.1 \\
DAPPER (ours) & 92.3 & 97.4 & 96.1 & 89.1 & 96.1 & 96.5 & 94.9 & 90.8 & 89.0 & 90.8 & 91.2 & 95.3 & 94.7 & 90.9 & 96.6 & 89.1 & 95.3 & \textbf{93.3}\\
    \addlinespace[0.02cm] 
    \hline
    \addlinespace[0.2cm] 
    \multicolumn{19}{l}{\textit{Adaptation algorithm: SHOT}}\\
    \hline
    \addlinespace[0.04cm] 
    
TgtLabel & 100.0 & 100.0 & 100.0 & 100.0 & 100.0 & 100.0 & 100.0 & 100.0 & 100.0 & 100.0 & 100.0 & 100.0 & 100.0 & 100.0 & 100.0 & 100.0 & 100.0 & 100.0 \\
FixedEpoch & N/A & N/A & N/A & N/A & N/A & N/A & N/A & N/A & N/A & N/A & N/A & N/A & N/A & N/A & N/A & N/A & N/A & N/A \\
SrcLabel & 89.5 & 82.2 & 75.9 & 80.7 & 91.6 & 86.3 & 95.8 & 83.3 & 91.1 & 76.5 & 92.3 & 92.9 & 79.0 & 90.0 & 92.9 & 79.0 & 90.0 & 86.4 \\
SoftmaxScore & 75.1 & 82.9 & 91.4 & 74.9 & 74.0 & 76.8 & 47.7 & 33.0 & 50.4 & 34.6 & 65.9 & 54.5 & 89.5 & 74.7 & 54.5 & 89.5 & 74.7 & 67.3 \\
DIR & 75.5 & 79.4 & 58.3 & 67.6 & 78.9 & 72.9 & 79.2 & 77.5 & 72.5 & 75.5 & 68.0 & 79.9 & 65.0 & 75.5 & 79.9 & 65.0 & 75.5 & 73.3 \\
EnsRM & 48.6 & 41.6 & 28.6 & 43.1 & 50.1 & 47.8 & 70.0 & 82.5 & 72.8 & 86.3 & 60.8 & 71.9 & 29.6 & 47.3 & 71.9 & 29.6 & 47.3 & 54.7 \\
DAPPER (ours) & 95.6 & 90.2 & 96.9 & 90.7 & 91.2 & 96.2 & 91.9 & 92.3 & 92.4 & 91.4 & 90.4 & 96.2 & 96.2 & 88.6 & 96.2 & 96.2 & 88.6 & \textbf{93.0}\\

    \Xhline{2\arrayrulewidth}
\end{tabular}%
}
\end{table}

\subsection{Overall Results}\label{sec:experiment:overall}

Table~\ref{tab:overall_result} is the overall result that shows how effective \system{} is compared with the baselines. The best values among different methods are highlighted in bold, except for TgtLabel. The similarity in TgtLabel is 100, as it is the ground truth. Measuring similarity is not available in FixedEpoch, and thus marked as ``N/A.'' For each adaptation method, we specify similarities per each domain (D1$\sim$D$n$) and the average of them.

While both TgtLabel and SrcLabel use labeled data for performance validation, the estimation by SrcLabel is often inaccurate as it assumes that the performance is generalizable via hold-out source data without any information from the target. \rev{Using the softmax score for performance estimation (SoftmaxScore) is not reliable due to the well-known incorrect calibration issue~\cite{10.5555/3305381.3305518}.} We also observe that the performance estimation by DIR is far below \system{}. This is due to its assumption that an accurate check model is available~\cite{chuang2020estimating}. We found that this assumption often does not hold in mobile sensing, where a high-quality check model is not always assured. Note that DIR uses DANN for training the check model for each target domain. Unlike traditional vision domains with thousands of target data and well-defined domains, we discovered that the check model's errors were often high in mobile sensing data, which aligns with our observation of performance dynamics in Figure~\ref{fig:motivation1}. Note that EnsRM performs generally worse than DIR in our experiments. We attribute this to its assumptions that (i)~an ensemble of check models has small errors on points that the target model is correct and (ii)~the ensemble models have a proper covariance with the target model, i.e., ``the covariance should not be negative or too positive'', as noted by the authors~\cite{ensrm}. These assumptions could fail when there is no guarantee for the performance of the ensemble models or the correlation between the ensemble models and the target model, e.g., due to high variability of performance in heterogeneous mobile sensing.

Overall, \system{} outperformed the baselines consistently across four datasets. On average, \system{} \rev{achieved 12.8\% similarity improvement over SrcLabel, 17.5\% improvement over SoftmaxScore, 21.9\% improvement over DIR, and 39.8\%} improvement over the state-of-the-art method, EnsRM. This highlights the effectiveness of our proposed features coupled with our training strategies. We believe the results indicate that \system{} is the most appropriate performance estimation method for heterogeneous mobile sensing. \rev{
We however found some challenging cases where DAPPER showed low similarities (e.g., WESAD D4 in Fine-tuning). These cases have unique patterns that are not well generalized with our virtual training data, which means that domain randomization might not be very effective in those cases. There might be several ways to overcome this, such as generating more virtual training data and increasing the number of source data.
}

\subsection{Qualitative Analysis}~\label{sec:experiment:qualitative} 
\begin{figure*}[t]
\captionsetup[subfigure]{justification=centering} 
    \centering
    \begin{subfigure}[t]{0.95\textwidth}
        \centering
        \begin{subfigure}[t]{0.25\textwidth}
            \centering
            \includegraphics[width=0.95\linewidth]{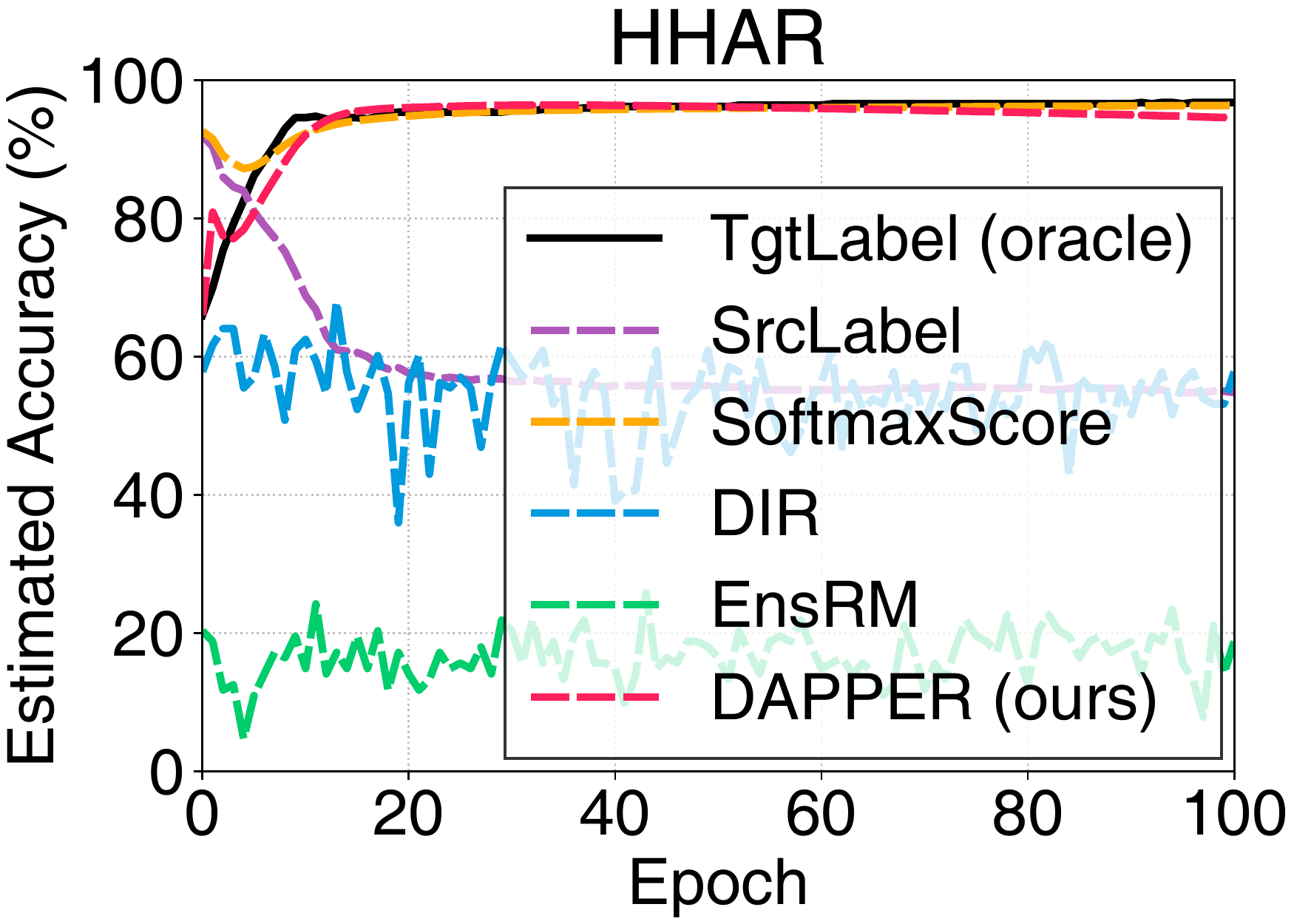}
        \end{subfigure}
        ~
        \begin{subfigure}[t]{0.25\textwidth}
            \centering
            \includegraphics[width=0.95\linewidth]{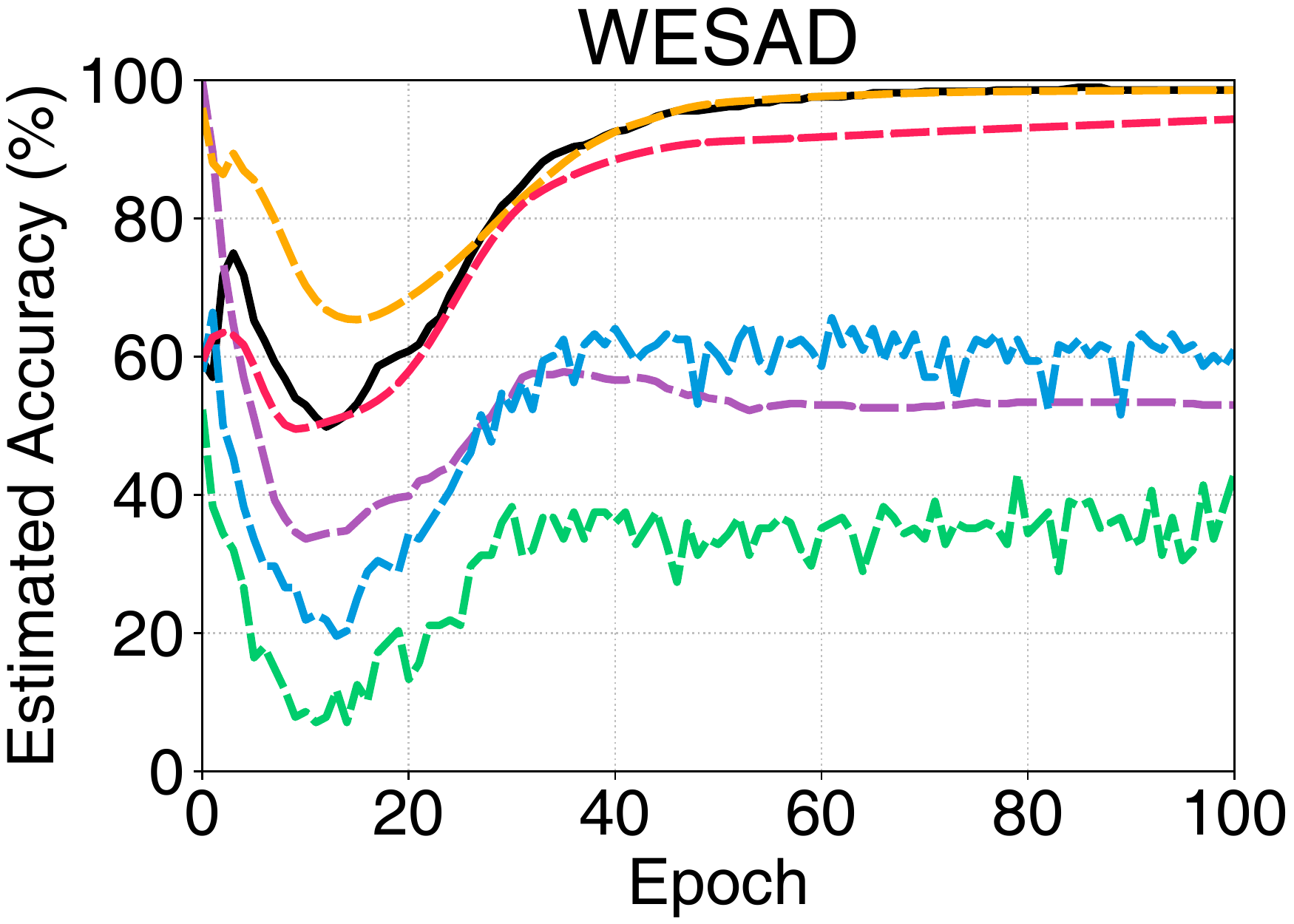}
        \end{subfigure}
        ~
        \begin{subfigure}[t]{0.25\textwidth}
            \centering
            \includegraphics[width=0.95\linewidth]{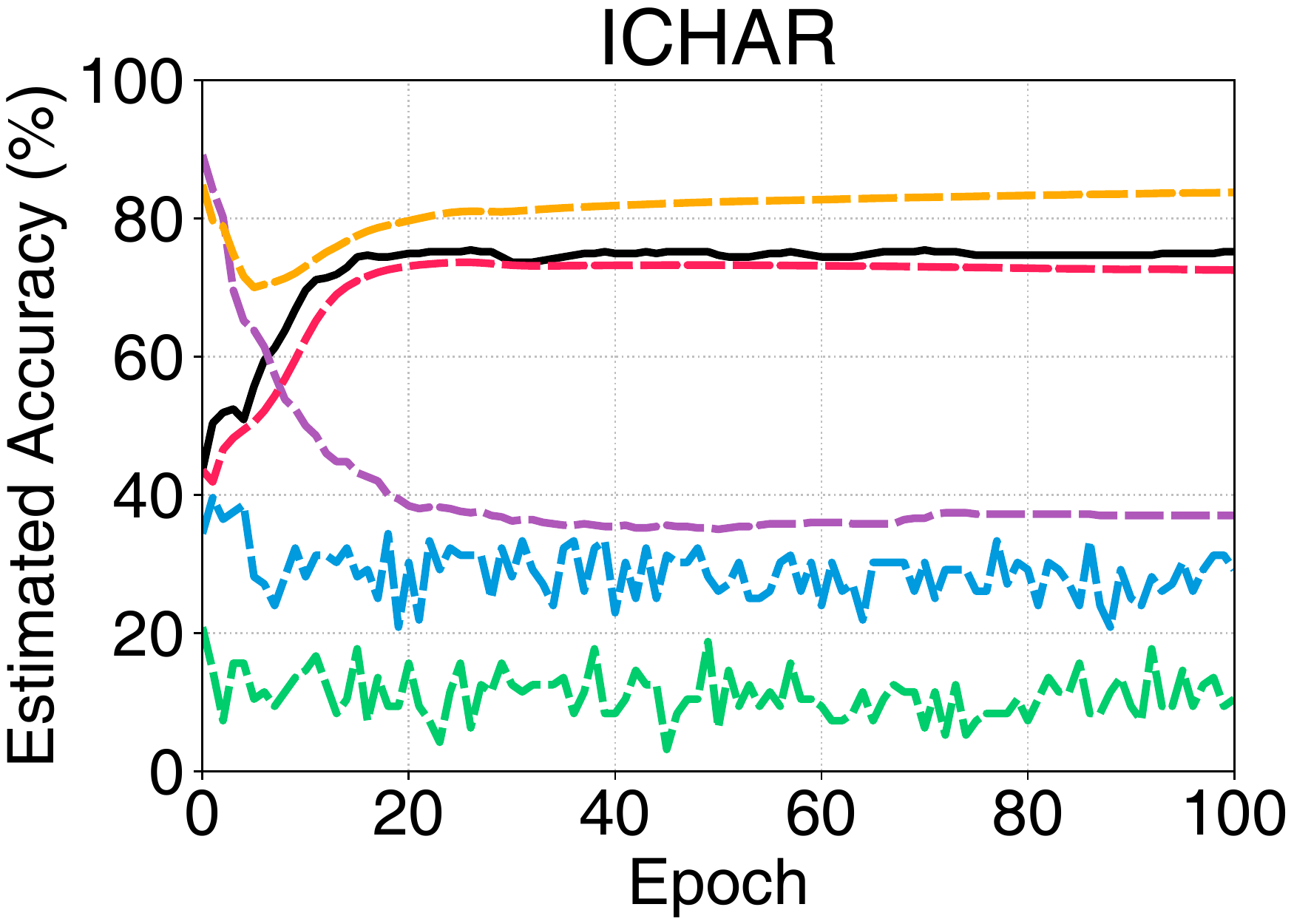}
        \end{subfigure}
        ~
        \begin{subfigure}[t]{0.25\textwidth}
            \centering
            \includegraphics[width=0.95\linewidth]{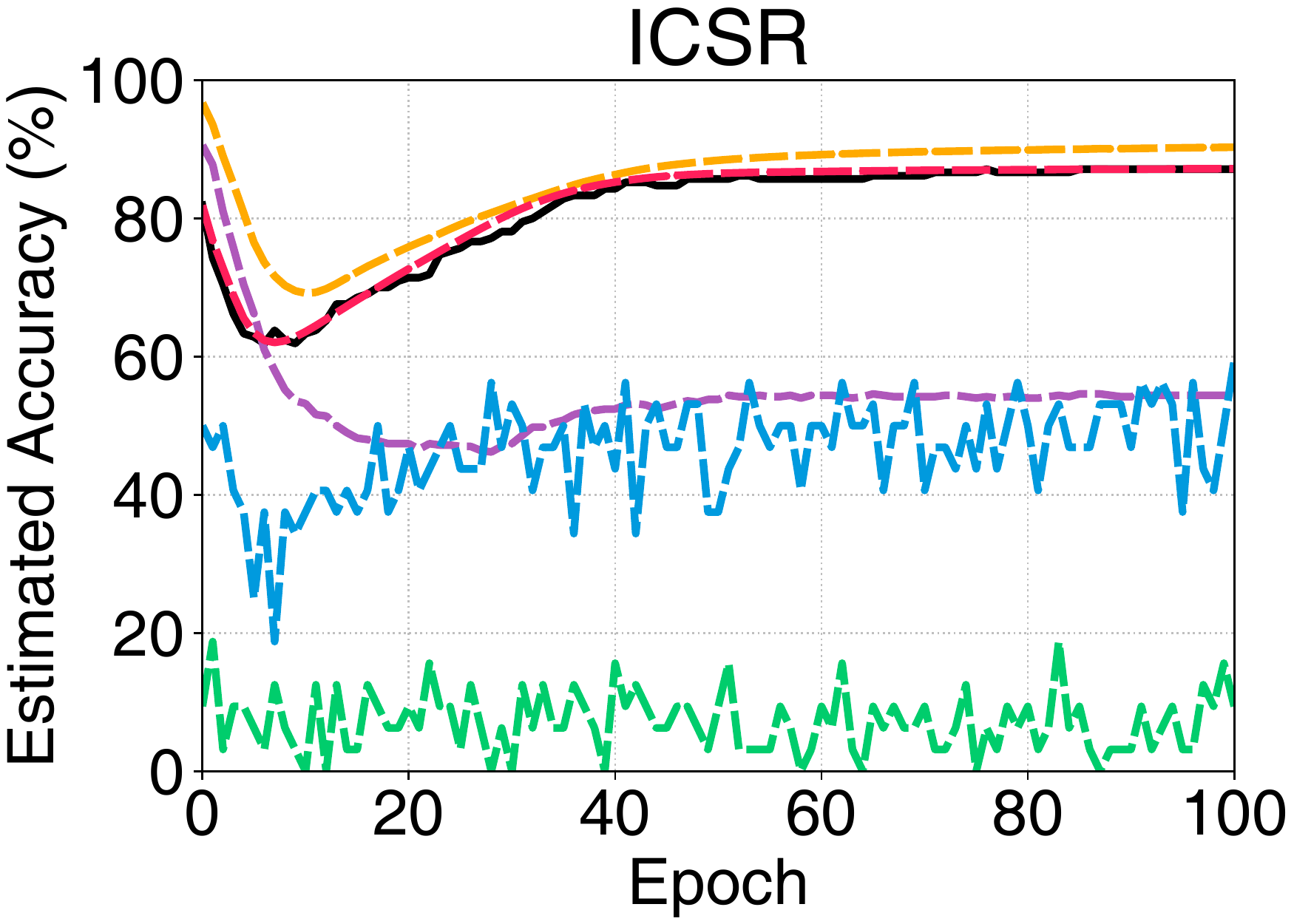}
        \end{subfigure}
    \centering
        \vspace{-0.6cm}
        \caption{Case 1: Increasing accuracy.}
        \vspace{0.3cm}
        \label{fig:qualitative:case1}
    \end{subfigure}

    \begin{subfigure}[t]{0.95\textwidth}
        \centering
        \begin{subfigure}[t]{0.25\textwidth}
            \centering
            \includegraphics[width=0.95\linewidth]{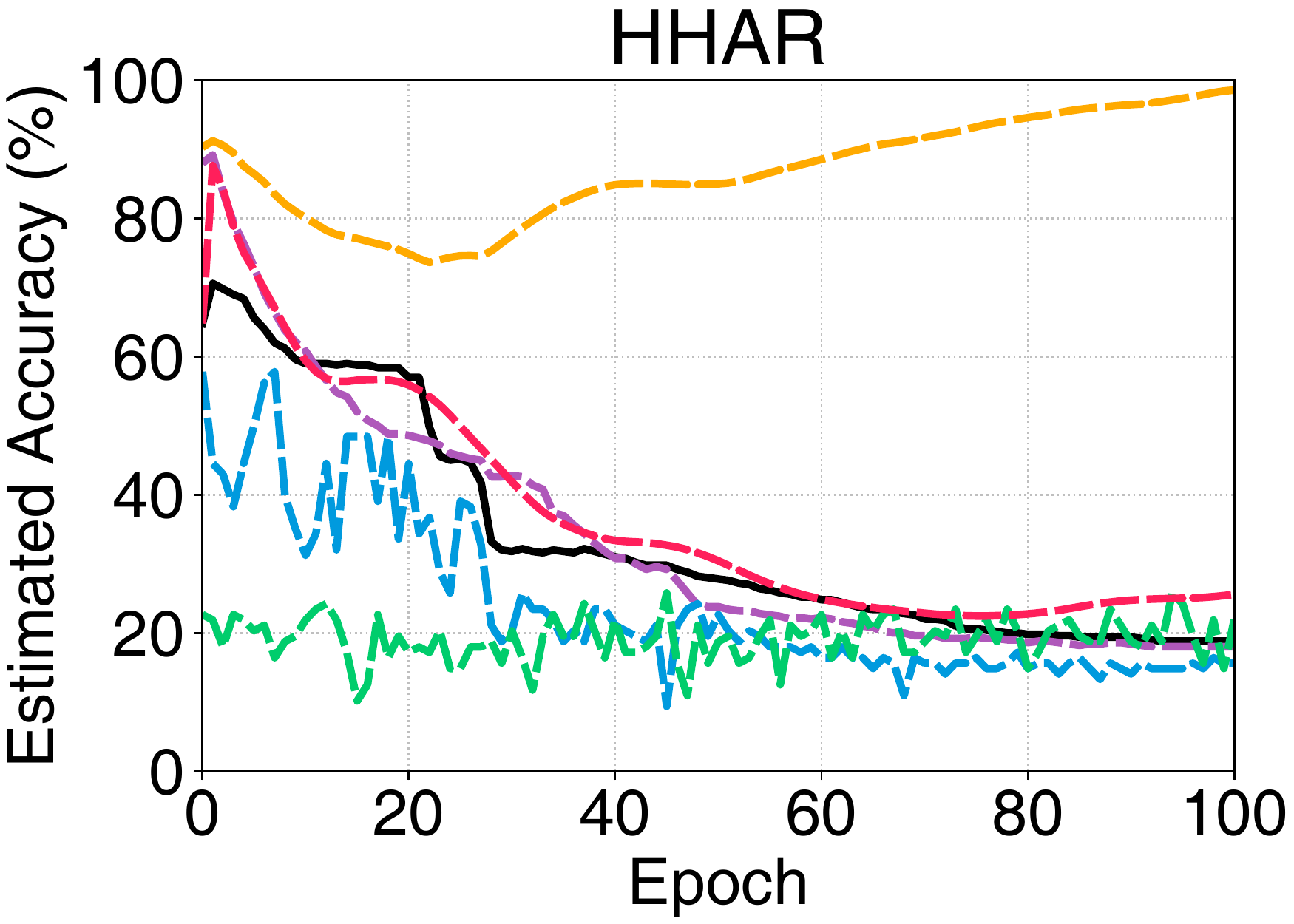}
        \end{subfigure}
        ~
        \begin{subfigure}[t]{0.25\textwidth}
            \centering
            \includegraphics[width=0.95\linewidth]{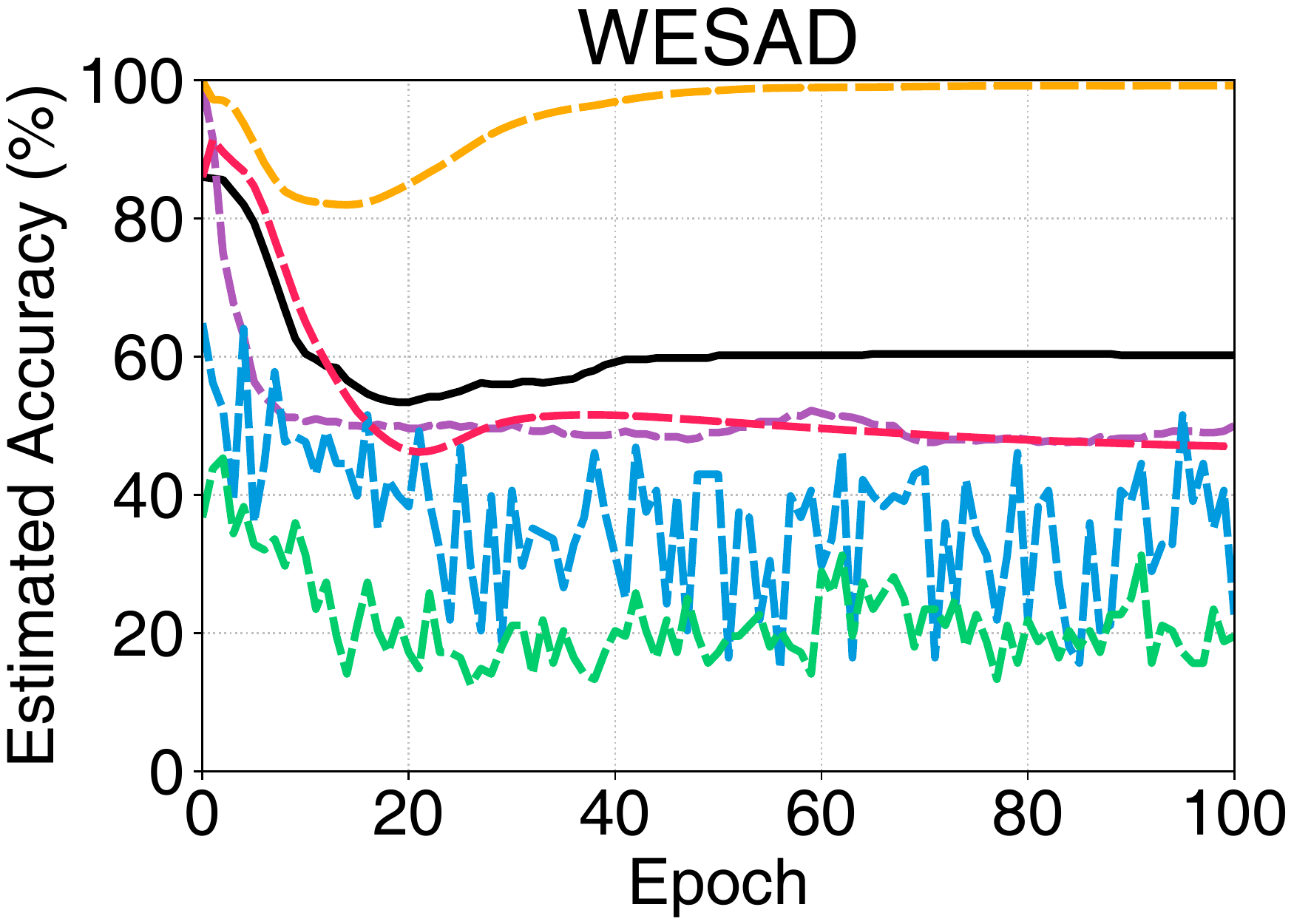}
        \end{subfigure}
        ~
        \begin{subfigure}[t]{0.25\textwidth}
            \centering
            \includegraphics[width=0.95\linewidth]{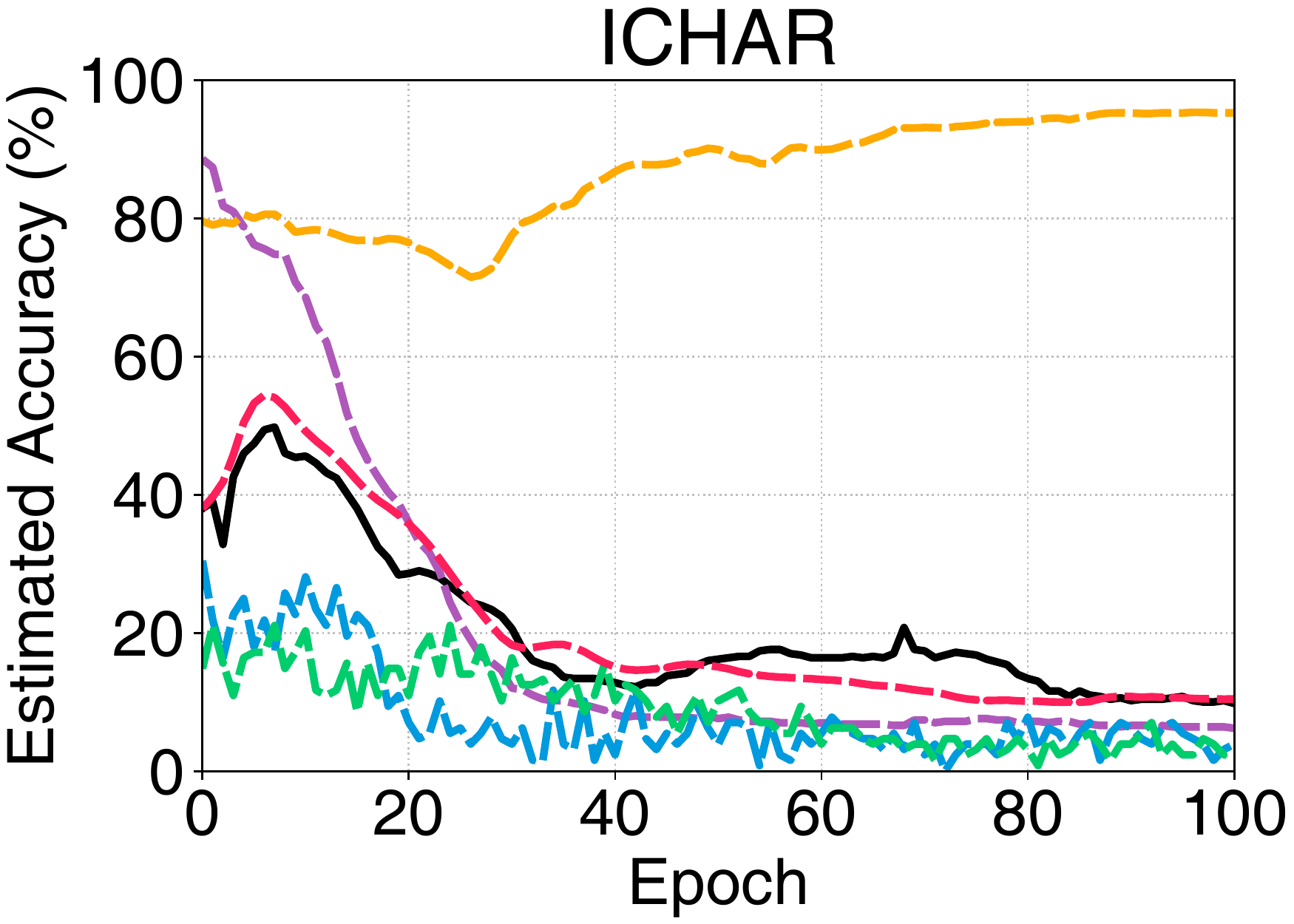}
        \end{subfigure}
        ~
        \begin{subfigure}[t]{0.25\textwidth}
            \centering
            \includegraphics[width=0.95\linewidth]{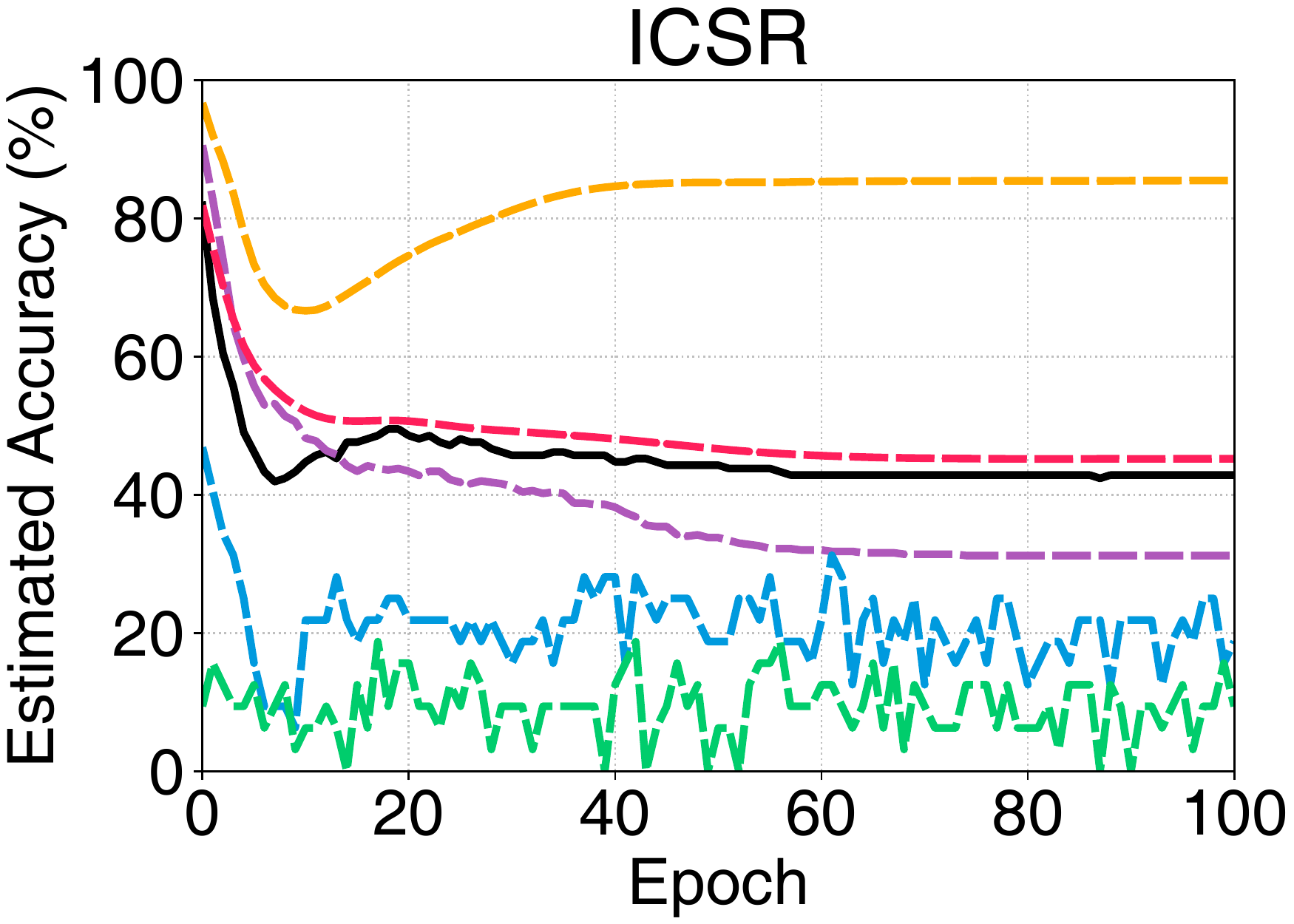}
        \end{subfigure}
        \vspace{-0.6cm}
        \caption{Case 2: Decreasing accuracy.}
        \vspace{0.3cm}
        \label{fig:qualitative:case2}
    \end{subfigure}
    
    \begin{subfigure}[t]{0.95\textwidth}
        \centering
        \begin{subfigure}[t]{0.25\textwidth}
            \centering
            \includegraphics[width=0.95\linewidth]{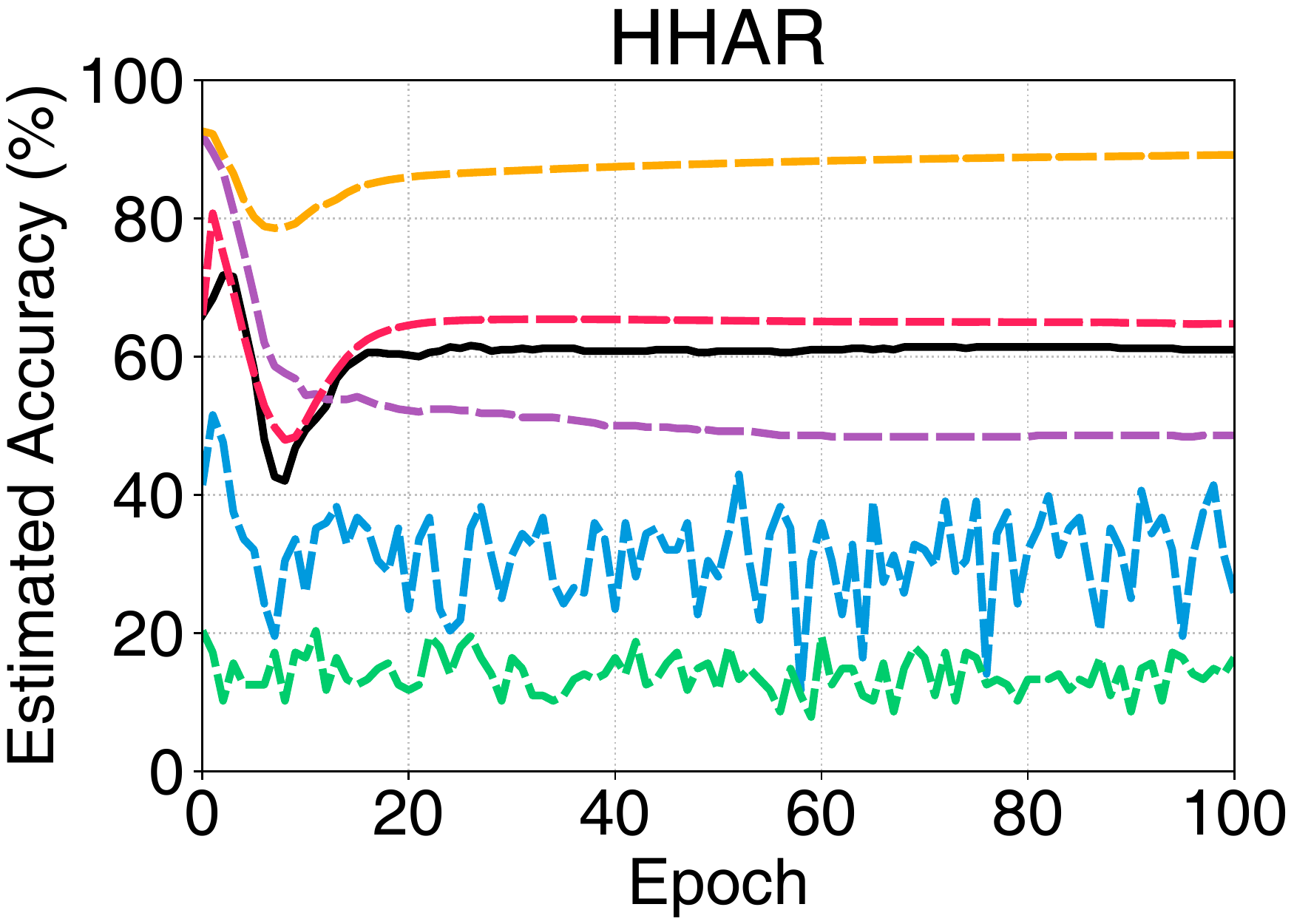}
        \end{subfigure}
        ~
        \begin{subfigure}[t]{0.25\textwidth}
            \centering
            \includegraphics[width=0.95\linewidth]{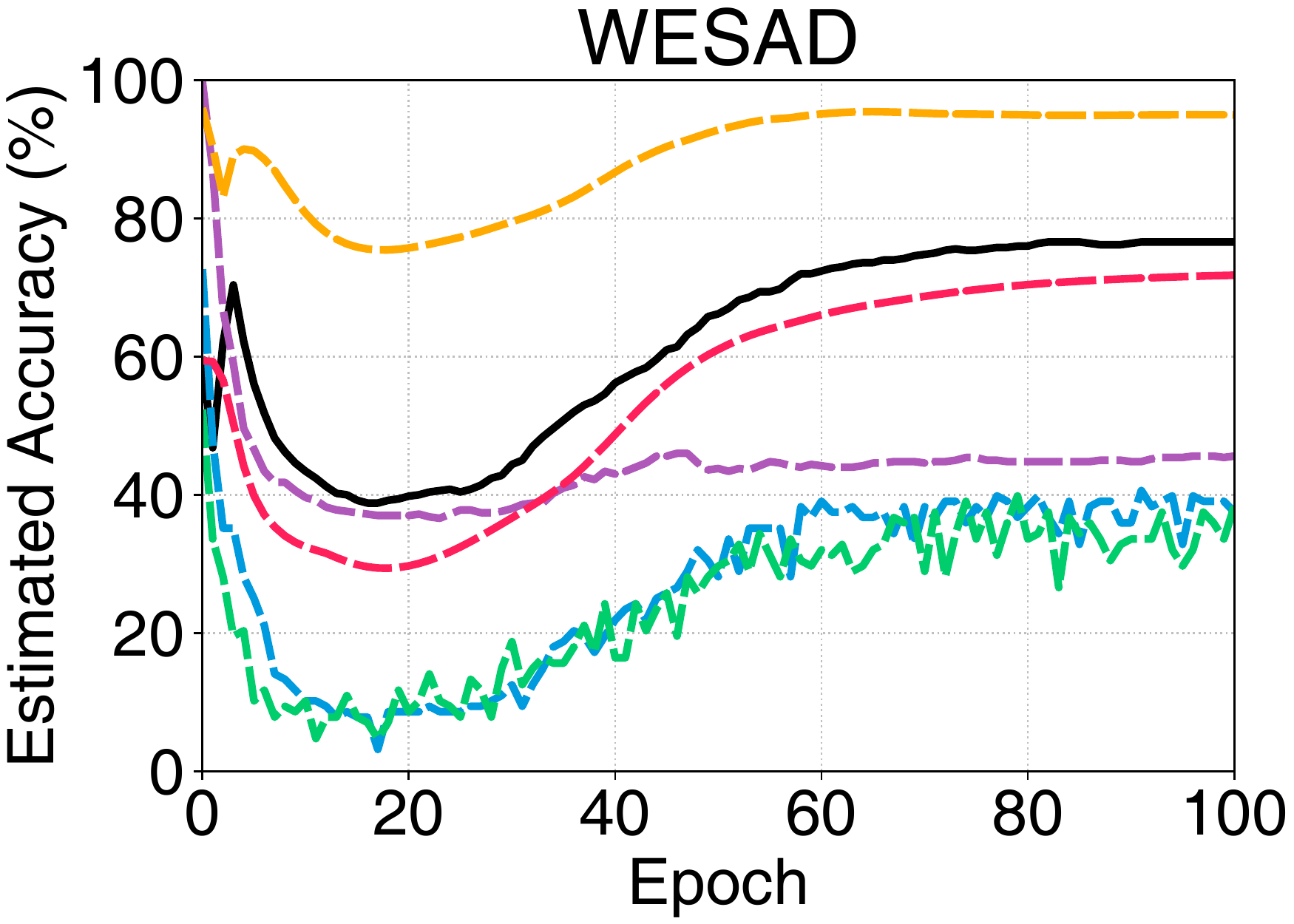}
        \end{subfigure}
        ~
        \begin{subfigure}[t]{0.25\textwidth}
            \centering
            \includegraphics[width=0.95\linewidth]{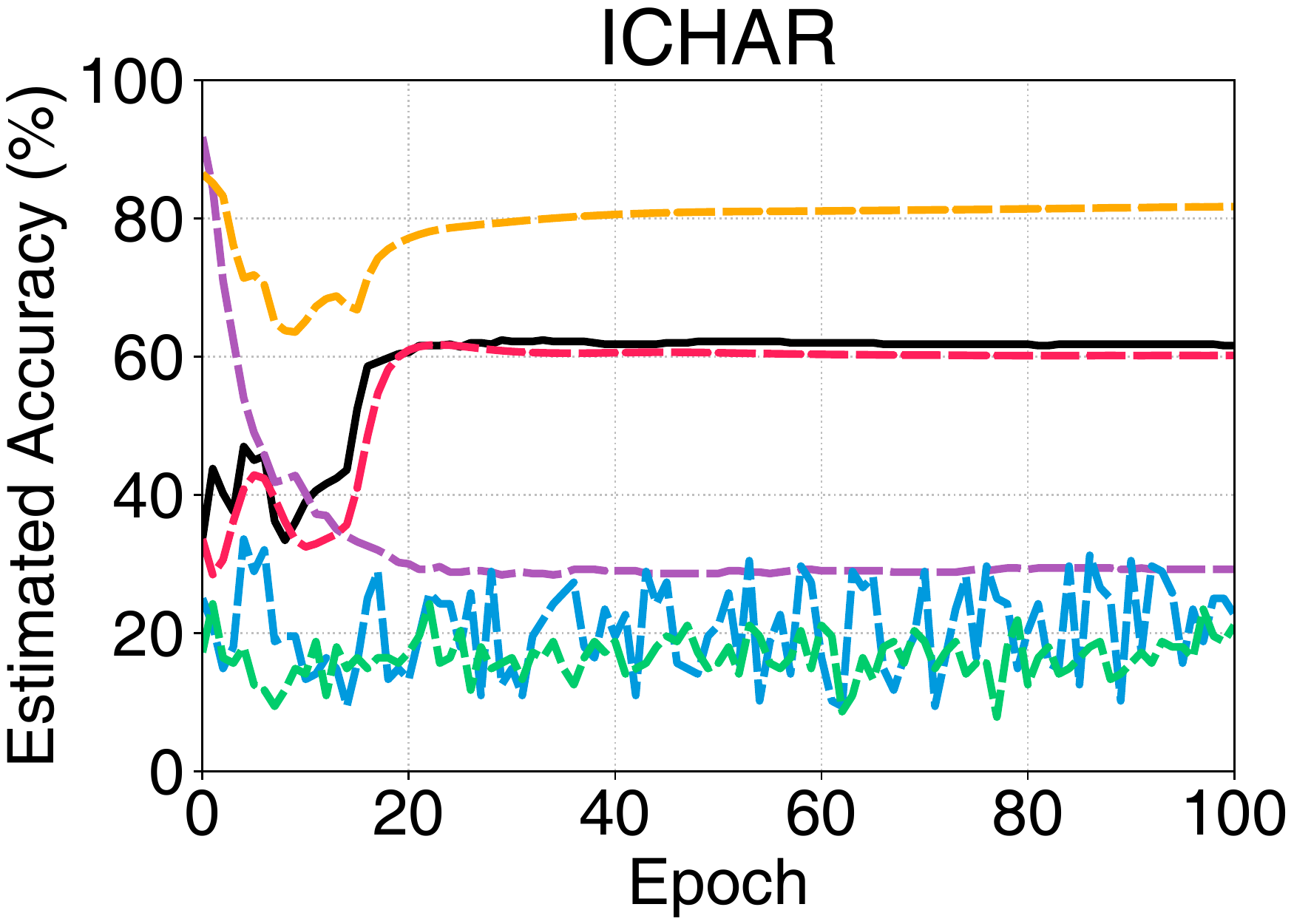}
        \end{subfigure}
        ~
        \begin{subfigure}[t]{0.25\textwidth}
            \centering
            \includegraphics[width=0.95\linewidth]{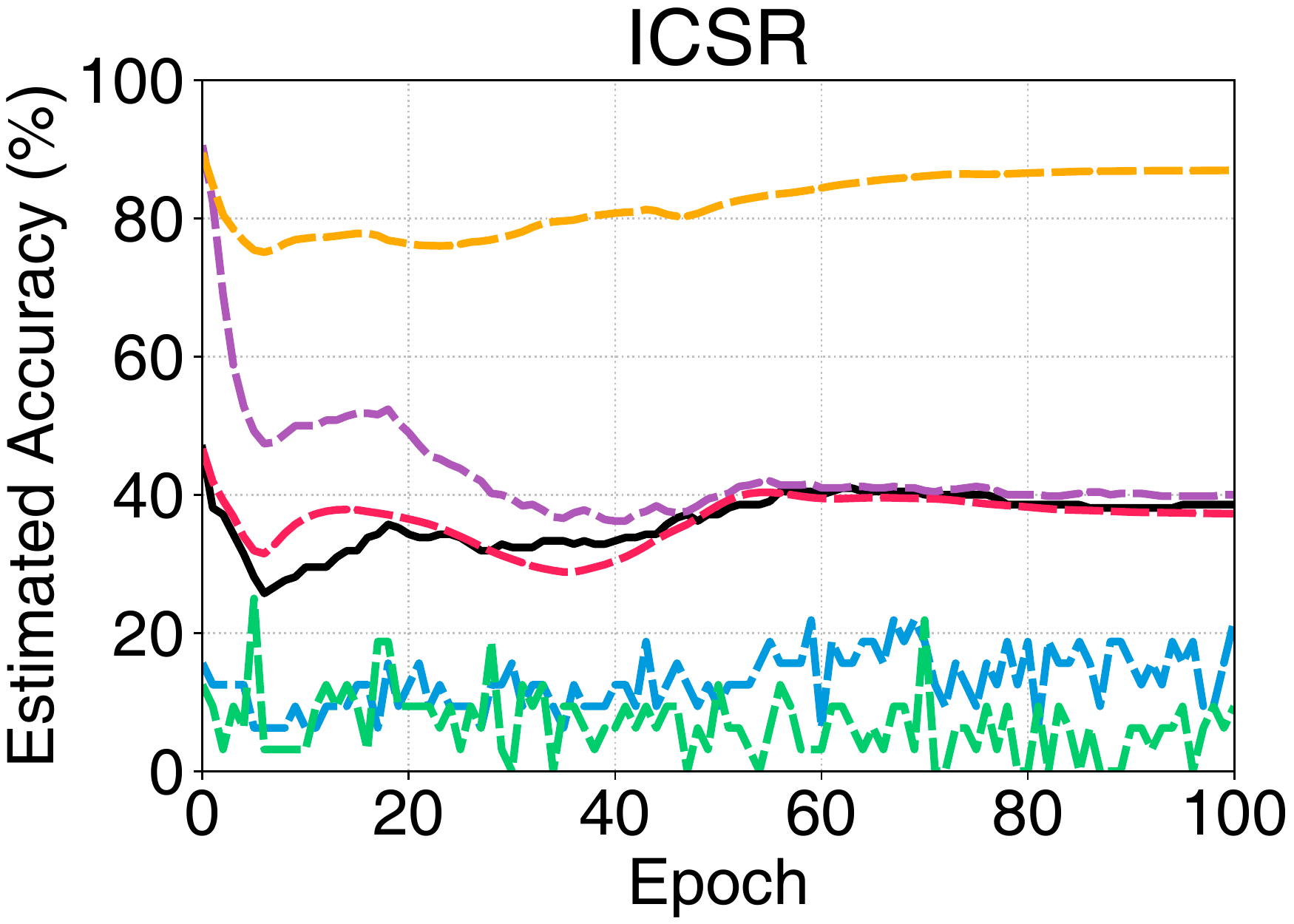}
        \end{subfigure}
        
        \vspace{-0.2cm}
        \caption{Case 3: Fluctuating accuracy.}
        \label{fig:qualitative:case3}
    \end{subfigure}
    
    \vspace{-0.3cm}
    \caption{Qualitative analysis with three categories: (a) increasing accuracy, (b) decreasing accuracy, and (c) fluctuating accuracy.}
    \vspace{-0.1cm}
    \label{fig:qualitative}
\end{figure*}
We conducted a qualitative analysis to understand further the behaviors of the baselines of \system{}. Specifically, we investigated how similar the estimated performance is to the true performance. Figure~\ref{fig:qualitative} compares \rev{different methods (SrcLabel, SoftmaxScore, DIR, EnsRM and \system{}) to TgtLabel (oracle)} under three different performance patterns: (i) increasing accuracy, (ii) decreasing accuracy, and (iii) fluctuating accuracy. 
While the other baselines directly predict the accuracy, \system{} predicts relative accuracy change. We added the true accuracy at epoch zero to the estimated accuracy by \system{} for better visualization.

The estimated performance by SrcLabel generally decreased as training proceeded, regardless of whether the actual performance improved or not. This means validation with hold-out source data could not be applied to unseen targets due to domain shifts. Thus, the estimated performance decreased as the model adapted to the target data. \rev{SoftmaxScore often showed high estimated accuracy in the decreasing case as the model became overconfident for misclassifications due to the incorrect calibration~\cite{10.5555/3305381.3305518}.} The estimations by DIR and EnsRM were noisy as they require updating the check model(s) for every adaptation epoch. Since the learned representations change as the training continues, the estimations fluctuate accordingly. In addition, the estimated accuracy by DIR and EnsRM was generally lower than the other algorithms, indicating that the disagreement between the target model and the check model(s) was higher than it should be, probably due to the low accuracy of the check model(s).

\system{} predicted accuracy better than other baselines in general. As \system{} relies on the target data to estimate the performance, it does not have the domain shift issue of SrcLabel mentioned above. Moreover, \system{} uses a pre-trained estimator for performance estimation and does not incur noisy estimations as DIR and EnsRM.

\subsection{Ablation Study}\label{sec:experiment:ablation}

\begin{table}[t]
\centering
\caption{The ablation study result of \system{}. Average similarities (\%) for four datasets under the four adaptation settings are reported. The numbers are averaged across target domains. Dataset names are abbreviated for better visibility (HHA:HHAR, WES: WESAD, ICH: ICHAR, and ICS:ICSR). Bold type numbers indicate those of the highest average similarity.}
\label{tab:ablation}
\resizebox{\columnwidth}{!}{%
\begin{tabular}{
lrrrrrrrrrrrrrrrrr
}
\Xhline{2\arrayrulewidth}
\addlinespace[0.08cm]
\multirow{2}{*}{\textbf{Method}} & \multicolumn{4}{c}{\textbf{Fine-tuning}} & \multicolumn{4}{c}{\textbf{DANN}}  & \multicolumn{4}{c}{\textbf{CDAN}} & \multicolumn{4}{c}{\textbf{SHOT}} & \multicolumn{1}{c}{\multirow{2}{*}{\textbf{Avg}}} \\
\cmidrule(lr){2-5} \cmidrule(lr){6-9} \cmidrule(lr){10-13} \cmidrule(lr){14-17}
 & \multicolumn{1}{c}{HHA} & \multicolumn{1}{c}{WES} & \multicolumn{1}{c}{ICH} & \multicolumn{1}{c}{ICS} 
 & \multicolumn{1}{c}{HHA} & \multicolumn{1}{c}{WES} & \multicolumn{1}{c}{ICH} & \multicolumn{1}{c}{ICS} 
 & \multicolumn{1}{c}{HHA} & \multicolumn{1}{c}{WES} & \multicolumn{1}{c}{ICH} & \multicolumn{1}{c}{ICS} 
 & \multicolumn{1}{c}{HHA} & \multicolumn{1}{c}{WES} & \multicolumn{1}{c}{ICH} & \multicolumn{1}{c}{ICS} 
 & \multicolumn{1}{c}{} \\
 
\addlinespace[0.02cm] 
\hline
\addlinespace[0.04cm]

GD & 86.2 & 77.7 & 91.5 & 91.1 & 94.4 & 94.3 & 92.4 & 95.1 & 95.4 & 91.3 & 93.5 & 93.4 & 93.3 & 84.7 & 89.2 & 87.6 & 90.7 \\
GD+IU & 91.0 & 81.0 & 94.4 & 89.7 & 94.8 & 94.4 & 93.0 & 95.7 & 95.1 & 93.1 & 92.2 & 93.5 & 90.4 & 84.5 & 89.4 & 88.5 & 91.3 \\
GD+IU+PD & 95.5 & 89.2 & 95.5 & 90.8 & 93.9 & 87.7 & 92.9 & 94.8 & 94.6 & 91.3 & 93.6 & 93.7 & 93.5 & 91.7 & 88.8 & 93.6 & \textbf{92.6} \\
\hline
Feat & 77.2 & 77.4 & 54.5 & 73.8 & 40.5 & 60.1 & 78.5 & 71.6 & 58.6 & 58.9 & 83.3 & 72.6 & 73.1 & 70.6 & 87.2 & 75.2 & 69.6 \\
$\Delta$Feat & 77.0 & 63.7 & 80.0 & 85.4 & 76.5 & 79.2 & 90.6 & 95.5 & 78.4 & 91.1 & 85.5 & 95.6 & 72.4 & 67.1 & 80.3 & 87.0 & 81.6 \\
Feat+$\Delta$Feat & 95.5 & 89.2 & 95.5 & 90.8 & 93.9 & 87.7 & 92.9 & 94.8 & 94.6 & 91.3 & 93.6 & 93.7 & 93.5 & 91.7 & 88.8 & 93.6 & \textbf{92.6} \\
\hline
MLP & 85.8 & 82.1 & 92.1 & 88.7 & 94.0 & 93.2 & 92.2 & 94.7 & 94.5 & 90.4 & 93.1 & 94.4 & 92.2 & 88.9 & 90.3 & 92.6 & 91.2 \\
LSTM & 95.5 & 89.2 & 95.5 & 90.8 & 93.9 & 87.7 & 92.9 & 94.8 & 94.6 & 91.3 & 93.6 & 93.7 & 93.5 & 91.7 & 88.8 & 93.6 & \textbf{92.6}\\

\Xhline{2\arrayrulewidth}
\end{tabular}%
}
\end{table}

We conducted ablative experiments on \system{} to analyze the effectiveness of our design choices in training \system{} (\cref{sec:method:dapper:training}). Table~\ref{tab:ablation} is the result. We first inspected the effect of the feature modalities used in training \system{}. We sequentially added features, global diversity (GD), individual uncertainty (IU), and prediction distribution (PD), and compared the average similarity values. As our suggested features were added, the average similarity became higher in general. This shows the effectiveness of our features for training the estimator. Second, we compared the effectiveness of using both the original features ($\mathit{Feat}$) and the difference of them ($\Delta\mathit{Feat}$) for performance estimation. We observe that using both the original and the difference features significantly improved the similarity further than using only one of them. This attests the importance of considering both the original value of the features and the difference of them from the previous epoch for accurate performance estimation. Finally, we compared our LSTM architecture with a simple multi-layer perceptron~(MLP) architecture to understand the effectiveness of LSTM. Specifically, we used a two-layer MLP with a hidden dimension of 20 and a ReLU activation, which is the same architecture used in our estimator, excluding the LSTM part. We found that leveraging LSTM shows more accurate predictions than MLP, demonstrating the effectiveness of utilizing sequential modeling in performance estimation.

\begin{table}[t]
\centering
\caption{Average similarities (\%) of the baselines and \system{} under two cross-dataset adaptation scenarios (from ICHAR to HHAR and from HHAR to ICHAR). Bold type numbers indicate those of the highest average similarity.}\label{tab:cross_dataset}
\resizebox{0.65\columnwidth}{!}{%
\begin{tabular}{
lrrrrrrrrrr
}
    \Xhline{2\arrayrulewidth}
    \addlinespace[0.12cm] 
 \multicolumn{1}{c}{\multirow{2}{*}{\textbf{Method}}} & \multicolumn{6}{c}{\textbf{ICHAR} $\rightarrow$ \textbf{HHAR}} & \multicolumn{3}{c}{\textbf{HHAR} $\rightarrow$ \textbf{ICHAR}} & \multicolumn{1}{c}{\multirow{2}{*}{\textbf{Avg.}}} \\
 \cmidrule(lr){2-7} \cmidrule(lr){8-10}
 & \multicolumn{1}{c}{D1} & \multicolumn{1}{c}{D2} & \multicolumn{1}{c}{D3} & \multicolumn{1}{c}{D4} & \multicolumn{1}{c}{D5} & \multicolumn{1}{c}{D6} & \multicolumn{1}{c}{D1} & \multicolumn{1}{c}{D2} & \multicolumn{1}{c}{D3} & \multicolumn{1}{c}{} \\

    \addlinespace[0.02cm] 
    \hline
    \addlinespace[0.2cm] 
    \multicolumn{11}{l}{\textit{Adaptation algorithm: Fine-tuning}}\\
    \hline
    \addlinespace[0.04cm] 
    
TgtLabel & 100.0 & 100.0 & 100.0 & 100.0 & 100.0 & 100.0 & 100.0 & 100.0 & 100.0 & 100.0 \\
FixedEpoch & N/A & N/A & N/A & N/A & N/A & N/A & N/A & N/A & N/A & N/A \\
SrcLabel & 73.2 & 75.3 & 59.4 & 51.3 & 66.2 & 62.4 & 69.2 & 63.1 & 74.4 & 66.1 \\
SoftmaxScore & 81.1 & 84.7 & 91.3 & 89.7 & 86.1 & 89.8 & 80.0 & 86.2 & 85.4 & 86.0 \\
DIR & 52.5 & 58.7 & 55.7 & 50.3 & 50.1 & 53.4 & 60.5 & 31.8 & 58.2 & 52.4 \\
EnsRM & 50.8 & 53.8 & 41.2 & 39.6 & 47.8 & 47.5 & 58.1 & 45.7 & 53.0 & 48.6 \\
DAPPER (ours) & 91.7 & 95.3 & 89.5 & 92.6 & 87.4 & 90.5 & 95.0 & 65.6 & 86.3 & \textbf{88.2} \\

    \addlinespace[0.02cm] 
    \hline
    \addlinespace[0.2cm] 
    \multicolumn{11}{l}{\textit{Adaptation algorithm: DANN}}\\
    \hline
    \addlinespace[0.04cm] 
    
TgtLabel & 100.0 & 100.0 & 100.0 & 100.0 & 100.0 & 100.0 & 100.0 & 100.0 & 100.0 & 100.0 \\
FixedEpoch & N/A & N/A & N/A & N/A & N/A & N/A & N/A & N/A & N/A & N/A \\
SrcLabel & 83.2 & 77.4 & 64.0 & 65.4 & 59.2 & 52.6 & 48.8 & 43.0 & 61.3 & 61.6 \\
SoftmaxScore & 87.6 & 81.6 & 67.2 & 72.7 & 69.1 & 62.7 & 47.1 & 51.7 & 67.4 & 67.5 \\
DIR & 69.4 & 69.4 & 90.0 & 77.7 & 83.1 & 90.7 & 73.2 & 77.5 & 86.2 & 79.7 \\
EnsRM & 52.4 & 54.9 & 67.9 & 64.9 & 75.3 & 79.8 & 85.6 & 82.0 & 79.6 & 71.4 \\
DAPPER (ours) & 73.1 & 91.5 & 75.7 & 59.6 & 87.9 & 85.6 & 93.6 & 78.5 & 95.4 & \textbf{82.3} \\

    \addlinespace[0.02cm] 
    \hline
    \addlinespace[0.2cm] 
    \multicolumn{11}{l}{\textit{Adaptation algorithm: CDAN}}\\
    \hline
    \addlinespace[0.04cm] 
    
TgtLabel & 100.0 & 100.0 & 100.0 & 100.0 & 100.0 & 100.0 & 100.0 & 100.0 & 100.0 & 100.0 \\
FixedEpoch & N/A & N/A & N/A & N/A & N/A & N/A & N/A & N/A & N/A & N/A \\
SrcLabel & 83.5 & 79.0 & 68.3 & 64.2 & 56.2 & 65.4 & 50.2 & 56.1 & 66.0 & 65.4 \\
SoftmaxScore & 71.1 & 68.1 & 55.3 & 53.9 & 46.8 & 55.1 & 36.8 & 45.0 & 55.8 & 54.2 \\
DIR & 67.1 & 67.2 & 78.5 & 90.0 & 87.8 & 85.6 & 76.9 & 74.6 & 91.0 & 79.8 \\
EnsRM & 53.0 & 56.1 & 68.5 & 67.9 & 84.1 & 69.3 & 95.4 & 79.7 & 75.7 & 72.2 \\
DAPPER (ours) & 74.5 & 91.2 & 87.3 & 66.2 & 87.4 & 89.2 & 95.3 & 67.9 & 97.3 & \textbf{84.0}\\

    \addlinespace[0.02cm] 
    \hline
    \addlinespace[0.2cm] 
    \multicolumn{11}{l}{\textit{Adaptation algorithm: SHOT}}\\
    \hline
    \addlinespace[0.04cm] 
    
TgtLabel & 100.0 & 100.0 & 100.0 & 100.0 & 100.0 & 100.0 & 100.0 & 100.0 & 100.0 & 100.0 \\
FixedEpoch & N/A & N/A & N/A & N/A & N/A & N/A & N/A & N/A & N/A & N/A \\
SrcLabel & 91.7 & 74.5 & 59.4 & 59.8 & 60.5 & 47.7 & 49.1 & 43.8 & 64.7 & 61.2 \\
SoftmaxScore & 91.8 & 79.0 & 59.7 & 66.2 & 67.1 & 58.0 & 35.9 & 32.6 & 54.7 & 60.6 \\
DIR & 49.2 & 66.8 & 86.1 & 85.9 & 91.2 & 91.2 & 74.9 & 80.1 & 91.1 & 79.6 \\
EnsRM & 39.3 & 55.1 & 71.7 & 69.2 & 72.6 & 86.4 & 94.9 & 85.9 & 73.6 & 72.1 \\
DAPPER (ours) & 72.8 & 93.2 & 71.2 & 78.1 & 85.2 & 87.6 & 95.3 & 73.1 & 96.6 & \textbf{83.7}\\

    \Xhline{2\arrayrulewidth}
\end{tabular}%
}
\end{table}

\begin{figure*}[t]
\captionsetup[subfigure]{justification=centering} 
    \centering
    \begin{subfigure}[t]{0.475\textwidth}
        \centering
        \begin{subfigure}[t]{0.5\textwidth}
            \centering
            \includegraphics[width=0.95\linewidth]{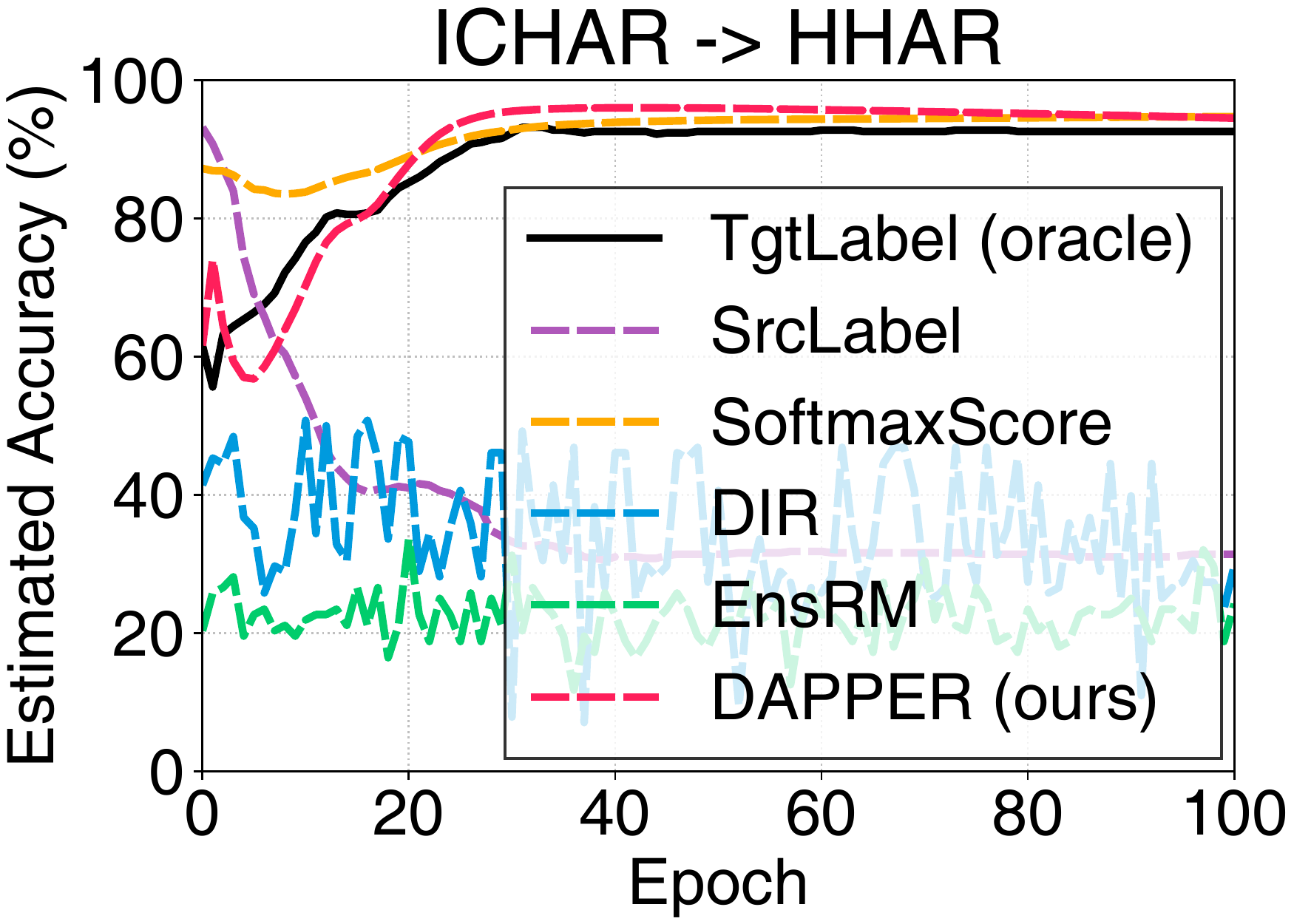}
        \end{subfigure}
        ~
        \begin{subfigure}[t]{0.5\textwidth}
            \centering
            \includegraphics[width=0.95\linewidth]{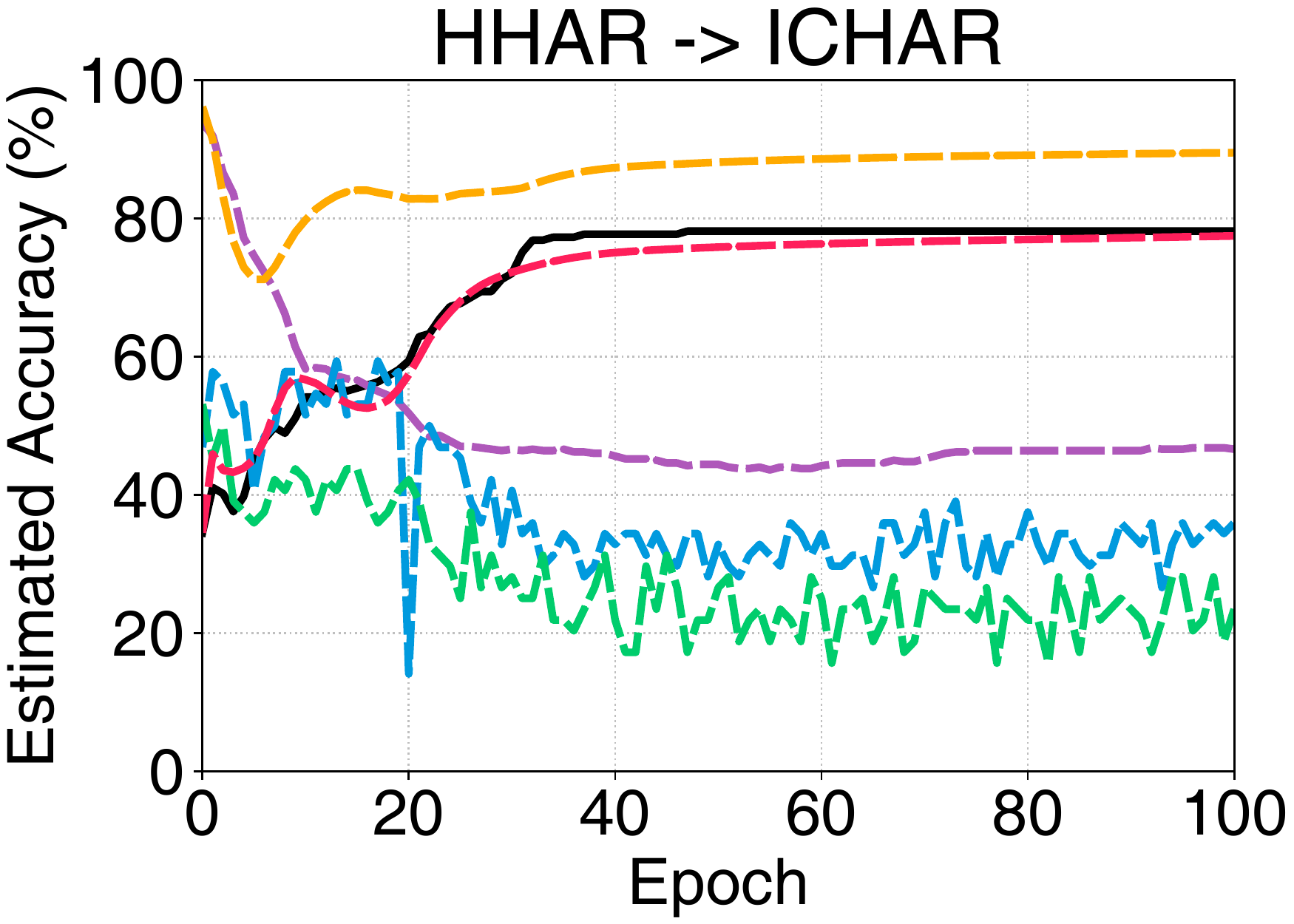}
        \end{subfigure}
    \centering
        \vspace{-0.6cm}
        \caption{Case 1: Increasing accuracy.}
        \vspace{0.3cm}
        \label{fig:qualitative_cross_data:case1}
    \end{subfigure}
    ~
    \begin{subfigure}[t]{0.475\textwidth}
        \centering
        \begin{subfigure}[t]{0.5\textwidth}
            \centering
            \includegraphics[width=0.95\linewidth]{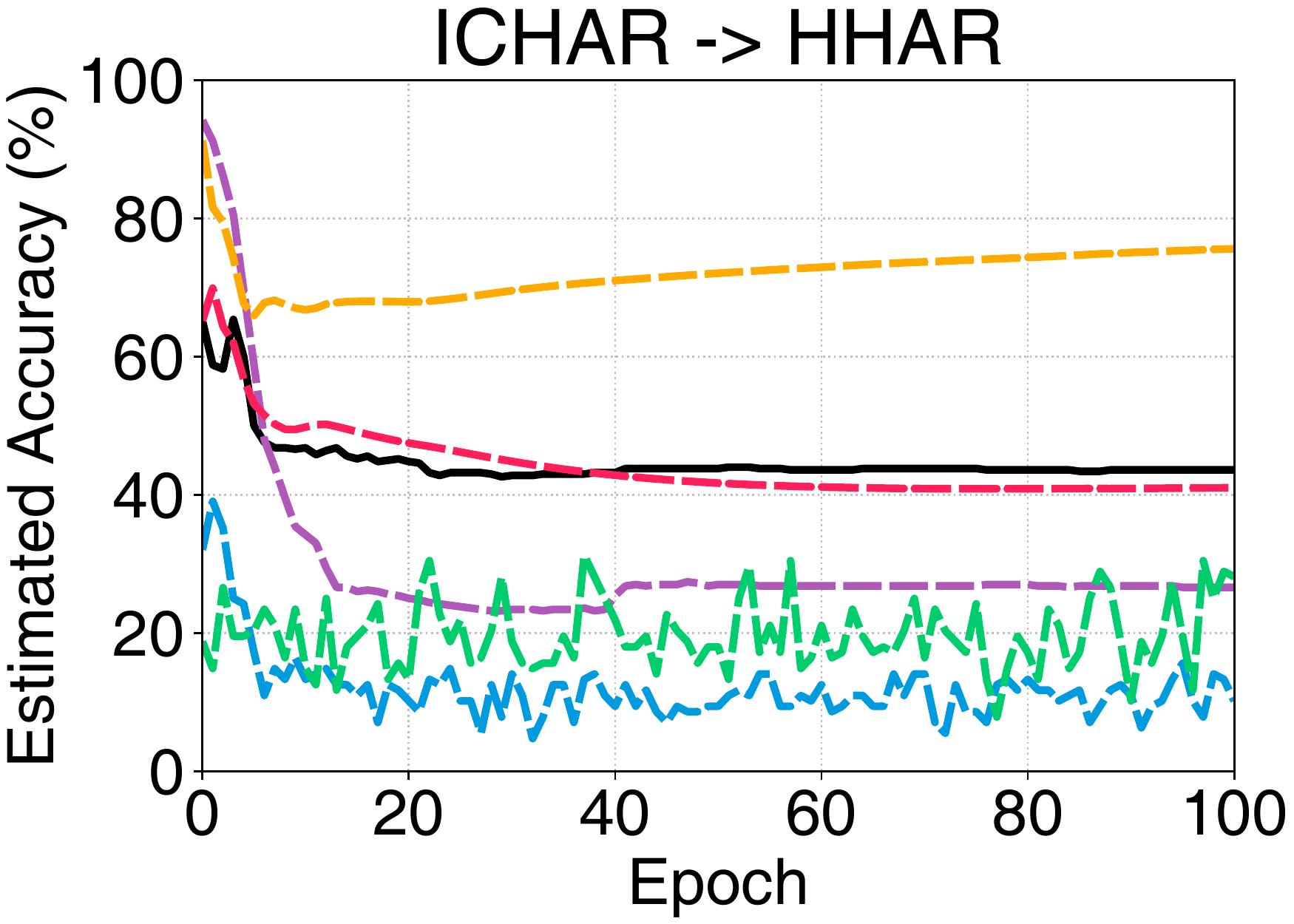}
        \end{subfigure}
        ~
        \begin{subfigure}[t]{0.5\textwidth}
            \centering
            \includegraphics[width=0.95\linewidth]{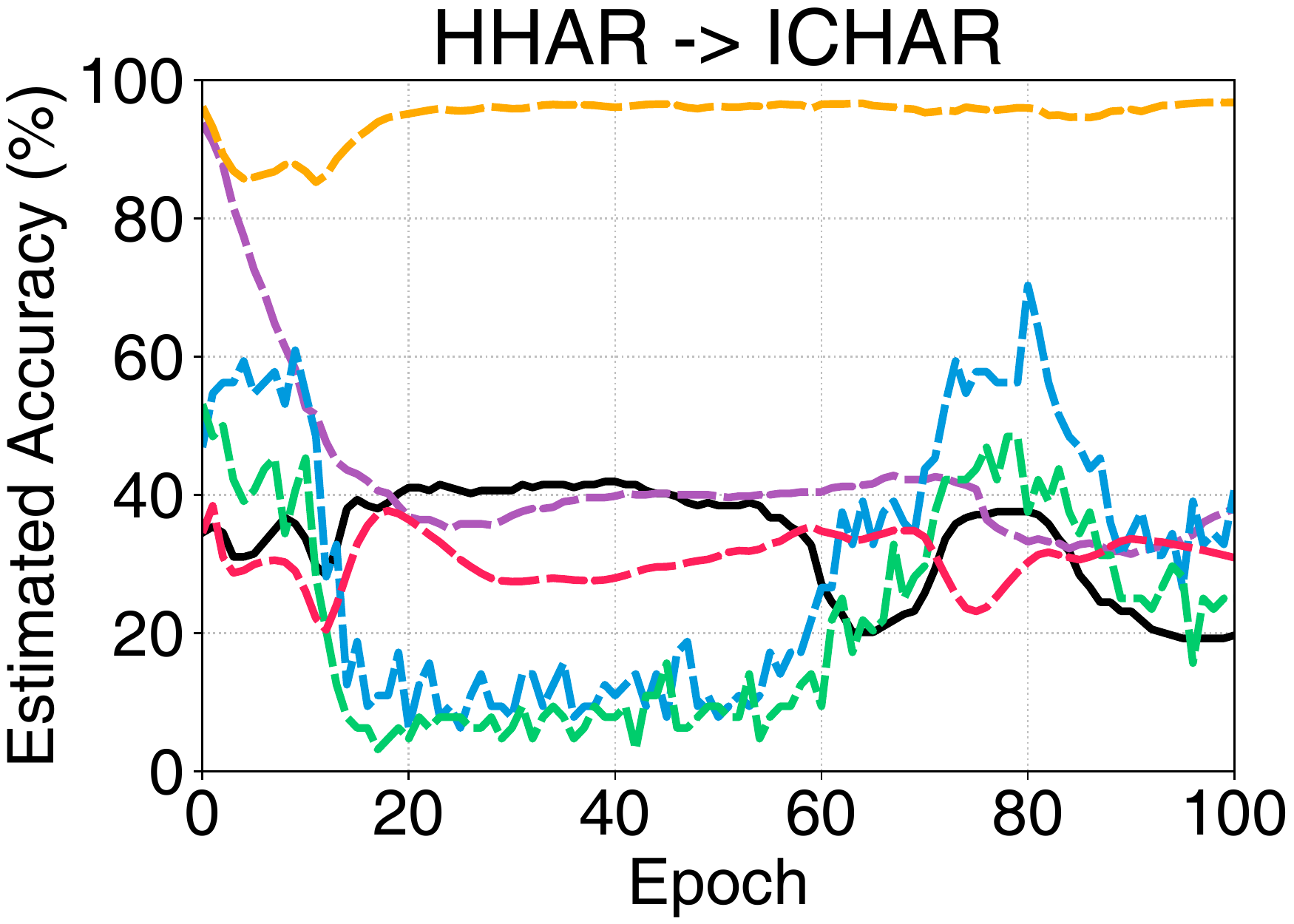}
        \end{subfigure}
        \vspace{-0.6cm}
        \caption{Case 2: Decreasing accuracy.}
        \vspace{0.3cm}
        \label{fig:qualitative_cross_data:case2}
    \end{subfigure}
    
    \begin{subfigure}[t]{0.95\textwidth}
        \centering
        \begin{subfigure}[t]{0.25\textwidth}
            \centering
            \includegraphics[width=0.95\linewidth]{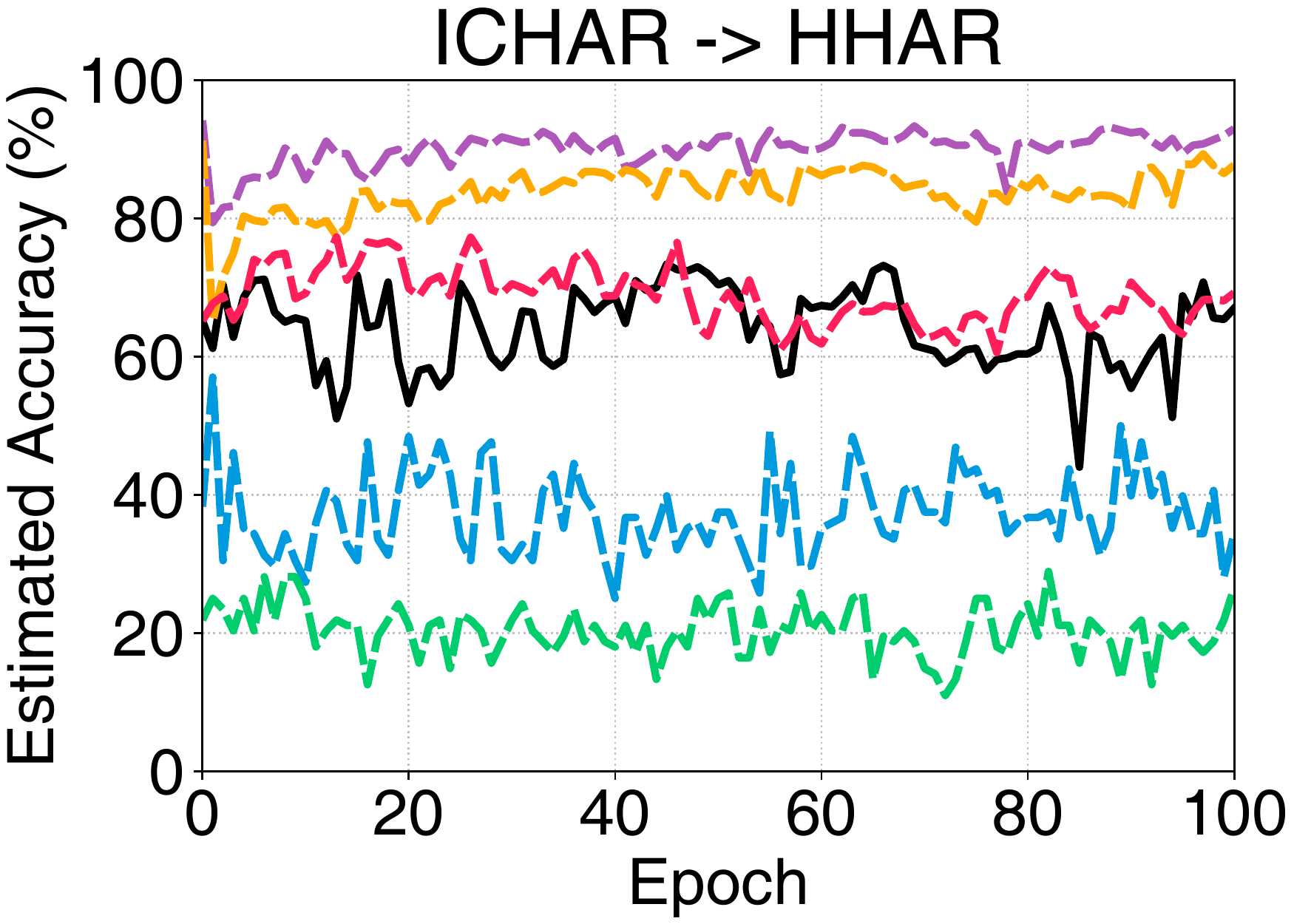}
        \end{subfigure}
        ~
        \begin{subfigure}[t]{0.25\textwidth}
            \centering
            \includegraphics[width=0.95\linewidth]{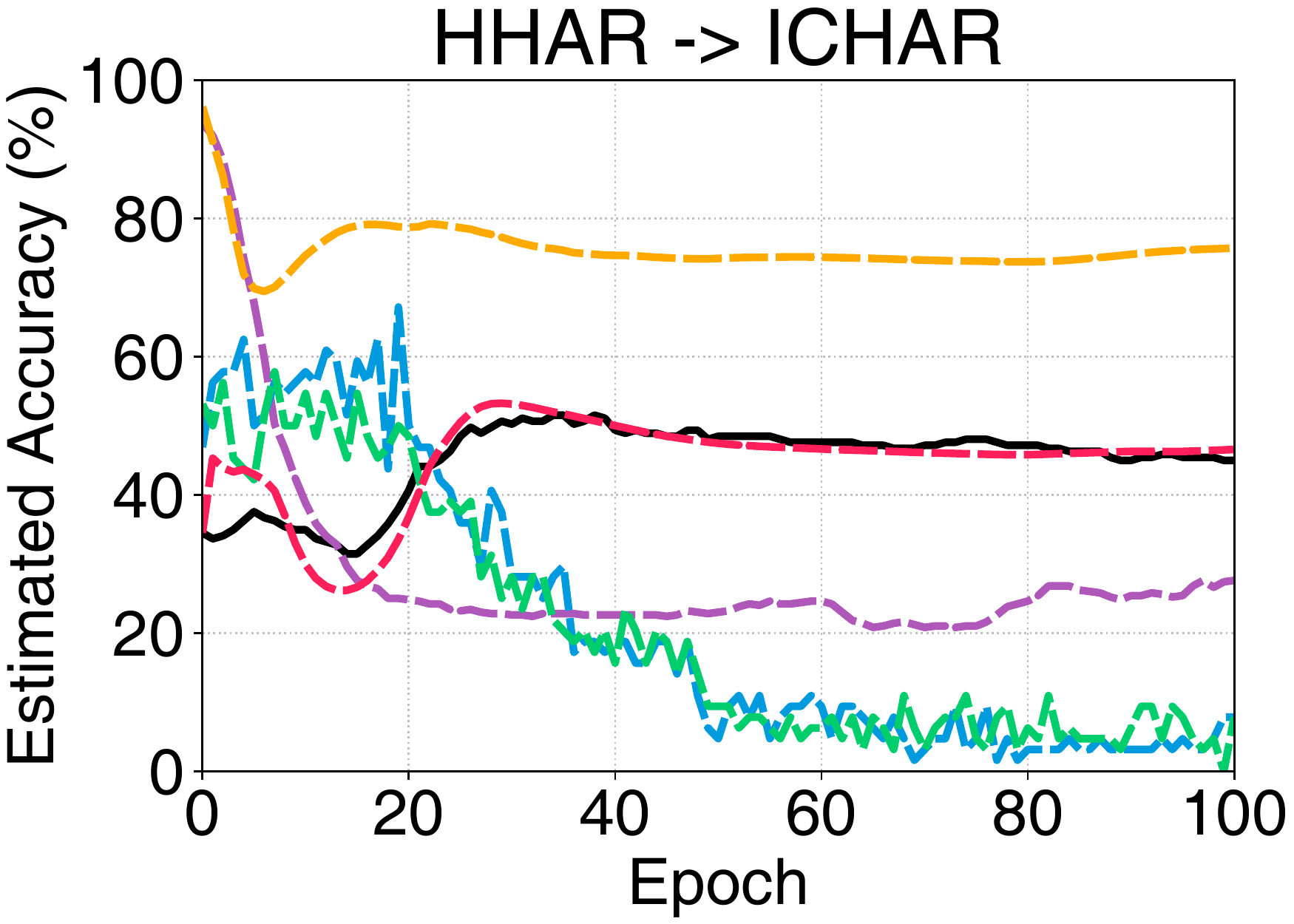}
        \end{subfigure}
        
        \vspace{-0.2cm}
        \caption{Case 3: Fluctuating accuracy.}
        \label{fig:qualitative_cross_data:case3}
    \end{subfigure}
    
    \caption{\rev{Qualitative analysis for two cross-dataset adaptation scenarios (from ICHAR to HHAR and from HHAR to ICHAR) with three categories: (a) increasing accuracy, (b) decreasing accuracy, and (c) fluctuating accuracy.}}
    \label{fig:qualitative_cross_data}
\end{figure*}

\subsection{Cross-dataset Adaptation}\label{sec:experiment:cross_dataset}
Cross-dataset adaptation is challenging but beneficial as it utilizes well-established existing data for a new related task. It has been investigated in mobile sensing scenarios such as human activity recognition~\cite{10.1145/3369818}. We evaluate \system{} under this challenging setting to further understand the estimated accuracy and possible limitations of \system{} under such a setting. Specifically, we selected the five common classes between HHAR and ICHAR datasets (sit, stand, walk, stairs up, and stairs down). We consider two scenarios: adaptation from HHAR to ICHAR and adaptation from ICHAR to HHAR. We use the same source and target domains used in the previous experiment in Table~\ref{tab:overall_result}. 

\rev{Table~\ref{tab:cross_dataset} shows the result and Figure~\ref{fig:qualitative_cross_data} illustrates the qualitative analysis. The result} shows that the overall similarities decreased compared with the within-dataset experiments. This might be attributed to more significant domain gaps, demonstrating the fundamental difficulty of the task. Still, the trend is similar to the previous within-dataset experiments, and \system{} overall shows better performances than the other baselines.

\subsection{Impact of Validation Data Distribution}\label{sec:experiment:imbalance}

For the other experiments, we used 500 randomly sampled validation data, which naturally follows the target data distribution. In practice, however, it is possible for target validation data to be imbalanced. Taking human activity recognition for instance, one could stay home during sick leaves, which would contain more static activities (sleeping, sitting, etc.) than dynamic activities. In this case, performance estimation would be conducted under skewed validation data. We investigated how resilient \system{} is on such an imbalanced distribution of target validation data. Specifically, for each dataset, we randomly selected half of the classes and dropped 20\%, 40\%, 60\%, and 80\% out of the 500 samples for the selected classes. The number of samples used for validation decreased accordingly.

\begin{figure}[t]
\captionsetup[subfigure]{justification=centering} 

    \centering
    \begin{subfigure}[t]{1\linewidth}
        \centering
        \begin{subfigure}[t]{0.25\linewidth}
            \centering
            \includegraphics[width=0.95\linewidth]{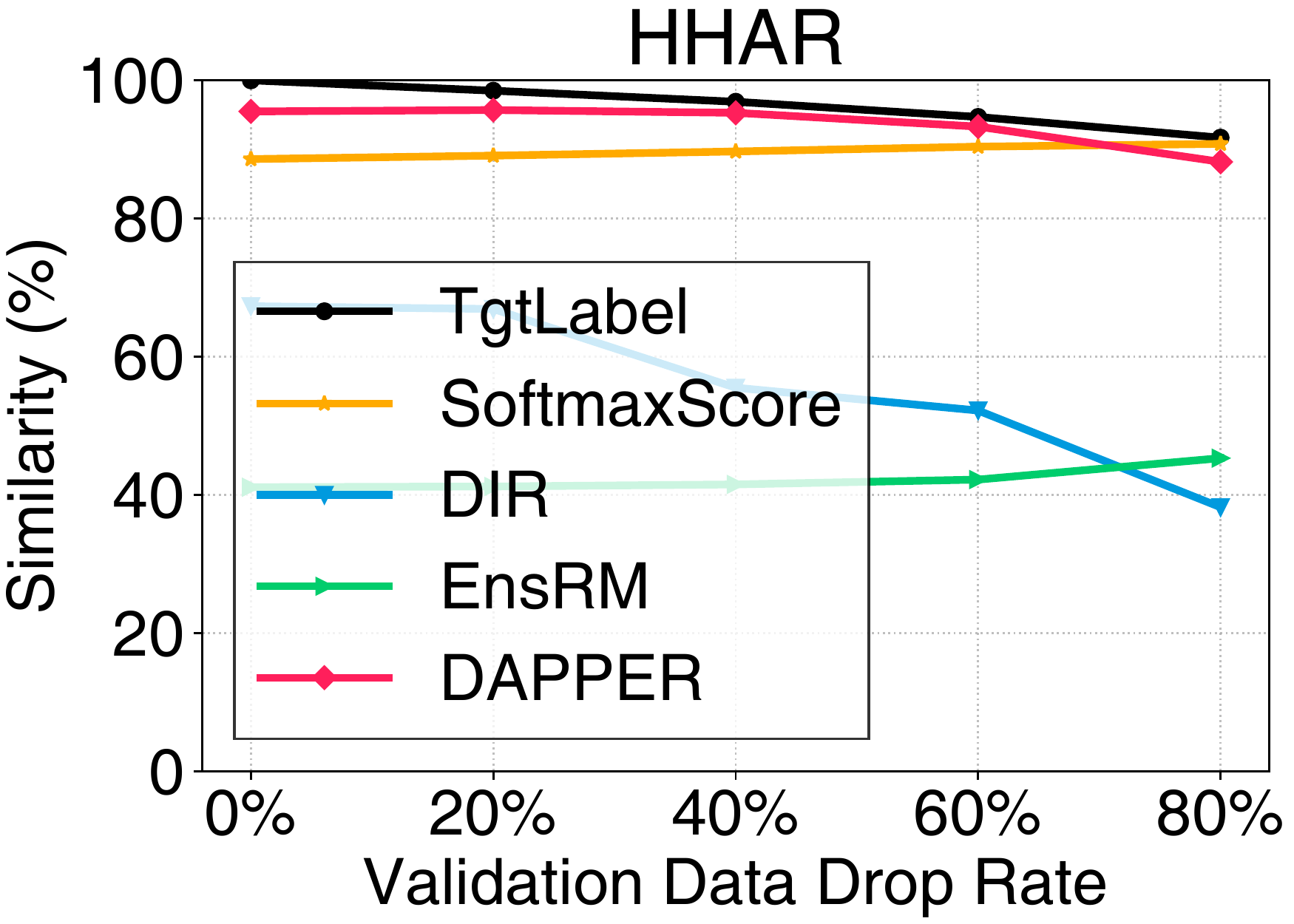}
            \label{fig:imb:sda:hhar}
        \end{subfigure}
        ~
        \begin{subfigure}[t]{0.25\linewidth}
            \centering
            \includegraphics[width=0.95\linewidth]{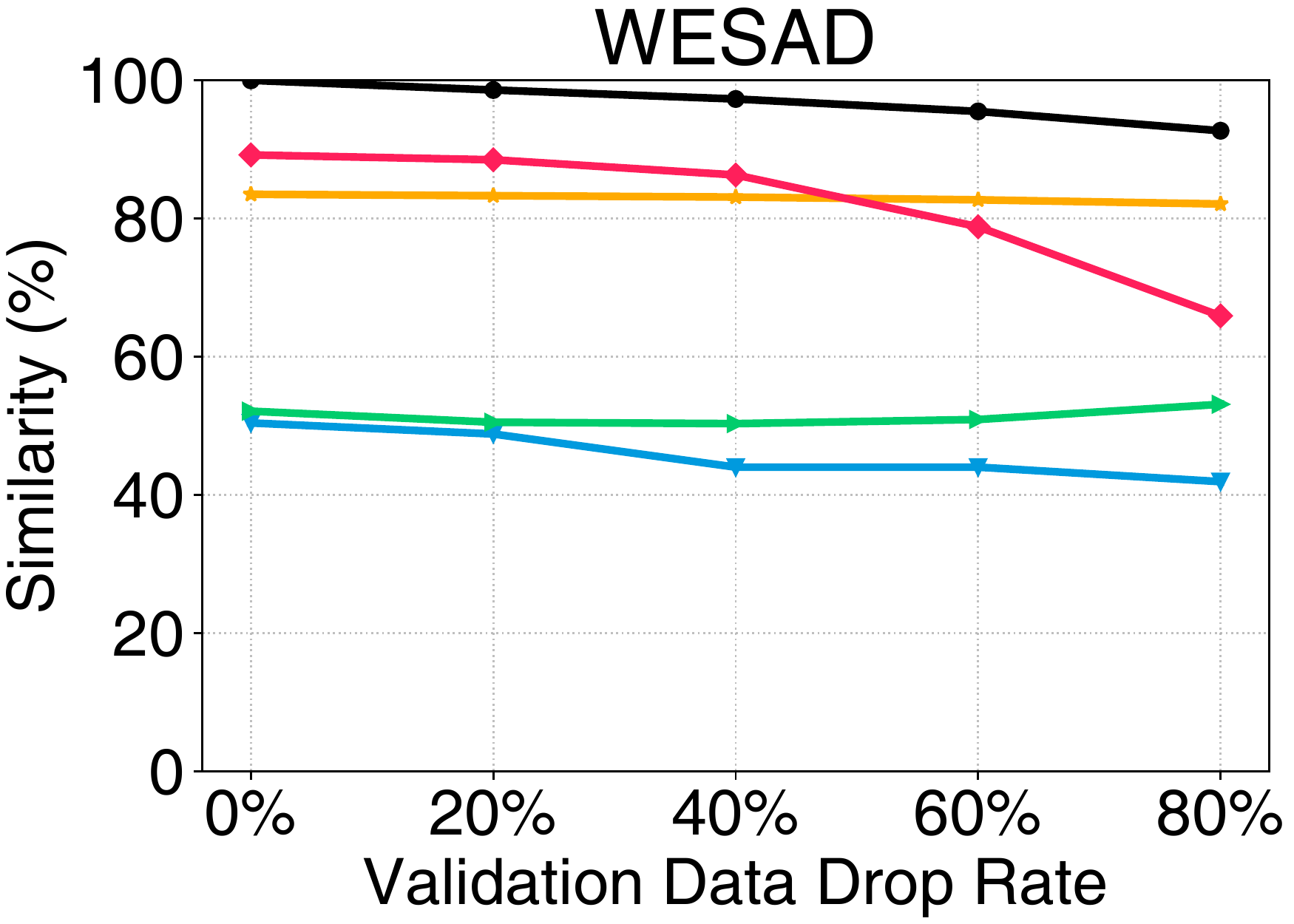}
            \label{fig:imb:sda:wesad}
        \end{subfigure}
        ~
        \begin{subfigure}[t]{0.25\linewidth}
            \centering
            \includegraphics[width=0.95\linewidth]{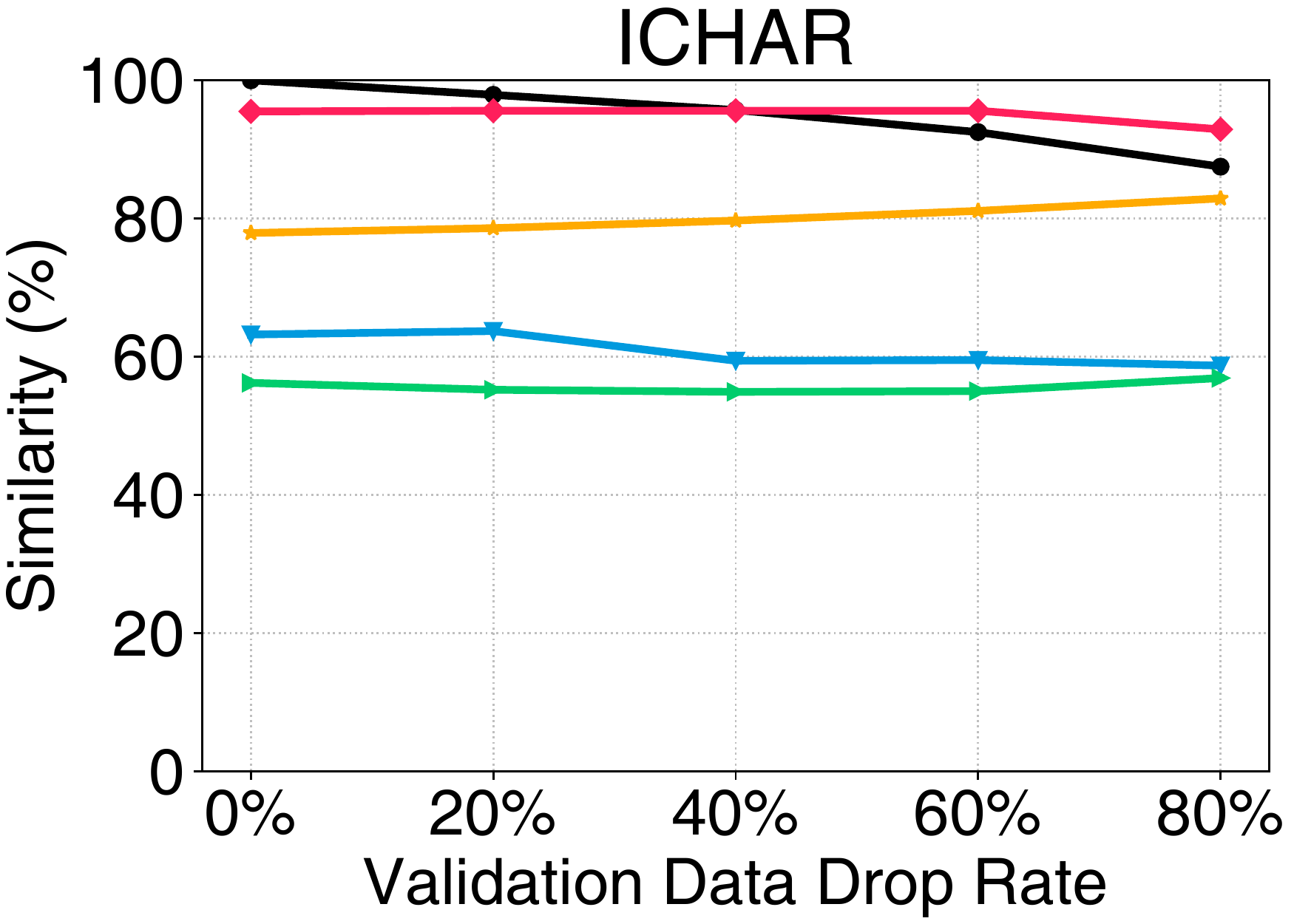}
            \label{fig:imb:sda:act}
        \end{subfigure}
        ~
        \begin{subfigure}[t]{0.25\linewidth}
            \centering
            \includegraphics[width=0.95\linewidth]{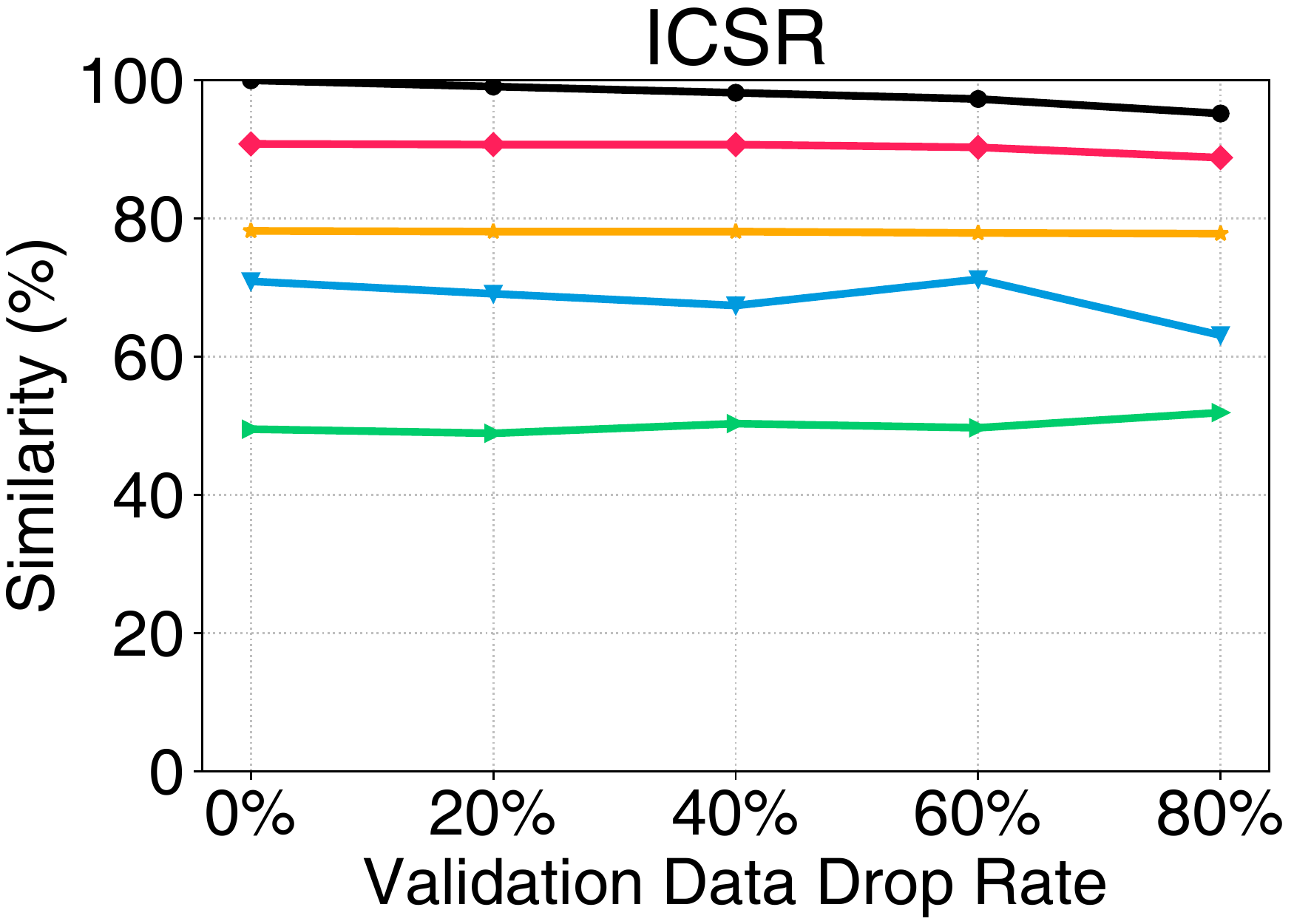}
            \label{fig:imb:sda:spe}
        \end{subfigure}
    \centering
    \end{subfigure}
    
    \caption{Result with imbalanced target validation data.}
    \label{fig:imb}
\end{figure}

Figure~\ref{fig:imb} shows the results with different distributions of the validation data in the fine-tuning scenario (similar trends were found in other adaptation algorithms). Here, we set the TgtLabel with balanced data (0\%) as the oracle and calculated the average similarity with respect to the oracle. Note that TgtLabel is also affected by the target data distribution; even if we use the labeled target data, the estimation becomes gradually inaccurate as data distribution becomes imbalanced. To some extent of imbalance, \system{} showed relatively stable performance. Interestingly, we found that the average similarity degradation in \system{} from 0\% to 80\% drop (5.9\%) was lower than that of the oracle, TgtLabel (7.6\%). We conjecture that this robustness was attributed to the diverse simulated adaptation data (10k) used to train \system{}. We note that the similarity in DIR deteriorated more with higher drop rates than that of EnsRM. We believe EnsRM is more robust to the dataset distribution owing to its ensemble models.

\subsection{On-Device Computational Overhead}~\label{sec:experiment:overhead}
On-device deep learning has been increasingly important owing to its advantages over central training. A variety of private information, such as location, health states, emotions, and identifiable information, could be inferred by leaking personal sensing data~\cite{10.1145/3309074.3309076, 10.1145/2789168.2790121}, which could be protected via on-device training. Additionally, on-device training can save communication bandwidth and cloud management cost~\cite{10.1117/12.2518469, 9111023}. We inspect the computational overhead of \system{} and the baselines in terms of on-device training and inference.

We implemented the adaptation methods, the performance validation baselines, and \system{} with the Mobile Neural Network (MNN) framework by Alibaba~\cite{alibaba2020mnn}. MNN supports both training and inference and achieves state-of-the-art performance with optimized assembly code. We included adaptation methods (Fine-tuning, DANN, CDAN, and SHOT) to compare the adaption time of those algorithms with the computation time of performance validation. FixedEpoch was excluded as it does not have additional computation. SrcLabel and SoftmaxScore were excluded as they had almost the same computation as TgtLabel. We used three commodity smartphones (Google Pixel 5, Google Pixel 2, and LG Nexus 5) for this experiment. We reproduced the experiment with the HHAR dataset under those mobile devices to measure the computational overhead. Both training and inference were made through the CPU with four threads. We set the number of source and target data used for each method as 500 to isolate the impact of the number of data from the algorithmic overhead. The overall trend was kept similar with different numbers of data. For each case, we measured the computational overhead 20 times and reported the average and standard deviation.

\begin{figure}[t]
    \begin{subfigure}[t]{0.44\linewidth}
        \centering
  \includegraphics[width=0.95\linewidth]{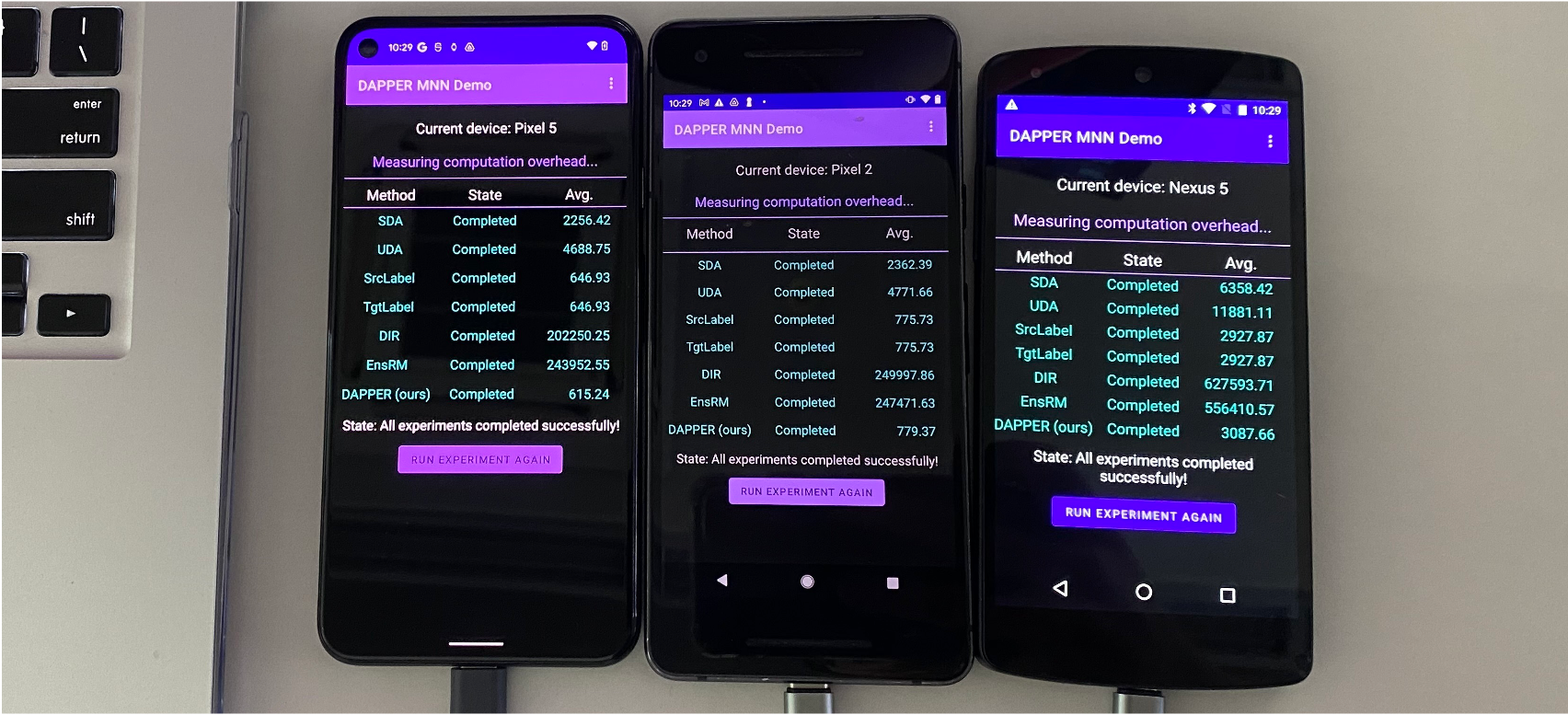}
        \caption{Testbed setup.}
        \label{fig:overhead:testbed}
    \end{subfigure}
    ~
    \begin{subfigure}[t]{0.5\linewidth}
        \centering
  \includegraphics[width=0.95\linewidth]{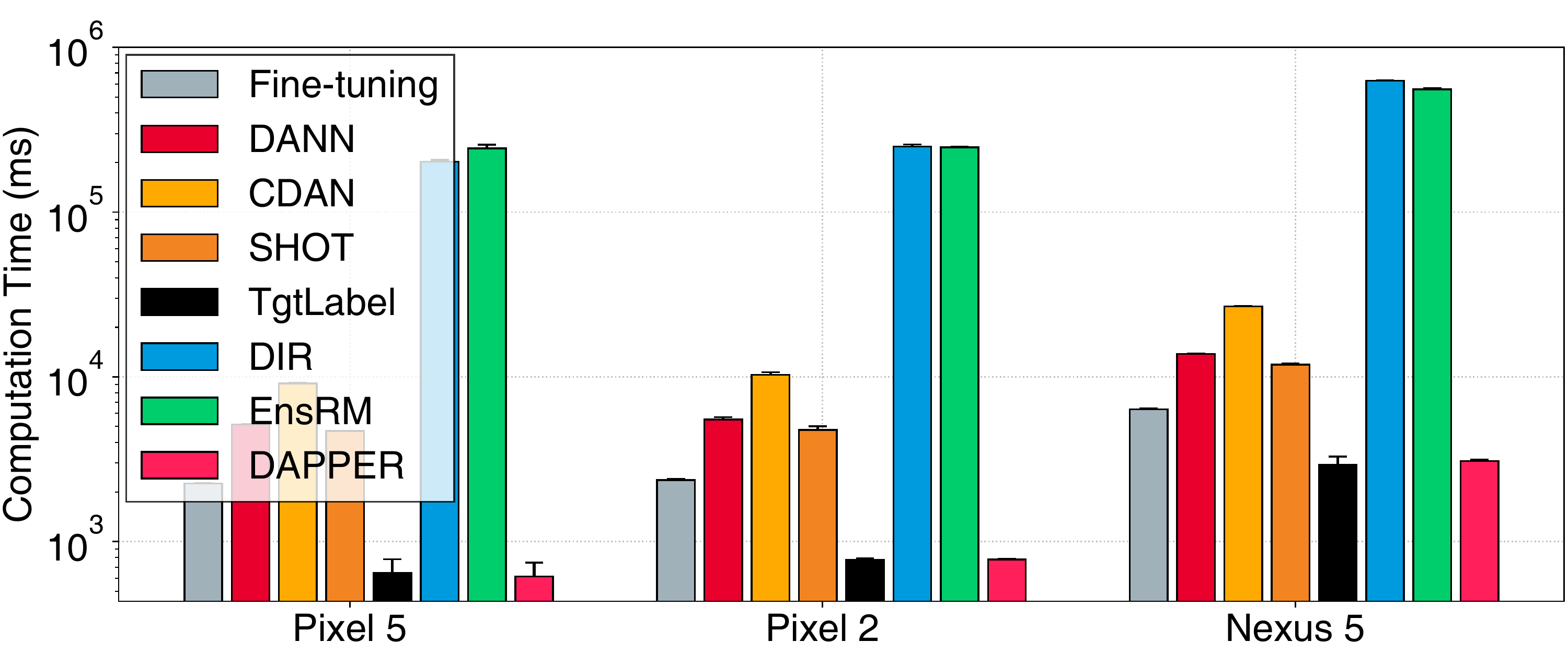}
        \caption{Result of the computational overhead measurement.}
        \label{fig:overhead:result}
    \end{subfigure}
  \caption{On-device computational overhead measurement with three off-the-shelf smartphones (Pixel 5, Pixel 2, and Nexus 5). The y-axis in the result (b) is in the logarithm scale.}
    \label{fig:overhead}
\end{figure}

Figure~\ref{fig:overhead} shows that the computational overhead of the state-of-the-art baselines (DIR and EnsRM) was much larger than that of \system{} (180$\sim$396$\times$ larger) and even larger than the adaptation algorithms. Their computational bottleneck was updating the check model(s) for every adapted model. This suggests that regardless of the estimation accuracy, requiring extra training by the user could harm the user experience with heavy computations. 
TgtLabel and \system{} showed almost identical computation time, as the primary computational overhead is from forwarding the validation samples to the target model.
The computational difference between TgtLabel and \system{} is from \system{} requiring (i) the calculation of the features (GD, IU, and PD) and (ii) additional forward-pass in the estimation network. The calculation of the features took only 0.6$\sim$2.1~ms and additional forwarding took only 0.7$\sim$3.6~ms, depending on the device. Compared with the forward-pass in the target model (613$\sim$3081~ms), additional overhead by \system{} was negligible, which took only around 0.2\% of the total computation in estimation. In summary, not only \system{} accurately estimates the performance, but it also generates affordable computation overhead similar to TgtLabel. 
\section{Related Work}

We summarize existing studies that utilize unlabeled data to validate performance.
Several schemes utilize unlabeled data for performance estimation but require assumptions that might not hold in mobile sensing scenarios. Limberg et al.~\cite{beyond_cross2020} estimated the accuracy of four shallow machine learning classifiers, such as k-Nearest Neighbors,
with a semi-supervised approach for robot object recognition tasks. However, it was experimentally shown that it worked well only when the training and test conditions were the same, which might not be applicable in heterogeneous mobile sensing. 
Platonis et al.~\cite{estimating_accuracy2014} estimated accuracy with unlabeled data by the agreement of multiple boolean functions. Their subsequent work~\cite{DBLP:conf/nips/PlataniosPMH17} improved estimation accuracy given logical constraints among classes. However, both studies were limited to binary classifiers and might be difficult to extend to general multi-class classifiers. 
Steinhardt et al.~\cite{NIPS2016_f2d887e0} estimated unsupervised risk (error) with unlabeled data by assuming three conditionally independent views of a sample are available, which limits the practicality of its approach.

Recent studies utilized deep learning to estimate the performance under domain shifts~\cite{dev, chuang2020estimating, ensrm}. Deep Embedded Validation~(DEV)~\cite{dev} proposed a model selection method by estimating the ratio between the source and the target density (i.e., density ratio). Specifically, DEV predicts the target risk by training a domain discriminator from the unlabeled target data and estimating the density ratio. A recent work utilizing Domain-Invariant Representations (DIR)~\cite{chuang2020estimating} estimates performance under domain shift. DIR estimates the performance of a target model by comparing it with a ``check model.'' Ensemble via Representation Matching (EnsRM)~\cite{ensrm} is the state-of-the-art performance estimation algorithm. Similar to DIR, EnsRM pre-trains a check model. EnsRM further improves the estimation accuracy over DIR using an ensemble of check models. All these algorithms assume that their domain discriminator and check models are accurate, which might not hold in mobile sensing due to a variety of factors affecting performance as explained in~\cref{sec:background}. Note that they require heavy computations after deployment for training the domain discriminator and the check model(s). As DEV focuses on the model selection and does not provide a comparable performance metric, we compared \system{} with DIR and EnsRM in our experiments.

\section{Discussion}\label{sec:discussion}

\subsection{Implications to Mobile Sensing}\label{sec:discussion:implication}
Previous research on the domain shift problem in mobile sensing~\cite{metasense, systematic, scaling, xhar2020, generalization_fitness2021, multi_source2020} has focused on improving the accuracy of DA algorithms. Our work disclosed the pitfall of DA in mobile sensing caused by various factors 
that often lead to performance degradation or unnecessary computational overhead. This phenomenon motivated our research of performance estimation by leveraging only unlabeled data that are naturally obtained in mobile sensing applications. 
This performance estimation framework would be beneficial to design quantitatively efficient DA systems in mobile sensing by selecting better models while reducing 
computational costs.

Furthermore, this study opens up a new direction for human-AI interaction in mobile sensing DA. Under the existing DA frameworks, users have to blindly use mobile sensing applications without knowing how well it operates. Demonstrating the model's performance can be seen as a way of interaction between AI and users, and it is known to improve user trust in AI systems~\cite{10.1145/3290605.3300509, 10.1145/3287560.3287590, 10.1145/3301275.3302277}. This would facilitate involving users in the loop of personalizing AI, making better models for themselves. 
We believe that providing performance is key to enabling a human-AI interaction toward overcoming the domain shift problem in mobile sensing.

\subsection{\rev{Model Calibration}}\label{sec:discussion:calibration}
In order to treat uncertain samples, model calibration~\cite{10.5555/3305381.3305518} aims to make a model produce probability estimates representative of the true correctness likelihood for a given sample. Our objective differs from model calibration in that we estimate the accuracy of the adapted model under domain shift given a number of unlabeled data. One might argue that model calibration could be utilized for our scenario, but we instead take a learning-based approach due to the following limitations of model calibration. First, recent studies found that model calibration significantly deteriorates under domain shift~\cite{10.5555/3454287.3455541, hendrycks2020augmix}. Second, model calibration methods often generate undesirable accuracy degradation~\cite{10.5555/3305381.3305518} or computational overhead~\cite{10.5555/3045390.3045502}. 
Third, calibrated models still have errors while uncalibrated models show a reasonable correlation with the true likelihood~\cite{10.5555/3305381.3305518}. 
Lastly, while designing our method, we conducted a preliminary experiment with a widely-used model calibration method, temperature scaling~\cite{10.5555/3305381.3305518}, to understand its effectiveness in our scenario. We found that it was not very effective, and the case without calibration (i.e., when the temperature was equal to one) on average showed the best correlation with accuracy.

\subsection{Limitations and Extensions}\label{sec:discussion:lim}

\subsubsection{Training and Validation Data Split}
In our experiments, we used 500 validation data for performance estimation. While we focused on
performance estimation with unlabeled data, the ratio between training and validation data could be an important decision, especially for real deployments. 
For instance, given 1,000 unlabeled data collected from a target user, the ratio between the number of training and validation samples could affect both the model's accuracy and the estimation accuracy. Furthermore, when only a few labeled data instances are available, utilizing some of them along with unlabeled data could improve performance estimation. Note that there is a trade-off in selecting the ratio between the training data and validation data; a larger number of training data would result in better adaptation, while a larger number of validation data would provide better estimation, as shown in~\cref{sec:experiment:imbalance}. This is a similar discussion to the efficient training/validation data split in machine learning~\cite{reitermanova2010data, xu2018splitting}, and their valuable findings could provide insights into our scenario. 

\subsubsection{Handling Continuous Data Stream}
We focused on predicting the performance within a single adaptation episode given a set of train/validation data. 
However, for the continuous sensor data streams from the target users, it might involve multiple adaptations as data arrive.
We argue that performance validation becomes even more important in such scenarios due to continuous domain shifts that might cause performance drift over time. We believe our insights and findings could be applicable to the continuous data stream case. However, issues that need to be addressed remain, such as how to manage distribution shifts as time passes, how much data can be kept, what data should be discarded, etc. We leave this direction as future work.

\section{Conclusion}

We investigated the uncertainty of domain adaptation performance in mobile sensing and the importance of performance validation. As a solution, we proposed \system{}, a performance estimation framework that utilizes only unlabeled data from a target domain. As unlabeled data are collected naturally in mobile sensing, it does not require additional manual user effort for labeling. Our evaluation with four real-world sensing datasets showed that \system{} is the most accurate performance estimator compared to the baselines. In addition, our on-device computation analysis showed that \system{} reduced the computational overhead up to 396$\times$ compared with the existing training-based estimation. We believe our study discloses an important problem of performance estimation in mobile sensing domain adaptation and proposes a promising solution that takes a meaningful stride toward overcoming the performance dynamics in mobile sensing.

\begin{acks}

\rev{This work was supported in part by the Institute of Information \& communications Technology Planning \& Evaluation (IITP) grant funded by the Korea government (MSIT) (No.2022-0-00495) and the Institute of Information \& communications Technology Planning \& Evaluation (IITP) grant funded by the Korea government (MSIT) (No. 2022-0-00064).}


\end{acks}

%
%
%


\bibliographystyle{ACM-Reference-Format}
\bibliography{references}

\end{document}